\documentclass{article}
\usepackage[dblblindworkshop,final]{neurips_2026}
\makeatletter
\renewcommand{\@noticestring}{Accepted at the AI for Science workshop (NeurIPS 2026).}
\makeatother

\usepackage{amsmath,amsfonts,bm}

\def\eqref#1{equation~\ref{#1}}

\def\1{\bm{1}}

\DeclareMathAlphabet{\mathsfit}{\encodingdefault}{\sfdefault}{m}{sl}
\SetMathAlphabet{\mathsfit}{bold}{\encodingdefault}{\sfdefault}{bx}{n}

\usepackage[utf8]{inputenc}
\usepackage[T1]{fontenc}
\usepackage{amsmath,amssymb,amsthm}
\usepackage{graphicx}
\usepackage{float}
\usepackage{enumitem}
\usepackage{booktabs}
\usepackage{multirow}
\usepackage[hidelinks]{hyperref}
\usepackage{url}
\usepackage{caption}
\usepackage{placeins}
\usepackage[most]{tcolorbox}
\usepackage{microtype}
\usepackage{titletoc} 
\graphicspath{{./figures/}{./}}

\newcommand{\ResultAMaterialsKendallW}{0.95}
\newcommand{\ResultAProteinsKendallW}{0.69}
\newcommand{\ResultAKendallSysB}{0.80}
\newcommand{\ResultAKendallGBOne}{0.59}
\newcommand{\ResultAKendallGRBTwo}{0.82}
\newcommand{\ResultAKendallYAPOne}{0.91}

\newcommand{\ResultBMaterialsMedianR}{0.94}

\newcommand{\ResultBYAPWeakR}{0.44}

\newcommand{\ResultDPoolEndog}{0.63}
\newcommand{\ResultDLOSOSysBOrig}{-0.18}

\newcommand{\ResultDCrossDirOneSysB}{-0.36}

\newcommand{\ResultEDeltaRatio}{1.000}

\newcommand{\ResultEVordSlopeCovering}{0.38}
\newcommand{\ResultEVordSlopeCoveringLo}{+0.34}
\newcommand{\ResultEVordSlopeCoveringHi}{+0.44}

\newcommand{\ResultProbeEepiProteinsSpearman}{0.75}

\newcommand{\ResultLOSOVarianceOnly}{-0.04}
\newcommand{\ResultLOSOGPcalOnly}{-0.11}
\newcommand{\ResultLOSOVarianceGPcal}{+0.08}
\newcommand{\ResultLOSOVarianceGPcalSpMed}{0.54}

\newcommand{\ResultLOSOVarianceOnlySpMed}{0.48}
\newcommand{\ResultLOSOGPcalOnlySpMed}{0.49}

\newcommand{\ResultAURCOracle}{0.123}
\newcommand{\ResultAURCRandom}{0.156}
\newcommand{\ResultAURCVariance}{0.142}
\newcommand{\ResultAURCGPcal}{0.141}
\newcommand{\ResultAURCVarGPcal}{0.140}
\newcommand{\ResultAURCDisagreement}{0.190}

\newcommand{\ResultAURCVarianceGapClosed}{44}
\newcommand{\ResultAURCGPcalGapClosed}{45}

\newcommand{\ResultNSDesignRatioDoptMax}{4.14}
\newcommand{\ResultNSDesignRsimplex}{6}

\newcommand{\ResultNSVThreeDoptNTvsRandomDelta}{0.011}
\newcommand{\ResultNSVThreeDoptNTvsRandomFrac}{88}
\newcommand{\ResultNSVThreeNTvsGPvarDeltaFifteen}{0.005}
\newcommand{\ResultNSVThreeNTvsGPvarFracFifteen}{67}
\newcommand{\ResultNSVThreeNTvsGPvarSysA}{12}
\newcommand{\ResultNSVThreeNTvsGPvarSysB}{9}
\newcommand{\ResultNSVThreeNTvsGPvarSysC}{7}
\newcommand{\ResultNSVThreeNTvsGPvarSysD}{12}
\newcommand{\ResultNSVThreeLOEODeltaNTGPvarLo}{0.003}
\newcommand{\ResultNSVThreeLOEODeltaNTGPvarHi}{0.005}
\newcommand{\ResultNSVThreeLOEOFracNTGPvarLo}{64}
\newcommand{\ResultNSVThreeLOEOFracNTGPvarHi}{71}

\newcommand{\ResultNSVFiveNTvsPCASixNTevalDelta}{0.015}
\newcommand{\ResultNSVFiveNTvsPCASixNTevalFrac}{88}
\newcommand{\ResultNSVFiveNTvsPCASixMatchedDelta}{0.026}
\newcommand{\ResultNSVFiveNTvsPCASixMatchedFrac}{98}
\newcommand{\ResultNSVFiveHybridvsGPvarDelta}{0.003}
\newcommand{\ResultNSVFiveHybridvsGPvarFrac}{68}
\newcommand{\ResultNSVFiveHybridvsNTDelta}{-0.001}
\newcommand{\ResultNSVFiveHybridvsNTFrac}{42}
\newcommand{\ResultNSVFiveHybridvsRandomDelta}{0.010}
\newcommand{\ResultNSVFiveHybridvsRandomFrac}{82}

\newcommand{\ResultEQuadRestrictedA}{0.16}
\newcommand{\ResultEQuadRestrictedALo}{0.13}
\newcommand{\ResultEQuadRestrictedAHi}{0.19}
\newcommand{\ResultEQuadRestrictedB}{0.21}
\newcommand{\ResultEQuadRestrictedBLo}{0.20}
\newcommand{\ResultEQuadRestrictedBHi}{0.23}
\newcommand{\ResultEQuadRichNaiveA}{0.27}
\newcommand{\ResultEQuadRichNaiveALo}{0.24}
\newcommand{\ResultEQuadRichNaiveAHi}{0.30}
\newcommand{\ResultEQuadRichNaiveB}{0.21}
\newcommand{\ResultEQuadRichNaiveBLo}{0.20}
\newcommand{\ResultEQuadRichNaiveBHi}{0.24}
\newcommand{\ResultEAnalyticalEhSq}{0.214}

\newcommand{\ResultDCrossDirOneBootMed}{0.043}
\newcommand{\ResultDCrossDirOneBootLo}{-0.42}
\newcommand{\ResultDCrossDirOneBootHi}{+0.49}
\newcommand{\ResultDCrossDirOneFisherR}{-0.003}
\newcommand{\ResultDCrossDirOneFisherLo}{-0.38}
\newcommand{\ResultDCrossDirOneFisherHi}{+0.37}
\newcommand{\ResultDCrossDirOneNodePerm}{0.16}

\newcommand{\ResultDLOSODisjointRsq}{-0.53}
\newcommand{\ResultDLOSODisjointSp}{-0.17}
\newcommand{\ResultDLOSOMatchedRsq}{-0.45}
\newcommand{\ResultDLOSOMatchedSp}{-0.03}
\newcommand{\ResultDLOSOEndogRsq}{0.11}
\newcommand{\ResultDLOSOEndogSp}{0.64}
\newcommand{\ResultDLOSOEndogSysBRsq}{-1.76}

\newcommand{\ResultDPoolEndogBootMed}{0.62}
\newcommand{\ResultDPoolEndogBootLo}{+0.37}
\newcommand{\ResultDPoolEndogBootHi}{+0.79}
\newcommand{\ResultDPoolEndogFisherR}{0.54}
\newcommand{\ResultDPoolEndogFisherLo}{+0.31}
\newcommand{\ResultDPoolEndogFisherHi}{+0.71}

\newcommand{\ResultBICCTertileAnova}{0.39}
\newcommand{\ResultBICCContextDomAnova}{0.43}
\newcommand{\ResultBICCRegimeAnova}{0.70}

\newcommand{\ResultBICCTertileJackLo}{0.15}
\newcommand{\ResultBICCTertileJackHi}{0.45}
\newcommand{\ResultBICCContextDomJackLo}{0.26}
\newcommand{\ResultBICCContextDomJackHi}{0.49}
\newcommand{\ResultBICCRegimeJackLo}{0.52}
\newcommand{\ResultBICCRegimeJackHi}{0.73}

\newcommand{\ResultBTertileNodePermR}{0.55}
\newcommand{\ResultBContextDomNodePermR}{0.54}
\newcommand{\ResultBRegimeNodePermR}{0.75}

\newcommand{\ResultMPSupportGapEhull}{+0.023}
\newcommand{\ResultMPSupportFracEhull}{90}
\newcommand{\ResultMPSupportGapBandgap}{+0.006}

\newcommand{\ResultMPMatchedFracEhullLo}{87}
\newcommand{\ResultMPMatchedFracEhullHi}{90}
\newcommand{\ResultMPMatchedFracBandgapLo}{63}
\newcommand{\ResultMPMatchedFracBandgapHi}{80}
\newcommand{\ResultMPRecoveryFracEhull}{26}

\newcommand{\ResultIoptGPvarMSEFrac}{72}
\newcommand{\ResultIoptGPvarMSERatio}{0.94}
\newcommand{\ResultIoptGPvarSpearmanFrac}{65}
\newcommand{\ResultIoptGPvarSpearmanDelta}{+0.003}

\newcommand{\ResultFacRepRatio}{0.75}          
\newcommand{\ResultFacRepWin}{98}              
\newcommand{\ResultFacRepCIlo}{0.67}
\newcommand{\ResultFacRepCIhi}{0.87}
\newcommand{\ResultFacRepRatioMthirty}{0.80}   
\newcommand{\ResultFacRepSysLo}{0.70}          
\newcommand{\ResultFacRepSysHi}{0.79}
\newcommand{\ResultFacSelRatio}{0.89}          
\newcommand{\ResultFacSelWin}{95}
\newcommand{\ResultFacSelCIlo}{0.86}
\newcommand{\ResultFacSelCIhi}{0.92}
\newcommand{\ResultFacSelRatioMthirty}{0.95}   

\newcommand{\ResultFacAcqRatio}{0.998}         
\newcommand{\ResultFacAcqCIlo}{0.992}
\newcommand{\ResultFacAcqCIhi}{1.004}
\newcommand{\ResultFacAcqRatioMthirty}{0.994}
\newcommand{\ResultFacNTvarSp}{0.832}
\newcommand{\ResultFacNTioptSp}{0.832}
\newcommand{\ResultFacGenvarSp}{0.795}
\newcommand{\ResultFacGenioptSp}{0.791}
\newcommand{\ResultFacNullJ}{64}
\newcommand{\ResultFacNullFracStrong}{100}     
\newcommand{\ResultFacFullNullJ}{48}           
\newcommand{\ResultFacFullNullBeatPct}{100}    
\newcommand{\ResultFacFullNullOverlapMed}{0.49}
\newcommand{\ResultFacFullNullRhoMFifteen}{-0.05} 
\newcommand{\ResultFacFullNullRhoMthirty}{-0.10}  
\newcommand{\ResultSynthMisspecRatio}{4.7}     

\newtheorem{proposition}{Proposition}
\newtheorem{corollary}[proposition]{Corollary}
\newtheorem{lemma}[proposition]{Lemma}
\theoremstyle{definition}

\title{Stable but Hidden: Identifiability and Experimental
Design under Missing Compositional Support}
\workshoptitle{AI for Science: Verification in the Age of AI
Scientists at NeurIPS 2026}

\author{%
  Harry Sevi \\
  Altrove, Paris, France \\
  \texttt{harry.sevi@altrove.ai} \\
  \And
  Joonatan E.~M.~Laulainen \\
  Altrove, Paris, France \\
  \texttt{joonatan.laulainen@altrove.ai} \\
}

\begin{document}
\maketitle

\begin{abstract}
AI-guided scientific discovery often uses inexpensive predictors to allocate scarce
verification measurements. We study a structural failure when
relevant components have been observed separately but never jointly: distinct
admissible response functions can agree on all observed compositions while disagreeing
on the missing combinations. Additional same-support data cannot resolve this
ambiguity, and agreement among fitted models need not expose it.

We characterize the unresolved directions as an invisible subspace
$\mathcal{N}_T(\mathcal{F}) = \ker \mathcal{R}_T$, whose geometry determines oracle
zero-shot ambiguity and minimax risk under a symmetric, bounded ambiguity class. When
the missing response lies in an $r$-dimensional, well-conditioned linear basis, optimally designed joint
measurements achieve minimax integrated risk $\Theta(\sigma^2 r/m)$, with classical
$I$-optimality the fixed-design criterion.

We validate on materials. In electrocatalysis, a six-dimensional support-derived
structural basis (a prior on candidate directions, not an identified physical model)
reduces held-out MSE against a dimension-matched generic basis (median paired ratio
$\ResultFacRepRatio$, improving $\ResultFacRepWin\%$ of $60$ element pairs at $m = 15$), while
$I$-optimal and same-basis posterior-variance acquisition show no detected difference.
In Materials Project, removing joint property labels reduces $E_{\text{above hull}}$
ranking on $\ResultMPSupportFracEhull\%$ of $30$ element pairs, more than matched
data-removal controls, under a frozen pretrained representation. Pair difficulty is
reproducible across learners but poorly predicted by zero-label ensemble disagreement;
$20$ joint measurements enable transferable pair-level rankings without reliable
absolute calibration.

In practice, experiment selection should ask which relationships the observed
support leaves unidentified before asking where a fitted model is least certain.
\end{abstract}

\section{Introduction}
\label{sec:intro}

A learner is often asked to predict on a region of input space
whose relevant features never co-occurred in training. In materials
this arises when two elements $A, B$ have been measured
separately but never together; the same setup describes double
mutants inferred from single mutants, drug combinations from
individual compounds, and reactions from unpaired reagents. Let
$T_{AB} = \{x : x_A x_B = 0\}$ be the training support (at least
one of $A, B$ absent) and $C_{AB} = \{x : x_A x_B > 0\}$ the
target region (both present). Standard distribution-shift
analysis asks how a fitted predictor behaves when
$p_{\mathrm{train}} \neq p_{\mathrm{test}}$; we argue that this
regime raises a strictly harder question: whether the observed
data \emph{identify} the target quantity at all.

The difficulty is visible in three lines. Take $h(x) = x_A x_B$: on $T_{AB}$,
$h \equiv 0$; on $C_{AB}$, $h > 0$. So $g$ and $g + \lambda h$ predict
identically on every training input but differ on $C_{AB}$ for any nonzero
admissible $\lambda$: no non-copresence data can identify $\lambda$. A learner
still predicts, but from its inductive bias, not evidence about the interaction;
several learners can share that bias and agree while all being off by
$\lambda h$, so ensemble variance can vanish exactly where the target is least
identified. The obstacle is structural non-identifiability, not train--test
distance in any standard sense.

\emph{Model uncertainty and structural non-identifiability are
different objects.} Model uncertainty concerns variability of
predictions \emph{within} a chosen predictive representation:
finite-sample noise, model-family variance, and so on. Structural
non-identifiability concerns distinctions among admissible
worlds that the observed support itself cannot resolve, even for
an ideal learner with unlimited data on the same support. Pretraining
does not automatically resolve it either: a foundation-model representation
transfers prior knowledge but does not by itself identify a downstream interaction
from a label set whose support omits it (\S\ref{sec:mp-support-ablation},
frozen MACE-MP embeddings).

\textbf{Three questions, answered here.}
\emph{Q1: can the missing interaction be inferred from non-copresence data?}\ \ No: invisible
directions carry information $T_{AB}$-only data cannot identify (\S\ref{sec:prop1};
witness $h = x_A x_B$).
\emph{Q2: can standard uncertainty or model disagreement flag it?}\ \ Not reliably: under
leakage-safe evaluation, zero-label ensemble disagreement fails to predict pair-level
error (\S\ref{sec:empirical-summary}, Appendix~\ref{app:zerolabel}).
\emph{Q3: what experiments repair the gap?}\ \ When $\dim\mathcal{N}_T = r < \infty$, sparse
joint measurements targeting $\mathcal{N}_T$ turn the gap into experimental design, where
$I$-optimal design is exactly minimax (Prop.~\ref{prop:i-opt}); empirically the decisive
lever is \emph{which directions the model represents}, not the acquisition rule
(\S\ref{sec:real-setup}).

These questions separate three often-conflated objects: $D_m$, the error realized
by a fitted learner; $\Delta_{\mathcal{F}}$, the ambiguity the observations leave
across all admissible worlds; and $\widehat{\Delta}_{\mathcal{H}}$, the disagreement
within a chosen finite model family. The third need not reveal the second: several
fitted models can agree through a shared continuation bias while the observations
stay compatible with different responses on $C_{AB}$
(Appendix~\ref{app:three-quantities}). Empirically, realized pair difficulty $D_m$
is reproducible across learners (``stable'') yet zero-label disagreement fails to
predict it (``hidden''; \S\ref{sec:empirical-summary}).

\begin{figure}[!ht]\centering
\includegraphics[width=0.80\linewidth]{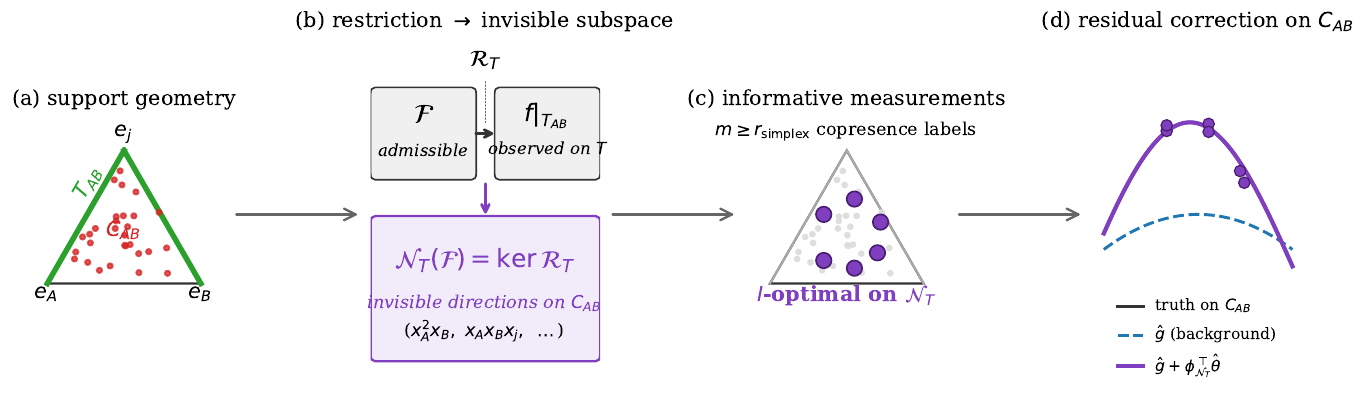}
\caption{\textbf{From missing support to informative measurements.}
(a) Composition simplex: training support $T_{AB}$ (edges, green)
and copresence region $C_{AB}$ (interior, red).
(b) The restriction $\mathcal{R}_T$ collapses all worlds in
$\mathcal{N}_T = \ker \mathcal{R}_T$ to the same $T_{AB}$-only
observations, the invisible subspace no $T_{AB}$-only estimator
can identify.
(c) support-aware design picks $m \geq r$ copresence measurements with full-rank
$\Phi_S$ to resolve $\mathcal{N}_T$; Prop.~\ref{prop:i-opt} makes $I$-optimality the
exact minimax criterion under the finite-basis model.
(d) the residual correction improves prediction on $C_{AB}$ over the background
$\hat g$.}
\label{fig:concept}
\end{figure}

The witness $x_A x_B$ generalizes to all response directions that vanish on the
observed support. Let $\mathcal{R}_T : f \mapsto f|_{T_{AB}}$ be the
support-restriction operator; its kernel
$\mathcal{N}_T(\mathcal{F}) = \ker \mathcal{R}_T$, the invisible subspace, collects
the distinctions no $T_{AB}$-only observation can resolve (Figure~\ref{fig:concept}).
For symmetric linear classes its geometry fixes both the ambiguity among
data-compatible worlds and the minimax zero-shot risk on $C_{AB}$; for
degree-bounded polynomials on the complete two-facet support, $\mathcal{N}_T$ is
spanned by monomials carrying $x_A x_B$ (Lemma~\ref{lem:simplex-dim}).

The missing information can be finite-dimensional: for degree-$\leq 3$ interactions
on the six-component simplex the complete two-facet support has
$r_{\mathrm{geom}} = 6$ invisible directions (Lemma~\ref{lem:simplex-dim}), and on
finite observed support these form a canonical $r_{\mathrm{struct}} = 6$ structural
subspace. Joint measurements then turn impossible zero-shot prediction into
finite-dimensional estimation, making $\mathcal{N}_T$ an experimental-design target
(Prop.~\ref{prop:i-opt}): identifiability-directed design asks which measurement
resolves distinctions the data cannot, not where a fitted model is least certain.

The construction is class-agnostic ($\mathcal{N}_T(\mathcal{F}) = \ker
\mathcal{R}_T$ for any admissible $\mathcal{F}$); turning it into tractable design
needs a usable finite-dimensional representation of the invisible directions,
which we supply via a structural polynomial basis: a \emph{prior} on candidate
directions, not an assumption that the response is polynomial.
\S\ref{sec:theory}--\S\ref{sec:empirical-summary} follow the outline above; full
analyses and exploratory extensions are in the appendices.

\section{Missing-Support Theory}
\label{sec:theory}

The introduction showed one interaction invisible from non-copresence data; this
section makes that general. Even one invisible direction creates unavoidable
zero-shot error (\S\ref{sec:prop1}); agreement among fitted models need not reveal
the ambiguity (\S\ref{sec:corollary}); collecting all invisible directions into
$\mathcal{N}_T = \ker \mathcal{R}_T$ gives a subspace that exactly determines
zero-shot minimax risk (\S\ref{sec:null-space}): given $P_C$ and the admissible
ball $B_\Lambda$, the risk is the squared $L^2(P_C)$ radius of
$\mathcal{N}_T \cap B_\Lambda$ (the kernel alone does not fix a number). Finally,
when $\mathcal{N}_T$ is finite-dimensional, joint measurements convert
non-identifiability into ordinary estimation and experimental design
(\S\ref{sec:prop2}). All proofs are deferred to Appendix~\ref{app:proofs}.

\paragraph{Setting and function class.}
Let $\mathcal{X} \subseteq \mathbb{R}_{\ge 0}^d$, fix coordinates $(A, B)$, and write
$T_{AB} = \{x : x_A x_B = 0\}$ ($x_A = 0$ or $x_B = 0$) and
$C_{AB} = \{x : x_A x_B > 0\}$ (both present). Training data are drawn from $T_{AB}$;
fix a distribution $P_C$ on $C_{AB}$ with norm $\|\cdot\|_{L^2(P_C)}$. The admissible
class for Prop.~\ref{prop:missing-support} is the \emph{coefficient slab} $\mathcal{F}^{\mathrm{slab}}_\Lambda = \{f = g +
\lambda h : g \in \mathcal{G},\, \lambda \in [-\Lambda, \Lambda]\}$, with
$\mathcal{G}$ a background class and $h(x) = 0$ for $x \in T_{AB}$,
$0 < \mathbb{E}_C[h^2] < \infty$ (this slab, the local invisible ball $K_\Lambda$,
and the global norm ball are distinct; Appendix~\ref{app:three-quantities}).
Prop.~\ref{prop:missing-support} grants oracle knowledge of $g$, so no assumption
about recovering $g$ from $T_{AB}$ is needed; observations are
$y = f_{\lambda^\star}(x) + \varepsilon$ with Gaussian $\varepsilon$.

\emph{Structural assumption.} Throughout, $\mathcal{F}$ is specified independently
of the observed responses: the theory asks what that class and the support geometry
make identifiable, not what the responses reveal about the class.

\subsection{One invisible interaction already makes zero-shot prediction impossible}
\label{sec:prop1}

We begin with the strongest test: suppose the learner already knows the
background $g$ exactly, so the only unknown is the coefficient of a single
direction $h$ that vanishes on $T_{AB}$. If prediction is impossible even here,
the obstruction is the support geometry, not model fitting.

\begin{proposition}[Missing-support minimax lower bound]
\label{prop:missing-support}
Fix $\lambda \in (0, \Lambda]$ and a specific background
$g \in \mathcal{G}$. Grant the learner oracle knowledge of $g$: the
estimator may depend on $g$ arbitrarily. For any estimator $\widehat{f}$
that additionally depends only on data $D_T$ drawn from $T_{AB}$,
\begin{equation}
\inf_{\widehat{f}}\;\sup_{s \in \{+, -\}}\;
\mathbb{E}_{D_T \mid f_{s\lambda}}\!\bigl\|\widehat{f} - f_{s\lambda}\bigr\|_{L^2(P_C)}^2
\;\geq\;
\lambda^2\,\mathbb{E}_{x \sim P_C}\bigl[h(x)^2\bigr]. \label{eq:prop1}
\end{equation}
The bound therefore isolates non-identifiability along $h$; no
assumption about ``identifiability from $T_{AB}$'' of $g$ is required.
\end{proposition}

\emph{Interpretation.} The bound is a property of the missing-support geometry and
the admissible class (independent of the estimator, the sample size $n_T$ on
$T_{AB}$, or the shape of $g$) and it is tight: the oracle estimator
$\widehat{f} = g$ attains $\lambda^2 \mathbb{E}_C[h^2]$, so \eqref{eq:prop1} holds
with equality. More $T_{AB}$-data cannot resolve missing support (both endpoints
$g \pm \Lambda h$ stay admissible at every $n_T$); only measurements constraining
invisible directions can (\S\ref{sec:prop2}).

\subsection{Model agreement does not certify identifiability}
\label{sec:corollary}

Uncertainty is often assessed from a finite set of fitted models, so we distinguish
the ambiguity allowed by the data ($\Delta^{\mathrm{emp}}_{\mathcal{F}}$, the
$L^2(P_C)$ diameter of the training-compatible version space
$\mathcal{V}_{\mathcal{F}}(D_T; \varepsilon)$) from the disagreement within a finite
family $\mathcal{H}$ ($\widehat{\Delta}_{\mathcal{H}} \leq
\Delta^{\mathrm{emp}}_{\mathcal{F}}$ since $\mathcal{H} \subseteq
\mathcal{V}_{\mathcal{F}}$; a rank-disagreement statistic is in
Appendix~\ref{app:three-quantities}).

\begin{corollary}[of Proposition \ref{prop:missing-support}]
\label{cor:blind-spot}
Assume $\mathcal{H} \neq \varnothing$,
$\mathcal{H} \subseteq \mathcal{V}_{\mathcal{F}}(D_T; \varepsilon)$,
and that there is a training-compatible $g \in \mathcal{G}$ whose
two endpoints are admissible,
$g \pm \Lambda h \in \mathcal{V}_{\mathcal{F}}(D_T; \varepsilon)$
(automatic for the slab class $\mathcal{F}^{\mathrm{slab}}_\Lambda$; a stated premise
otherwise). Then $\widehat{\Delta}_{\mathcal{H}} \leq \Delta^{\mathrm{emp}}_{\mathcal{F}}$
always. Moreover, if every member $f \in \mathcal{H}$ shares an identical
continuation of a common background $\tilde g$ ($f(x) = \tilde g(x)$ on
$C_{AB}$ for a common $\tilde g \in \mathcal{G}$), and excludes the
direction $h$, then
\begin{equation*}
\widehat{\Delta}_{\mathcal{H}} \;=\; 0
\qquad \text{while} \qquad
(\Delta^{\mathrm{emp}}_{\mathcal{F}})^2 \;\geq\; 4\Lambda^2\,\mathbb{E}_C[h^2].
\end{equation*}
More generally, take any finite family $\mathcal{H}$ whose $C$-side continuations
are fixed independently of $\Lambda$ and disagree only on $g$ (not on $h$). Then
$\widehat{\Delta}_{\mathcal{H}}$ is $O(1)$ in $\Lambda$ (bounded by their
$g$-continuation dispersion), while
$\Delta^{\mathrm{emp}}_{\mathcal{F}} \geq 2\Lambda\,\|h\|_{L^2(P_C)}$ grows
linearly; hence the ratio
$\widehat{\Delta}_{\mathcal{H}}/\Delta^{\mathrm{emp}}_{\mathcal{F}} \to 0$ as
$\Lambda \to \infty$. (The quantifier order matters: $\mathcal{H}$ is held fixed
while $\Lambda$ is enlarged.)
\end{corollary}

Corollary~\ref{cor:blind-spot} formalises the shared blind-spot mechanism: low
empirical disagreement can coexist with large admissible ambiguity when the
ensemble excludes the invisible direction (synthetic witness and a real-data
zero-label failure in Appendices~\ref{app:synth},~\ref{app:zerolabel}).

\subsection{The full invisible subspace determines zero-shot ambiguity}
\label{sec:null-space}

Proposition~\ref{prop:missing-support} states impossibility along one direction
$h$; the fuller object is the whole space of directions vanishing on $T_{AB}$. We
work at the oracle level (treating $f|_T$ as known; Appendix~\ref{app:three-quantities}
relates this to the finite-sample version).

\paragraph{Setup.}
Let $(\mathcal{F}, \|\cdot\|_{\mathcal{F}})$ be a normed linear space with the
symmetric ball $B_\Lambda = \{f \in \mathcal{F} : \|f\|_{\mathcal{F}} \leq
\Lambda\}$. The restriction operator $\mathcal{R}_T f = f|_T$ has kernel (the
\emph{invisible subspace}) $\mathcal{N}_T(\mathcal{F}) = \ker \mathcal{R}_T =
\{h \in \mathcal{F} : h(x) = 0 \ \forall x \in T_{AB}\}$, and \emph{local invisible
ball} $K_\Lambda := \mathcal{N}_T(\mathcal{F}) \cap B_\Lambda$.
For a known $T$-compatible centre $f_0 = E(f|_T)$ (a fixed extension $E$ with
$\mathcal{R}_T E = \mathrm{id}$), the \emph{local oracle fibre} is
$\mathcal{V}_T(f_0) = f_0 + K_\Lambda = f_0 + (\mathcal{N}_T(\mathcal{F}) \cap B_\Lambda)$.
Its diameter below is the \emph{oracle} diameter $\Delta^{\mathrm{orc}}$
(distinct from the coefficient slab, the global norm ball, and the empirical
diameter $\Delta^{\mathrm{emp}}_{\mathcal{F}}$; Appendix~\ref{app:three-quantities}).
This is a centrally symmetric affine set ($f_0 \pm h \in \mathcal{V}_T(f_0)$
since $\mathcal{N}_T$ is linear and $B_\Lambda$ symmetric); every
$f \in \mathcal{V}_T(f_0)$ is observationally equivalent to $f_0$ on $T_{AB}$.

\begin{proposition}[Version-space diameter]
\label{prop:diameter}
Assume $\Lambda > 0$ and that
$K_\Lambda = \mathcal{N}_T(\mathcal{F}) \cap B_\Lambda$ is
norm-compact in $L^2(P_C)$ (automatic for the finite-dimensional
classes used here). The squared $L^2(P_C)$-diameter
of $\mathcal{V}_T(f_0)$ equals four times the maximum squared
$L^2(P_C)$-norm attained by an admissible invisible direction
(equivalently, the diameter equals twice the maximum norm):
\begin{equation}
\operatorname{diam}^2_{L^2(P_C)}\!\bigl(\mathcal{V}_T(f_0)\bigr)
\;=\; 4 \sup_{h \in \mathcal{N}_T(\mathcal{F}) \cap B_\Lambda}
\bigl\|h\bigr\|^2_{L^2(P_C)}.
\label{eq:diameter}
\end{equation}
Consequently, zero-shot prediction on $C$ from oracle knowledge of
$f|_T$ is \emph{structurally identifiable} ($P_C$-almost surely) if
and only if $\|h\|_{L^2(P_C)} = 0$ for every
$h \in \mathcal{N}_T(\mathcal{F})$.
\end{proposition}

Prediction risk is the corresponding \emph{radius} rather than the diameter;
central symmetry of $\mathcal{V}_T(f_0)$ gives the matching minimax result.

\begin{proposition}[Matching minimax risk]
\label{prop:minimax}
Under square $L^2(P_C)$ risk on $C$ and the oracle-observation model
in which the learner knows the \emph{centre} $f_0$ of the local
fibre (equivalently, knows $f|_T$ together with the fixed extension
$E$ above that determines $f_0$ on $C$), the
Chebyshev centre $f_0$ of $\mathcal{V}_T(f_0)$ is the minimax
estimator, and
\begin{equation}
R_{\mathrm{minimax}}\bigl(\mathcal{V}_T(f_0)\bigr)
\;=\;
\inf_{\widehat{f}}\;
\sup_{f \in \mathcal{V}_T(f_0)}\;
\bigl\|\widehat{f} - f\bigr\|^2_{L^2(P_C)}
\;=\;
\sup_{h \in \mathcal{N}_T(\mathcal{F}) \cap B_\Lambda}
\bigl\|h\bigr\|^2_{L^2(P_C)}.
\label{eq:minimax}
\end{equation}
\end{proposition}

Thus the minimax zero-shot risk is exactly the squared $L^2(P_C)$
radius of the admissible invisible set: within this oracle local-fibre
model, the invisible component (the kernel of the restriction operator)
is exactly the source of support-induced ambiguity. This is distinct
from finite-sample estimation uncertainty, which the taxonomy in
Appendix~\ref{app:three-quantities} treats separately.

Two properties of $\mathcal{N}_T$ play different roles: its \emph{radius} on the
target region (the supremum in Eq.~(\ref{eq:minimax})) sets how bad zero-shot
ambiguity can be, while its \emph{dimension} $r$ sets how many degrees of freedom
must be learned once joint measurements are available (\S\ref{sec:prop2}). The
latter presumes $\Sigma_C \succ 0$ (which the rate theorem assumes); if $P_C$
makes some invisible directions irrelevant ($\Sigma_C$ singular), the effective
predictive dimension is the rank of $\Sigma_C$ on $\mathcal{N}_T$, not raw
$\dim\mathcal{N}_T$.

\paragraph{Common function classes.}
Polynomial and anchored-ANOVA classes have closed-form $\mathcal{N}_T$ bases: for
degree-$\leq D$ polynomials on the simplex with complete two-facet support,
$\dim \mathcal{N}_{T_{AB}^{\mathrm{geom}}}(\mathcal{P}_D)|_{\Delta^{d-1}} = \binom{D+d-3}{d-1}$
(Lemma~\ref{lem:simplex-dim}; $r_{\mathrm{simplex}}=6$ for $D=3,d=6$). Anchored-ANOVA,
and why strict supports need not span $\mathcal{N}_T$, are in App.~\ref{app:invisible-bases}.

\subsection{Copresence measurements break the ambiguity}
\label{sec:prop2}

Suppose $\mathcal{N}_T$ has basis $\phi_1, \ldots, \phi_r$; the invisible part is
$h_\theta(x) = \phi(x)^\top \theta$, $\theta \in \mathbb{R}^r$. Each copresence
measurement gives one noisy linear equation
$y_i - g(x_i) = \phi(x)^\top \theta + \varepsilon_i$, turning structural
non-identifiability (\S\ref{sec:prop1}--\ref{sec:null-space}) into an
$r$-parameter design problem. The basis $\phi$ (hence $r$) comes from the
structural class and support geometry, not from the responses; only $\theta$ is
learned, and (since every $\phi_j$ vanishes on $T_{AB}$) copresence
measurements are the first observations to constrain it. Two questions follow:
how well does OLS recover $\theta$ at budget $m$ (Prop.~\ref{prop:few-shot}), and
which $m$ compositions to measure (Prop.~\ref{prop:i-opt})?

\begin{proposition}[Few-shot recovery in a finite invisible basis]
\label{prop:few-shot}
Let $\phi\colon \mathcal{X} \to \mathbb{R}^r$ be a fixed interaction-basis
feature map with $h_\theta(x) = \phi(x)^\top \theta$, $\theta \in \mathbb{R}^r$,
$\phi_j \in L^2(P_C)$. With $g$ known, observe $m$ copresence samples
$y_i = g(x_i) + \phi(x_i)^\top \theta + \varepsilon_i$,
$\varepsilon \sim \mathcal{N}(0, \sigma^2 I_m)$ independent of $X_C$.
Conditioning on the well-conditioning event
$\mathcal{E}_\kappa = \{\lambda_{\min}(\tfrac{1}{m}\Phi^\top\Phi) \geq \kappa\}$
(which implies full rank $r$), OLS satisfies
\begin{align}
\mathbb{E}_{\varepsilon}\bigl[\|\widehat{\theta} - \theta\|_2^2 \,\big|\, X_C\bigr]
  \;&\leq\; \frac{\sigma^2 r}{m \kappa}, \label{eq:prop2-param}\\
\mathbb{E}_{\varepsilon}\bigl[\|\widehat{h}_\theta - h_\theta\|_{L^2(P_C)}^2 \,\big|\, X_C\bigr]
  \;&\leq\; \frac{L\, \sigma^2 r}{m \kappa}, \label{eq:prop2-pred}
\end{align}
where $L = \lambda_{\max}(\Sigma_C)$, $\Sigma_C = \mathbb{E}_{P_C}[\phi\phi^\top]$.
(A high-probability random-design version needs extra assumptions, e.g.\ bounded
features and $\lambda_{\min}(\Sigma_C) > 0$; Appendix~\ref{app:proofs}.)
\end{proposition}

Eq.~(\ref{eq:prop2-param}) is the OLS bound; (\ref{eq:prop2-pred}) follows from
$\|\widehat{h}_\theta - h_\theta\|_{L^2(P_C)}^2 =
(\widehat{\theta} - \theta)^\top \Sigma_C (\widehat{\theta} - \theta)
\leq L \|\widehat{\theta} - \theta\|_2^2$. Estimation is now possible; the
sharper question is which $m$-point design is best.

Unlike the bounded ball $K_\Lambda$ of
Props.~\ref{prop:diameter}--\ref{prop:minimax}, the following fixed-design
theorem takes an \emph{unbounded} $\theta \in \mathbb{R}^r$ (for a bounded
coefficient set the displayed risk remains a uniform upper bound, but exact
bounded-parameter minimax design needs a separate analysis).

\begin{proposition}[Exact minimax I-optimal risk and $\Theta(r/m)$ scaling]
\label{prop:i-opt}
In the setting of Proposition~\ref{prop:few-shot} with
$\theta \in \mathbb{R}^r$, condition on any
fixed $|S| = m$ with $\Phi_S \in \mathbb{R}^{m\times r}$ of rank
$r$. \emph{Exact fixed-design risk:} the minimax risk over all
estimators (not merely unbiased) is
\begin{equation}
\inf_{\widehat h}\sup_{\theta\in\mathbb R^r}
\mathbb{E}_\theta\!\left[\|\widehat h - h_\theta\|_{L^2(P_C)}^2 \,\big|\, S\right]
\;=\; \sigma^2\,\mathrm{tr}\!\left[\Sigma_C\,(\Phi_S^\top\Phi_S)^{-1}\right],
\label{eq:iopt-risk}
\end{equation}
attained by OLS. \emph{Optimal design:} consequently, over the
full-rank design domain $\mathfrak{D}_m = \{S : |S| = m,\
\Phi_S^\top\Phi_S \succ 0\}$ (assumed nonempty, with the minimum
attained; otherwise take the infimum),
$S^\star \in \arg\min_{S \in \mathfrak{D}_m} \mathrm{tr}[\Sigma_C (\Phi_S^\top\Phi_S)^{-1}]$
is the exact minimax measurement design. \emph{Rate:} assume
$\Sigma_C \succ 0$, $\|\phi(x)\|_2^2 \leq B$ for every admissible
candidate $x$, and some admissible design $\widetilde S$ with
$|\widetilde S| = m$ satisfying
$\lambda_{\min}(\tfrac{1}{m}\Phi_{\widetilde S}^\top\Phi_{\widetilde S}) \geq \kappa$.
Then
\begin{equation}
c_-\,\frac{\sigma^2 r}{m}\;\leq\;R(S^\star)\;\leq\;\frac{\sigma^2\,L\,r}{m\,\kappa},
\qquad c_- \;=\; \frac{r\,\det(\Sigma_C)^{1/r}}{B},
\label{eq:iopt-scaling}
\end{equation}
so $R(S^\star) = \Theta(\sigma^2 r / m)$ whenever $B$ and
$\det(\Sigma_C)^{1/r}$ scale so that $c_-$ stays bounded away from
zero and $L/\kappa$ stays bounded above as $r$ varies.
\end{proposition}

\emph{Scope.} The result is for a \emph{fixed, nonadaptive} design (or
one chosen independently of the response noise) over the full-rank
domain $\mathfrak{D}_m$ (rank-deficient designs have infinite minimax
risk when $\Sigma_C \succ 0$); on real data $g$ is estimated and $\phi$
is a structural prior, so the I-optimal criterion is a plug-in,
model-based surrogate, not the exact minimax risk of the real-data
problem, and the minimaxity is worst-case over $\theta \in
\mathbb{R}^r$, not pointwise dominance. The attainment and finite-pool
conditions (nonemptiness, the $\arg\min$-vs-infimum distinction,
compactness for attainment) are stated with the theorem and detailed in
Appendix~\ref{app:proofs}.

Combined with $r_{\mathrm{simplex}} = \binom{D+d-3}{d-1}$, and under the
uniform-conditioning assumptions of Prop.~\ref{prop:i-opt} ($c_{-,r}$ bounded
below, $L_r/\kappa_r$ bounded above, feasible full-rank designs at the needed
budgets; detailed in Appendix~\ref{app:proofs}), reaching target integrated MSE
$\eta$ needs $m = \Theta\!\big(\max\{r,\, \sigma^2 r/\eta\}\big)$, valid within
the finite candidate pool ($m \le |\mathcal{X}_{\mathrm{cand}}|$). This does not
follow from Lemma~\ref{lem:simplex-dim} alone; an $r$-scaling sweep confirms
$\mathrm{MSE} \propto \sigma^2 r/m$ (Fig.~\ref{fig:r-scaling},
Appendix~\ref{app:main-tables}).

Finally, the design basis $\Phi$ typically only \emph{approximates}
$\mathcal{N}_T$. Splitting the true residual $h^\star = \Pi_\Phi h^\star +
h^\star_\perp$ ($\Pi_\Phi$ the $L^2(P_C)$-projection onto $\mathrm{span}(\Phi)$),
OLS on a fixed full-rank design $S$ has the exact risk decomposition
\begin{equation}
\mathbb{E}_\varepsilon \|\widehat h - h^\star\|^2_{L^2(P_C)}
= \|h^\star_\perp\|^2_{L^2(P_C)} + b_S(h^\star_\perp)
+ \sigma^2\,\mathrm{tr}\!\big[\Sigma_C (\Phi_S^\top\Phi_S)^{-1}\big]
\label{eq:misspec}
\end{equation}
(Prop.~\ref{prop:misspec}, Appendix~\ref{app:proofs}): an \emph{approximation}
error, a sample-aliasing term $b_S \geq 0$, and the design \emph{variance}; it
recovers Prop.~\ref{prop:i-opt} when $\Phi$ spans $\mathcal{N}_T$
($h^\star_\perp = 0$). The irreducible term $\|h^\star_\perp\|^2$ grows as the
representation rotates away from $\mathcal{N}_T$, the mechanism behind the
representation effect of \S\ref{sec:real-setup}, validated synthetically
(Appendix~\ref{app:synth-misspec}) and on real data by the $\mathcal{N}_T$-overlap
sweep of Appendix~\ref{app:main-tables}.

\section{From Invisible Directions to Informative Measurements}
\label{sec:null-space-design}

Section~\ref{sec:theory} turned missing support into an $r$-parameter measurement
problem; we now test whether this is useful beyond the idealized theory. A
controlled experiment tests the design rule when the finite-basis model is exactly
correct; electrocatalysis (\S\ref{sec:real-setup}) tests a chosen structural
subspace of $\mathcal{N}_T$ when the basis is only a prior; Materials Project
(\S\ref{sec:mp-support-ablation}) gives the reverse intervention: whether
removing existing joint support creates a deficit matched removals do not.

\paragraph{The measurement basis.} On the six-component simplex the $r = 6$
invisible directions per pair are the pair-gated monomials $x_A^2 x_B$,
$x_A x_B^2$, and $x_A x_B x_j$ ($j$ the four other components;
Lemma~\ref{lem:simplex-dim}). We write $\mathcal{N}_{\mathrm{struct}} :=
\operatorname{span}\{x_A^2 x_B,\, x_A x_B^2,\, x_A x_B x_j\} \subseteq
\mathcal{N}_{T_{\mathrm{obs}}}$ for this six-dimensional pair-gated subspace; on
finite observed support it is a structural subspace ($r_{\mathrm{struct}} = 6$) of
the empirical kernel, so the candidate interaction is $\theta \in \mathbb{R}^{6}$
(figures and tables label $\mathcal{N}_{\mathrm{struct}}$ as $\mathcal{N}_T$). We
fit a background $\hat g$ on $T_{AB}$; each $x_i \in C_{AB}$ supplies a residual
$z_i = y_i - \hat g(x_i) \approx \phi(x_i)^\top \theta + \varepsilon_i$, giving
$z = \Phi_S \theta + \varepsilon$, and the corrected predictor is
$\hat y(x) = \hat g(x) + \phi(x)^\top \widehat\theta$ (Figure~\ref{fig:concept}d).

Which $m$ rows of $\Phi$ to observe? $I$-optimality minimises the integrated
prediction variance $\mathrm{tr}[\Sigma_C (\Phi_S^\top \Phi_S)^{-1}]$ over the
target region \citep{atkinson2007optimum, pukelsheim2006optimal}, which
Prop.~\ref{prop:i-opt} shows is the minimax criterion under the finite-basis
model. This is distinct from GP uncertainty sampling, whose variance reflects a
chosen continuation model rather than the support-unresolved directions.

\paragraph{Controlled check.} On a synthetic where the finite invisible-basis model is
exactly correct ($d{=}6$, $D{=}3$, $r{=}6$; $80$ seeds), $D$/$I$-optimal selection cuts
median MSE by $3.5$--$\ResultNSDesignRatioDoptMax\times$ over random just above $m=r$,
narrowing as $m$ grows (full setup and Table~\ref{tab:null-design} in App.~\ref{app:main-tables}).

\subsection{Electrocatalysis: does the structural prior help on real data?}
\label{sec:real-setup}

The degree-$\leq 3$ invisible basis is a structural prior, not a claim that the
catalyst response is polynomial; we test whether measurements spanning it improve
prediction on unseen pairs. We use the four Rohr et al.\ ACE-I systems A--D
\citep{rohr2020} ($\sim 2200$ ternary/quaternary compositions each on a
six-component simplex; target $\text{FOM} = -\eta$). For each of the
$\binom{6}{2} = 15$ pairs per system ($60$ pairs), $T_{AB}$ is the
non-copresence region and $C_{AB}$ the held-out copresence region ($|C_{AB}|$
median $\sim 656$); the background $\hat g$ is a Mat\'ern-$5/2$ GP fitted per pair
on $T_{AB}$, and its $C_{AB}$ residual is the design target.
\emph{Fixed-test protocol} (offline counterfactual evaluation of prospective
selection; Appendix~\ref{app:main-tables})\emph{:} reserve a common
$U_{AB} \subset C_{AB}$ (random
$50\%$); each method selects $m$ measurements from $C_{AB} \setminus U_{AB}$,
fits the residual, and is evaluated on the same $U_{AB}$ ($15$ seeds;
$m \in \{8, \dots, 30\}$, $m \geq r_{\mathrm{simplex}} = 6$).

\paragraph{Comparisons.} Three controls address alternative explanations: \emph{Random}
(does design help at all?), \emph{GP posterior variance} (does ordinary
uncertainty find the right points?), and \emph{Generic-6D (PCA)} (the same
$D$-optimal criterion in a label-free PCA subspace of the non-$\mathcal{N}_T$
degree-$\leq 3$ features on $X_T$, asking whether any dimension-matched $6$-D
space does as well).
Against these we run $\mathcal{N}_T$-$D$-opt (equivalently, Bayesian
posterior-variance acquisition) and theorem-aligned $\mathcal{N}_T$-$I$-opt in the
analytical invisible basis, with $\widehat\Sigma_C$ estimated transductively from
covariates only; ``theorem-aligned'' is the ridge-regularized Bayesian analogue of
the Prop.~\ref{prop:i-opt} criterion, a finite-sample stabilizer that recovers
exact I-optimality as $\lambda\to0$ (Appendix~\ref{app:main-tables}).

\paragraph{Where does the gain come from? A representation\,$\times$\,acquisition
factorial.} To separate \emph{what to represent} from \emph{how to acquire}, we
cross representation $\in \{\mathcal{N}_T, \text{Generic-6D}\}$ with acquisition
$\in \{\text{pointwise variance (}D\text{-opt)}, \text{integrated } I\text{-opt}\}$,
using the same regularized Bayesian-linear model, split, seeds, and background GP
in all four cells (Table~\ref{tab:realdata-factorial}; $60$ pairs $\times$ $15$
seeds; GP variance and Random are external baselines). A clean three-level
hierarchy results.

\begin{figure}[!ht]\centering
\includegraphics[width=0.80\linewidth]{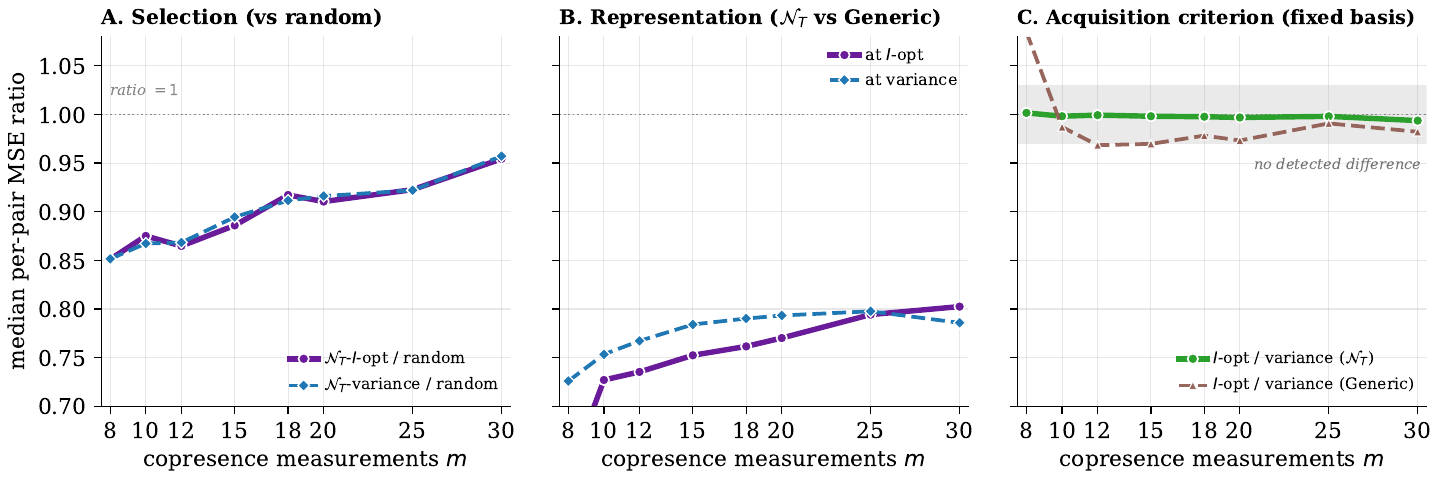}
\caption{\textbf{Decomposing the electrocatalysis gain: representation vs
acquisition} (one locked factorial run; median per-pair MSE ratio vs budget $m$,
reference line at $1$, lower favors the first term; Table~\ref{tab:realdata-factorial}).
\textbf{A,} \emph{selection}: targeted ($\mathcal{N}_T$) beats Random (ratio $<1$,
narrowing as $m$ grows). \textbf{B,} \emph{representation} (dominant): the
$\mathcal{N}_T$ basis beats the Generic-6D basis at either acquisition
($\approx\!\ResultFacRepRatio$, far below $1$, every budget). \textbf{C,}
\emph{acquisition}: once the representation is fixed, $I$-opt vs same-basis variance
stays in a ``no detected difference'' band ($\approx\!\ResultFacAcqRatio$; CI spans
$1$). GP-variance baseline, dependence-aware CIs, and per-system splits in
Appendix~\ref{app:main-tables}.}
\label{fig:learning-curve}
\end{figure}

\emph{(A) Targeted selection helps.} Holding the representation at
$\mathcal{N}_T$, targeted selection beats Random with median paired MSE ratio
$\ResultFacSelRatio$ at $m = 15$ ($\ResultFacSelWin\%$ of pairs, element-cluster
$95\%$ CI $[\ResultFacSelCIlo, \ResultFacSelCIhi]$), shrinking to
$\ResultFacSelRatioMthirty$ at $m = 30$ as random measurement catches up
(Figure~\ref{fig:learning-curve}A).

\emph{(B) The support-derived representation dominates.} Holding acquisition
fixed, the $\mathcal{N}_T$ basis beats the dimension-matched Generic-6D basis by
median MSE ratio $\ResultFacRepRatio$ at $m = 15$ ($\ResultFacRepWin\%$ of pairs,
element-cluster $95\%$ CI $[\ResultFacRepCIlo, \ResultFacRepCIhi]$), consistent
across all four systems (per-system ratios
$\ResultFacRepSysLo$--$\ResultFacRepSysHi$) and every leave-one-element-out fold.
And it is not merely a matter of \emph{including} interaction features: the
$\mathcal{N}_T$ basis beats the median of $\ResultFacFullNullJ$ random $6$-dim
bases drawn from the \emph{full} ambient degree-$\leq 3$ space (which can contain
interaction structure and partially overlap $\mathcal{N}_T$) on
$\ResultFacFullNullBeatPct\%$ of pairs, and greater $\mathcal{N}_T$-overlap
predicts lower MSE (Appendix~\ref{app:main-tables}). This is the largest effect
in the experiment.

\emph{(C) The acquisition criterion barely matters once the representation is
fixed.} Within the $\mathcal{N}_T$ basis we detect no difference between
integrated $I$-opt and pointwise posterior-variance (uncertainty) acquisition:
the median MSE ratio is extremely close to one ($\ResultFacAcqRatio$ at $m = 15$,
element-cluster $95\%$ CI $[\ResultFacAcqCIlo, \ResultFacAcqCIhi]$ spans $1$; no
equivalence test performed, so this is a null-difference finding, not a proof of
equivalence). The external GP-variance baseline (uncertainty geometry set by its continuation
kernel, not the invisible subspace) selects different measurements
(Jaccard $\in [0.05, 0.20]$; Table~\ref{tab:uncertainty-vs-ident}) and is worse on
paired MSE ($\mathcal{N}_T$-$I$-opt lower on $\ResultIoptGPvarMSEFrac\%$ of pairs,
ratio $\ResultIoptGPvarMSERatio$). So representation, not the acquisition criterion,
is the dominant mechanism.

\subsection{Materials Project: a support-removal counterfactual}
\label{sec:mp-support-ablation}

Materials Project (MP) supplies the support-removal counterpart: when joint
support exists, does removing it hurt more than matched removals do? For $30$
element pairs ($|C_{AB}| \geq 40$) from a $7467$-structure near-stable MP pool
\citep{jain2013materials}, with frozen MACE-MP embeddings
\citep{batatia2025foundation} and ridge/HGB predictors for $E_{\text{above hull}}$
and band gap, we compare full support against three quantity-matched removals:
random $T$-side (data-volume), nearest-neighbour to
$\overline{U_{AB}}$ (covariate-proximity), and another pair's region $C_{A'B'}$
(pair-specificity).

Deleting $C_{AB}$ drops per-pair Spearman on $U_{AB}$ (median support gap
$G_{AB} = \ResultMPSupportGapEhull$ on $E_{\text{above hull}}$,
$\ResultMPSupportFracEhull\%$ of pairs positive; $\ResultMPSupportGapBandgap$ on
band gap). None of the three matched removals reproduces it: zero-support loss
exceeds each matched-control loss on
$\ResultMPMatchedFracEhullLo$--$\ResultMPMatchedFracEhullHi\%$ of pairs for
$E_{\text{above hull}}$ (and $\ResultMPMatchedFracBandgapLo$--$\ResultMPMatchedFracBandgapHi\%$
for band gap); $k = 32$ random joint labels recover a median
$\ResultMPRecoveryFracEhull\%$ of the gap on the pre-declared subset, HGB
replicates, and element-cluster bootstrap / LOEO preserve the paired claim
(Appendix~\ref{app:mp-support}).
Absent joint support, $\mathcal{N}_T$-targeted measurements help;
present joint support, selectively removing it hurts more than
matched removals do.

\section{Pair Difficulty Is Stable but Hidden}
\label{sec:empirical-summary}

Not every pair suffers equally: can we tell which before spending the joint budget?
Realized difficulty $D_m^{\mathrm{rank}}$ (aggregated across GP, RF, and GBM,
leakage-safe; Appendix~\ref{app:calibration}) is stable: rankings agree across
learners ($W = \ResultAMaterialsKendallW$, $\geq 0.80$ in every system) and transfer
across third-element and ternary/quaternary splits. Yet zero-label ensemble
disagreement does \emph{not} ($\rho_S = \ResultDLOSOMatchedSp$,
$R^2 = \ResultDLOSOMatchedRsq$; Appendix~\ref{app:calibration}), consistent with
Cor.~\ref{cor:blind-spot} (the ``hidden''), while $k = 20$ random copresence labels
($\approx 3\%$ of $C_{AB}$) make the ordering transfer (LOSO $\rho_S$ up to $+0.54$).
So pair difficulty is a routing signal, not an acquisition score: it flags which
pairs merit measurement, not which composition to acquire within one.

\section{Related Work}
\label{sec:related}

\paragraph{Identifiability, multiplicity, and compositional
generalization.} Partial-identification theory
\citep{manski2003partial,tamer2010partial}, predictive-multiplicity /
underspecification work
\citep{fisher2018all,semenova2022existence,d2022underspecification}, and
classical incomplete factorial design \citep{byar1993incomplete}, which
characterises the interaction contrasts a chosen subset of combinations leaves
unestimable, study ambiguity compatible with the observations; we
specialise this to a concrete, support-induced object, the invisible subspace
$\mathcal{N}_T = \ker \mathcal{R}_T$, whose radius sets the zero-shot minimax risk
(Props.~\ref{prop:diameter},~\ref{prop:minimax}) and whose dimension controls
few-shot recovery (Prop.~\ref{prop:i-opt}). This differs from compositional generalization
\citep{brady2024}, compositional-intervention identifiability
\citep{yu2024compositional}, and materials-OOD benchmarks
\citep{li2025probing,omee2024structure,segal2025known}, where admissible
functions can agree on the \emph{entire} observed support yet differ on the
missing combination, even with oracle knowledge of the response on that support.

\paragraph{Uncertainty and experimental design.} Ensemble methods treat variation
among fitted predictors as uncertainty \citep{lakshminarayanan2017simple}, but
finite-family agreement does not certify identifiability when all admissible models
share the same continuation bias (Cor.~\ref{cor:blind-spot}). Classical
$I$-/$D$-optimal design \citep{atkinson2007optimum,pukelsheim2006optimal} and
scientific active-learning platforms for drug combinations \citep{tosh2025batchie}
select \emph{where} (or which combinations) to measure under a given feature
space $\phi$. Our contribution is not $\ker \mathcal{R}_T$ as an object (a null
space is elementary) but the step \emph{upstream}: the observed support, relative
to the admissible class, determines \emph{what} is unidentified, hence
\emph{which} subspace a \emph{support-derived experimental representation} must
span, one that ordinary predictive uncertainty may omit. Empirically
(\S\ref{sec:real-setup}) that choice dominates the acquisition rule: the
$\mathcal{N}_T$ basis beats a dimension-matched generic one, while rules
\emph{inside} it ($I$-optimal vs same-basis posterior variance) are nearly
interchangeable (Prop.~\ref{prop:misspec}). Identifying which distinctions the
observed support leaves unresolved is thus more decisive than any acquisition
criterion. A partial cross-domain replication in protein epistasis
\citep{olson2014,faure2022mapping}, with the materials-specific claims stated
explicitly, is in Appendix~\ref{app:crossdomain}.

\section{Conclusion}
\label{sec:conclusion}

Within an admissible response class, missing compositional support can create
directions no learner can identify from same-support data, regardless of
sample or ensemble size. The invisible subspace
$\mathcal{N}_T(\mathcal{F}) = \ker \mathcal{R}_T$ makes them explicit: its
target-region radius sets zero-shot minimax risk, its dimension the degrees of
freedom once joint measurements are available, and finite-model
agreement need not reveal the ambiguity, so ordinary uncertainty signals can be
silent where the target is least identified. When $\dim\mathcal{N}_T = r < \infty$,
$I$-optimal measurements achieve $\Theta(\sigma^2 r/m)$ minimax risk, distinct from
fitted-model variance. Empirically, the support-derived directions are the
dominant lever (MSE ratio $\ResultFacRepRatio$ vs a generic basis), the acquisition
rule within them nearly immaterial ($\ResultFacAcqRatio$). The
principle is representation- and domain-independent
(the basis a structural prior, not a physical model; evidence strongest in one
materials setting): before asking where a predictor is uncertain, ask whether the
observations identify the target; if not, the experiment must change the support.

\appendix

\startcontents[appendixtoc]
{\noindent\large\textbf{Appendix Contents}\par\nobreak\vspace{0.5em}}
{\small\printcontents[appendixtoc]{}{1}{\setcounter{tocdepth}{2}}}
\vspace{0.5em}\hrule\vspace{1.5em}

\section{The three quantities and notation}
\label{app:three-quantities}

\emph{Question.} Why does the paper carry three distinct
error-like quantities ($D_m$, $\Delta_{\mathcal{F}}$, and
$\widehat{\Delta}_{\mathcal{H}}$), rather than one, and how are they
related?

\emph{Role.} This appendix is the paper's notation contract. It
defines each object, states the exact identities that link them
under missing compositional support, and lists the specific
non-implications that the main text turns into empirical claims. All
subsequent appendices use these definitions verbatim.

\subsection{Why three quantities are needed}
\label{app:three-quantities:why}

Consider the witness of \S\ref{sec:intro}: given a background $g$ on
$T_{AB}$, an observationally equivalent world is
\[
    f_\lambda(x) \;=\; g(x) + \lambda\, x_A x_B,
    \qquad \lambda \in [-\Lambda, \Lambda].
\]
All $f_\lambda$ agree on $T_{AB}$ and differ on $C_{AB}$. Three
questions about the situation are distinct:
\[
    \boxed{\;
    \begin{aligned}
        D_m &: \;\text{How wrong did learner $m$ actually turn out to be?} \\
        \Delta_{\mathcal{F}} &: \;\text{How much do all admissible, data-compatible worlds differ?} \\
        \widehat{\Delta}_{\mathcal{H}} &: \;\text{How much do the particular fitted models differ?}
    \end{aligned}
    \;}
\]
The first is realized. The second is structural: a worst-case
property of $T_{AB}$ and the admissible class $\mathcal{F}$,
independent of any fitted model. The third is what an ensemble on
the table can be measured for. The paper studies when the third
fails to reveal the second (\S\ref{sec:empirical-summary}), and when
a small number of joint measurements can be used to resolve the
second directly (\S\ref{sec:null-space-design}).

\subsection{Formal definitions}
\label{app:three-quantities:defs}

Let $\mathcal{F}$ be the admissible function class,
$\mathcal{F}^{\mathrm{glob}}_\Lambda \subset \mathcal{F}$ its global
norm-bounded submodel (functions bounded by $\Lambda$ under the
class's norm), $B_\Lambda \subset \mathcal{F}$ the ball of radius
$\Lambda$ around zero used to constrain \emph{invisible
perturbations} around a fixed representative, $D_T = \{(x_i, y_i)\}$
the training data supported on $T_{AB}$, and $P_C$ the evaluation
distribution on $C_{AB}$. The symbols $\mathcal{F}^{\mathrm{glob}}_\Lambda$
and $B_\Lambda$ play different roles: $\mathcal{F}^{\mathrm{glob}}_\Lambda$
is a global size constraint on the entire function;
$B_\Lambda$ is a local size constraint on the invisible perturbation
direction, giving $K_\Lambda = \mathcal{N}_T(\mathcal{F}) \cap
B_\Lambda$. Neither coincides with the coefficient slab
$\mathcal{F}^{\mathrm{slab}}_\Lambda$ used in
Prop.~\ref{prop:missing-support}. Two version spaces appear in the
paper and must be distinguished; we use permanently distinct
diameter symbols $\Delta^{\mathrm{emp}}_{\mathcal{F}}$ (empirical,
below) and $\Delta^{\mathrm{orc}}$ (oracle local fibre,
Prop.~\ref{prop:diameter}).

\paragraph{Finite-sample empirical version space.}
Fix a \emph{named} admissible set $\mathcal{A}$ (either the
coefficient slab $\mathcal{F}^{\mathrm{slab}}_\Lambda$ or the global
ball $\mathcal{F}^{\mathrm{glob}}_\Lambda$) and a noise tolerance
$\varepsilon \ge 0$. Then
\[
    \mathcal{V}^{\mathrm{emp}}_{\mathcal{A}}(D_T; \varepsilon)
    \;=\;
    \bigl\{ f \in \mathcal{A} :
    \tfrac{1}{n}\textstyle\sum_i (f(x_i) - y_i)^2 \le \varepsilon^2 \bigr\};
    \qquad
    \Delta^{\mathrm{emp}}_{\mathcal{A}}
    := \operatorname{diam}_{L^2(P_C)} \mathcal{V}^{\mathrm{emp}}_{\mathcal{A}}(D_T; \varepsilon),
\]
subscripted by the specific class $\mathcal{A}$ because the slab and
the global ball can have different diameters at the same data and
$\Lambda$ (Corollary~\ref{cor:blind-spot} uses the slab). Where the
choice is fixed and clear we abbreviate
$\mathcal{V}_{\mathcal{F}}(D_T;\varepsilon) \equiv
\mathcal{V}^{\mathrm{emp}}_{\mathcal{A}}(D_T;\varepsilon)$ and
$\Delta^{\mathrm{emp}}_{\mathcal{F}} \equiv \Delta^{\mathrm{emp}}_{\mathcal{A}}$.
The subscript records that the diameter is taken
over the chosen admissible class $\mathcal{A}$. The slab makes the
$g \pm \Lambda h$ endpoint membership of Corollary~\ref{cor:blind-spot}
automatic; the global ball is used where explicitly stated, and there
endpoint membership is an assumed premise, not a symmetry consequence.
This is the object empirical procedures approximate; it can depend
on realized responses and $\varepsilon$, contains ordinary
in-support estimation uncertainty, and is in general \emph{not}
equal to the oracle fibre below.

\paragraph{Oracle support-compatible version space.}
Given a $T_{AB}$-compatible representative $f_0$,
\[
    \boxed{\;
    \mathcal{V}_T(f_0)
    \;:=\;
    f_0 \;+\; \bigl( \mathcal{N}_T(\mathcal{F}) \cap B_\Lambda \bigr).
    \;}
\]
Every member of $\mathcal{V}_T(f_0)$ has the same restriction as
$f_0$ on $T_{AB}$. Conversely, within this local perturbation
model, every admissible support-equivalent world differs from
$f_0$ by an element of $\mathcal{N}_T(\mathcal{F}) \cap B_\Lambda$.
The theory in \S\ref{sec:theory} operates on
$\mathcal{V}_T(f_0)$: it isolates the ambiguity created by missing
joint support \emph{after} granting oracle knowledge of the
restriction on $T_{AB}$. In finite-sample regimes the empirical
version space also carries ordinary in-support estimation
uncertainty; the paper does not conflate the two.

\paragraph{Realized error of a single learner.}
For a specific fitted predictor $\hat f_m$,
\[
    D_m(A, B) \;:=\; \mathbb{E}_{x \sim P_C}\bigl[(\hat f_m(x) - f^\star(x))^2\bigr].
\]

\paragraph{Consensus difficulty.}
For a panel of learners $\mathcal{M} = \{\hat f_m\}_{m=1}^M$,
$D_{\mathrm{cons}}(A, B)$ denotes the explicitly specified
learner-aggregated difficulty target used in analyses that require
one value per pair (a rank or a value; the aggregation rule is
stated at each point of use, e.g.\ \S\ref{sec:empirical-summary}
and Appendix~\ref{app:stability}). Kendall's $W$ is \emph{not}
$D_{\mathrm{cons}}$: $W$ is the concordance statistic testing
whether the learner-specific $D_m$ rankings are compatible enough
to justify forming such an aggregate.

\paragraph{Admissible ambiguity (oracle diameter).}
The theoretical quantity used by Propositions~\ref{prop:diameter}
and~\ref{prop:minimax} is the \emph{oracle} diameter
\[
    \Delta^{\mathrm{orc}}(A, B)
    \;:=\;
    \operatorname{diam}_{L^2(P_C)}
    \mathcal{V}_T(f_0)
    \;=\;
    \sup_{f_1, f_2 \in \mathcal{V}_T(f_0)}
    \|f_1 - f_2\|_{L^2(P_C)}
    \;=\;
    \sup_{h_1, h_2 \in K_\Lambda}
    \|h_1 - h_2\|_{L^2(P_C)}
\]
(independent of the choice of $f_0$: $\mathcal{V}_T(f_0)$ is
symmetric about $f_0$, so its diameter is a property of the
invisible perturbation set alone). We keep $\Delta^{\mathrm{orc}}$
permanently distinct from the empirical version-space diameter
$\Delta^{\mathrm{emp}}_{\mathcal{F}}$; the two are not in general
equal, and only $\Delta^{\mathrm{orc}}$ enters the minimax identity
$R_{\mathrm{minimax}} = \tfrac14 (\Delta^{\mathrm{orc}})^2$. Both
denote \emph{unsquared} $L^2$ diameters unless squared explicitly.

\paragraph{Finite-ensemble diameter (theory-side).}
For a chosen finite hypothesis family
$\mathcal{H} \subseteq \mathcal{F}$ (an ensemble, a GP with a
specific kernel, a set of trained models),
\[
    \widehat{\Delta}_{\mathcal{H}}(A, B)
    \;:=\;
    \operatorname{diam}_{L^2(P_C)}
    \bigl\{\hat f_h : h \in \mathcal{H}\bigr\}
    \;=\;
    \max_{h_1, h_2 \in \mathcal{H}}
    \|\hat f_{h_1} - \hat f_{h_2}\|_{L^2(P_C)}.
\]
This is the object Corollary~\ref{cor:blind-spot} bounds.

\paragraph{Practical average disagreement (empirics-side).}
Two zero-label ensemble statistics appear in the empirical
sections; both are averages, not diameters.
\begin{align*}
    V_{\mathcal{H}}^{\mathrm{avg,\,value}}(A, B)
    &\;:=\;
    \frac{1}{\binom{M}{2}}
    \sum_{i < j}
    \|\hat f_i - \hat f_j\|_{L^2(P_C)}^2, \\[2pt]
    V_{\mathcal{H}}^{\mathrm{avg,\,rank}}(A, B)
    &\;:=\;
    1 \;-\;
    \frac{1}{\binom{M}{2}}
    \sum_{i < j}
    \rho_S\!\bigl(\hat f_i(C_{AB}),\, \hat f_j(C_{AB})\bigr),
\end{align*}
where $\rho_S$ is Spearman rank correlation. Neither equals
$\widehat{\Delta}_{\mathcal{H}}$. The value statistic admits the
safe upper bound
$V_{\mathcal{H}}^{\mathrm{avg,\,value}} \le \widehat{\Delta}_{\mathcal{H}}^{\,2}$
(the maximum is at least the average); no analogous $L^2$-diameter
inequality is available for the rank statistic. The empirical
zero-label study of \S\ref{sec:empirical-summary} uses
$V_{\mathcal{H}}^{\mathrm{avg,\,rank}}$ as its primary operational
score; Corollary~\ref{cor:blind-spot} concerns
$\widehat{\Delta}_{\mathcal{H}}$. Neither should be identified with
the other.

\subsection{Exact mathematical relationships}
\label{app:three-quantities:identities}

Two identities pin down the geometric meaning of
$\Delta^{\mathrm{orc}}$ and its relation to minimax risk. Assume
the symmetric linear setting of
Propositions~\ref{prop:diameter}--\ref{prop:minimax}: $\mathcal{F}$
is a Banach space, $B_\Lambda$ is the local invisible-perturbation
ball, and the oracle version space around a $T_{AB}$-compatible
representative $f_0$ is
$\mathcal{V}_T(f_0) = f_0 + \bigl( \mathcal{N}_T(\mathcal{F}) \cap
B_\Lambda \bigr)$.

\paragraph{Diameter identity (Prop.~\ref{prop:diameter}).}
\[
    \Delta^{\mathrm{orc}}(A, B)^2
    \;=\;
    4 \sup_{h \in \mathcal{N}_T(\mathcal{F}) \cap B_\Lambda}
    \|h\|_{L^2(P_C)}^2.
\]
The factor $4$ is the diameter/radius conversion: the version space
is symmetric about its centre, its diameter is twice the largest
invisible direction in $B_\Lambda$, and squaring gives the four.
Symmetry is the only property used; compactness of $K_\Lambda$ is
needed only to attain the supremum as a maximum.
\emph{Identifiability converse (scaling step).} Suppose some
$h \in \mathcal{N}_T(\mathcal{F})$ has $0 < \|h\|_{L^2(P_C)} < \infty$
but $h \notin B_\Lambda$. Since $\Lambda > 0$, set
$\widetilde h = t h$ with $0 < t \le \Lambda / \|h\|_{\mathcal{F}}$;
then $\widetilde h \in K_\Lambda$ and
$\|\widetilde h\|_{L^2(P_C)} = t\|h\|_{L^2(P_C)} > 0$, so the
diameter is strictly positive and the target is not identifiable.
Hence identifiability is equivalent to $\|h\|_{L^2(P_C)} = 0$ for
every $h \in \mathcal{N}_T(\mathcal{F})$.

\paragraph{Minimax identity (Prop.~\ref{prop:minimax}).}
\[
    R_{\mathrm{minimax}}(A, B)
    \;=\;
    \sup_{h \in \mathcal{N}_T(\mathcal{F}) \cap B_\Lambda}
    \|h\|_{L^2(P_C)}^2.
\]
The minimax estimator on $C_{AB}$ returns the centre of the version
space; its worst-case squared error is then the radius squared.

\paragraph{Combining these.}
\[
    \boxed{\;
    R_{\mathrm{minimax}}(A, B)
    \;=\;
    \tfrac{1}{4}\,\Delta^{\mathrm{orc}}(A, B)^2.
    \;}
\]
So the three propositions play different roles:
\begin{itemize}[leftmargin=1.6em]
    \item Proposition~\ref{prop:missing-support} is a simpler
    two-world lower bound. A single invisible direction
    $h \in \mathcal{N}_T$ with $\mathbb{E}_C[h^2] > 0$ already forces
    a lower bound of order $\Lambda^2 \mathbb{E}_C[h^2]$ on any
    estimator's worst-case squared error, without invoking the
    full diameter identity.
    \item Proposition~\ref{prop:diameter} identifies the oracle
    version-space diameter with the invisible-subspace diameter.
    \item Proposition~\ref{prop:minimax} then converts the diameter
    into the matching minimax risk via the $1/4$ factor above.
\end{itemize}

\subsection{Non-implications and paper map}
\label{app:three-quantities:non-impl}

Realized error, admissible ambiguity, and empirical disagreement do
not imply each other. The paper studies three specific gaps.

\paragraph{Shared blind spot.}
$\widehat{\Delta}_{\mathcal{H}}$ small
$\;\not\Rightarrow\;$
$\Delta^{\mathrm{emp}}_{\mathcal{F}}$ small.
Corollary~\ref{cor:blind-spot} constructs finite families
$\mathcal{H}$ whose members agree on $C_{AB}$ while the admissible
class still contains the full $\pm \Lambda h$ separation along an
invisible direction $h \in \mathcal{N}_T$, so the empirical admissible
ambiguity $\Delta^{\mathrm{emp}}_{\mathcal{F}}$ remains large. Empirically,
training-compatible ensembles fail to predict $D_{\mathrm{cons}}$ at
leave-one-system-out scale (median LOSO
$R^2 = \ResultDLOSODisjointRsq$;
Appendix~\ref{app:zerolabel}).

\paragraph{Agreement without accuracy.}
$\widehat{\Delta}_{\mathcal{H}}$ small
$\;\not\Rightarrow\;$
$D_m$ small.
The implication fails constructively in
Corollary~\ref{cor:blind-spot} and in the controlled shared-blind-spot
synthetic of Appendix~\ref{app:synth}: two learners can agree
tightly on $C_{AB}$ while both are wrong because they miss the same
invisible direction. The Materials Project intervention
(\S\ref{sec:mp-support-ablation}) provides complementary evidence
that removing relevant joint support degrades downstream prediction
across learners; it does not, by itself, exhibit a shared invisible
direction.

\paragraph{Structural ambiguity is worst-case.}
$\Delta^{\mathrm{orc}}$ large
$\;\not\Rightarrow\;$
$D_m$ is large for every realized world.
$\Delta^{\mathrm{orc}}$ is a worst-case quantity over the admissible
class; the true $f^\star$ may lie near the continuation a particular
learner happens to pick, so $D_m$ can be small even when the
version space is wide. Proposition~\ref{prop:minimax} says no
estimator is uniformly small over the version space, not that every
predictor is large on every draw.

\paragraph{Paper map.}
\[
    \boxed{\;
    \begin{array}{lcl}
        \S\ref{sec:theory} &:& \Delta^{\mathrm{orc}}\ \text{and}\ \mathcal{N}_T \\
        \S\ref{sec:null-space-design} &:& \text{measurements targeting } \mathcal{N}_T \\
        \S\ref{sec:empirical-summary} &:& D_m,\; D_{\mathrm{cons}},\;
        \text{and practical } V_{\mathcal{H}}^{\mathrm{avg,\,rank}}
    \end{array}
    \;}
\]

\subsection{Terminology}
\label{app:three-quantities:terminology}

The following names are used verbatim throughout the paper and are
not interchangeable. The generic symbol $\Delta_{\mathcal{F}}$ is an
\emph{umbrella} for the two distinct diameters below (empirical
$\Delta^{\mathrm{emp}}_{\mathcal{F}}$ and oracle
$\Delta^{\mathrm{orc}}$); it is retained only in informal prose and
never appears in a formula or a non-implication, both of which name
the specific diameter they use.

\begin{center}
\begin{tabular}{ll}
\toprule
\textbf{Prose term} & \textbf{Formal object} \\
\midrule
invisible subspace & $\mathcal{N}_T(\mathcal{F}) = \ker \mathcal{R}_T$ \\
realized difficulty & $D_m$ \\
consensus difficulty & $D_{\mathrm{cons}}$ \emph{(only when aggregating)} \\
admissible ambiguity (umbrella) & $\Delta_{\mathcal{F}}$ \emph{(informal prose only)} \\
empirical admissible ambiguity & $\Delta^{\mathrm{emp}}_{\mathcal{F}}$ (diameter of $\mathcal{V}_{\mathcal{F}}(D_T;\varepsilon)$) \\
oracle local ambiguity & $\Delta^{\mathrm{orc}}$ (diameter of $\mathcal{V}_T(f_0)$) \\
finite-ensemble diameter & $\widehat{\Delta}_{\mathcal{H}}$ \\
practical average disagreement & $V_{\mathcal{H}}^{\mathrm{avg,\,value}}$ or $V_{\mathcal{H}}^{\mathrm{avg,\,rank}}$, as specified \\
\bottomrule
\end{tabular}
\end{center}

\emph{Takeaway.} Under missing compositional support, realized
difficulty, admissible ambiguity, and empirical disagreement are
distinct objects. The identity
\[
    R_{\mathrm{minimax}} \;=\; \tfrac{1}{4}\,(\Delta^{\mathrm{orc}})^2
\]
links structural ambiguity to minimax prediction risk.
Corollary~\ref{cor:blind-spot} shows formally why finite-model
disagreement can miss that ambiguity; the shared-blind-spot
construction (Appendix~\ref{app:synth}) further shows how such
agreement can coexist with realized prediction error.

\section{Invisible subspaces for concrete function classes}
\label{app:invisible-bases}

\emph{Question.} What does the abstract invisible subspace
$\mathcal{N}_T = \ker \mathcal{R}_T$ look like for concrete
scientific function classes, and why does the electrocatalysis
experiment have six (not seven) unresolved interaction
directions?

\emph{Role.} Section~\ref{sec:theory} characterizes
missing-support ambiguity abstractly through $\mathcal{N}_T$. This
appendix makes that object explicit for polynomial and
anchored-ANOVA classes. We begin with the six-component
degree-3 catalyst setting used in \S\ref{sec:null-space-design},
then derive the general simplex formula, and finally give the
anchored-ANOVA analogue.

\subsection{The catalyst example: seven ambient directions
become six on the simplex}
\label{app:invisible-bases:catalyst}

\begin{tcolorbox}[colback=black!3, colframe=black!45, boxrule=0.4pt,
    left=6pt, right=6pt, top=3pt, bottom=3pt, arc=1pt]
\small
\textbf{Intuition.} A degree-$\le 3$ interaction is invisible on $T_{AB}$
exactly when every monomial contains the factor $x_A x_B$; then it vanishes
whenever $A$ or $B$ is absent. The candidates are
$\{x_A x_B,\; x_A^2 x_B,\; x_A x_B^2,\; x_A x_B x_j\ (j \notin \{A,B\})\}$:
\emph{seven} directions in ambient $\mathbb{R}^6$, but the simplex constraint
$\sum_j x_j = 1$ removes one, leaving $r = 6$. The derivation below makes this
count exact.
\end{tcolorbox}

Take
\[
    d = 6, \qquad D = 3.
\]
Any polynomial interaction that is invisible whenever $A$ or $B$ is
absent must contain the factor $x_A x_B$. So in ambient
$\mathbb{R}^6$ the degree-$\le 3$ invisible monomials are
\[
    x_A x_B,
    \qquad
    x_A^2 x_B,\; x_A x_B^2,
    \qquad
    x_A x_B x_j \quad (j \notin \{A, B\}).
\]
There are four possible third elements $j \notin \{A, B\}$, so
\[
    1 + 2 + 4 \;=\; 7.
\]
If compositions were unconstrained vectors in $\mathbb{R}^6$, the
invisible space would indeed have dimension seven.

\paragraph{Impose the simplex constraint.} Real compositions
satisfy $\sum_{j=1}^6 x_j = 1$. Multiplying by $x_A x_B$,
\[
    x_A x_B
    \;=\; x_A x_B \sum_{j=1}^6 x_j
    \;=\; x_A^2 x_B + x_A x_B^2
    + \sum_{j \notin \{A, B\}} x_A x_B x_j.
\]
Hence the degree-$2$ monomial is not independent of the six
degree-$3$ generators once we restrict to the composition simplex,
and
\[
    \boxed{\;
    r_{\mathrm{ambient}} = 7
    \;\longrightarrow\;
    r_{\mathrm{simplex}} = 6.
    \;}
\]
The invisible-subspace basis used in \S\ref{sec:null-space-design}
can therefore be taken as
\[
    \phi_{AB}(x)
    \;=\;
    \bigl[
        x_A^2 x_B,\;
        x_A x_B^2,\;
        \{x_A x_B x_j\}_{j \notin \{A, B\}}
    \bigr]
    \;\in\; \mathbb{R}^6.
\]
The six unknown coefficients correspond to (i) nonlinear
$A$-loading modulation of the pair effect, (ii) nonlinear
$B$-loading modulation, and (iii) four third-element-conditioned
$A$-$B$ interactions, one per possible spectator. These six are
jointly linearly independent on $\Delta^5$. Write all six as
$\psi_j = x_A x_B \, x_j$ for $j = 1, \ldots, 6$ (with $x_A, x_B$
themselves among the $x_j$, so $\psi = x_A^2 x_B$ and $x_A x_B^2$
are included). Suppose $\sum_{j=1}^{6} c_j \psi_j = 0$ on
$\Delta^5$. Dividing by $x_A x_B > 0$ on the relatively open
copresence region gives $\sum_{j=1}^{6} c_j x_j = 0$ there. A
homogeneous linear form vanishing on a relatively open subset of
$\sum_j x_j = 1$ is identically zero, hence every $c_j = 0$. This
single relation rules out any mixed dependence among all six, not
only within the spectator or loading subsets.

\subsection{General polynomial invisible subspace (ambient)}
\label{app:invisible-bases:poly-ambient}

For degree-$\le D$ polynomials
$\mathcal{P}_D = \{p : \deg p \le D\}$ on $\mathbb{R}^d$, a monomial
$x^\alpha = \prod_j x_j^{\alpha_j}$ vanishes on
$T_{AB} = \{x \colon x_A x_B = 0\}$ if and only if
$\alpha_A \ge 1$ and $\alpha_B \ge 1$. Therefore
\begin{equation}
    \mathcal{N}_{T_{AB}}(\mathcal{P}_D)
    \;=\; \operatorname{span}\!\bigl\{\, x^\alpha \colon
    \alpha_A \ge 1,\; \alpha_B \ge 1,\; |\alpha| \le D \,\bigr\}.
    \label{eq:null-poly}
\end{equation}
Factoring out $x_A x_B$ gives $h(x) = x_A x_B\, q(x)$ with
$\deg q \le D - 2$, so, for $D \ge 2$ (and with the convention
$\mathcal{P}_k = \{0\}$ for $k < 0$, giving $r_{\mathrm{ambient}} = 0$
when $D < 2$), the ambient dimension is
\begin{equation}
    r_{\mathrm{ambient}}(D, d)
    \;=\; \binom{D - 2 + d}{d}.
    \label{eq:null-poly-dim}
\end{equation}
For $d = 6$: $r_{\mathrm{ambient}}(2, 6) = 1$
(only $x_A x_B$), $r_{\mathrm{ambient}}(3, 6) = 7$, and
$r_{\mathrm{ambient}}(4, 6) = 28$.

\subsection{Restriction to the simplex}
\label{app:invisible-bases:simplex}

On the composition simplex
$\Delta^{d-1} = \{x \ge 0 : \sum_j x_j = 1\}$, polynomials differing
by a multiple of $\ell(x) = \sum_j x_j - 1$ represent the same
function; write $L_D = (\ell) \cap \mathcal{P}_D$ for the
degree-$\le D$ multiples of $\ell$, and let
$\mathcal{N}^{\mathrm{amb}}_D = \mathcal{N}_{T_{AB}}(\mathcal{P}_D)$
denote the ambient pair-gated kernel. Write
$T_{AB}^\Delta = \Delta^{d-1} \cap (\{x_A = 0\} \cup \{x_B = 0\})$ for
the complete two-facet training support on the simplex. The invisible
space on the simplex is the restriction kernel
$\mathcal{N}^{\Delta}_D := \ker\!\bigl(\mathcal{R}_{T_{AB}^\Delta} :
\mathcal{P}_D/L_D \to \operatorname{Fun}(T_{AB}^\Delta)\bigr)$, and
restriction induces the \emph{quotient isomorphism} (not a literal
equality of function sets)
\[
    \mathcal{N}^{\Delta}_D
    \;=\; \frac{\mathcal{N}^{\mathrm{amb}}_D + L_D}{L_D}
    \;\cong\; \frac{\mathcal{N}^{\mathrm{amb}}_D}
    {\mathcal{N}^{\mathrm{amb}}_D \cap L_D}.
\]
Because $\mathcal{N}^{\mathrm{amb}}_D \cap L_D =
(\ell) \cap \mathcal{N}_{T_{AB}}(\mathcal{P}_D) =
\ell \cdot \mathcal{N}_{T_{AB}}(\mathcal{P}_{D-1})$, its dimension
is
\[
    \binom{D - 2 + d}{d} \;-\; \binom{D - 3 + d}{d},
\]
and Pascal's identity gives
\begin{equation}
    \boxed{\;
    r_{\mathrm{simplex}}(D, d)
    \;=\; \binom{D + d - 3}{d - 1}.
    \;}
    \label{eq:null-simplex-dim}
\end{equation}
For $D = 3, d = 6$ this returns $\binom{6}{5} = 6$, matching
\S\ref{app:invisible-bases:catalyst} directly. Lemma~\ref{lem:simplex-dim}
formalizes this quotient argument.

\subsection{Anchored-ANOVA analogue}
\label{app:invisible-bases:anova}

In an anchored ANOVA decomposition, each component
$f_S = \sum_{\alpha_S} c_{\alpha_S} \prod_{j \in S} \phi_j(x_j)^{\alpha_j}$
indexed by $S \subseteq \{1, \ldots, d\}$ (the sum runs over
multi-indices $\alpha_S$ with $\alpha_j \ge 1$ for every $j \in S$, so
the component genuinely depends on all of $S$) represents an interaction
involving exactly the variables in $S$, with each univariate factor
$\phi_j$ vanishing at its anchor $\phi_j(0) = 0$. A component
$f_S$ becomes invisible whenever $A$ or $B$ is absent if and only
if it contains \emph{both} $A$ and $B$: only then does setting
$x_A = 0$ or $x_B = 0$ zero the product. The same conclusion holds for
an arbitrary sum $f = \sum_S f_S$: if $f$ vanishes on the facet
$x_A = 0$, uniqueness of the anchored decomposition forces every
component with $A \notin S$ to vanish; applying the same argument on
$x_B = 0$ eliminates all remaining components with $B \notin S$. Hence
only terms with $S \supseteq \{A, B\}$ can survive. Consequently,
\emph{on a Cartesian
product domain $\mathcal{X} = \prod_j X_j$ with $0 \in X_j$, where the
anchored components form a unique direct sum},
\begin{equation}
    \boxed{\;
    \mathcal{N}_{T_{AB}}(\mathcal{F}_{\mathrm{ANOVA}})
    \;=\; \bigoplus_{S \supseteq \{A, B\}} \mathcal{F}_S.
    \;}
    \label{eq:null-anova}
\end{equation}
This identity is stated before restriction to the simplex. After
restriction the components need no longer form a direct sum (e.g.\ on
$\Delta^2$, $f_A = x_A - x_A^2$ and $f_{AC} = -x_A x_C$ are
non-$AB$ components whose sum $x_A(1 - x_A - x_C) = x_A x_B$ is
invisible), so a separate kernel analysis (as for the polynomial
class in Lemma~\ref{lem:simplex-dim}) is required on the simplex.
So missing $A + B$ leaves all interaction terms requiring both
$A$ and $B$ unidentified: $f_{AB}$, $f_{ABC}$, $f_{ABCD}$, and so
on. Under a generic centred ANOVA basis, whose non-anchored components
need not vanish when a participating coordinate is set to
zero, this characterization does not apply; interaction components
can leak into $T_{AB}$ through main effects.

\emph{Takeaway.} The invisible subspace is not an abstract artifact
of the theory. For concrete interaction classes it has an explicit
basis determined jointly by the model class and the observed
support. In the six-component degree-$3$ catalyst setting, on the
\emph{complete geometric two-facet} support, missing
$A + B$ leaves exactly six independent polynomial interaction
directions unresolved (on finite observed support these six
pair-gated functions form a canonical $r_{\mathrm{struct}} = 6$
structural subspace of the empirical kernel, not necessarily the whole
kernel): these are the coordinates targeted by the
measurement designs in \S\ref{sec:null-space-design}. The
anchored-ANOVA analogue shows the same principle in a non-polynomial
class. These bases characterize the oracle support-induced
component of ambiguity; finite-sample estimation uncertainty is a
separate enlargement of the version space, as defined in
Appendix~\ref{app:three-quantities}.

\subsection{General construction for an arbitrary feature representation}
\label{app:invisible-bases:general}

The analytic bases above exist only for the polynomial-simplex class, but the
invisible subspace has a purely computational definition for \emph{any} model
that supplies a feature map $\psi : \mathcal{X} \to \mathbb{R}^p$ (a neural
encoder, a graph model, a frozen MLIP, a sequence model). Stacking the features
on the observed support,
$\Psi_T = [\psi(x_1), \ldots, \psi(x_n)]^\top \in \mathbb{R}^{n \times p}$
($x_i \in T_{AB}$), the feature-space directions that vanish on $T_{AB}$ are
exactly $\mathcal{N}_T^\psi = \ker \Psi_T$, i.e.\ the right-singular vectors of
$\Psi_T$ with zero singular value. When the support is only \emph{approximately}
uninformative about a direction (finite, noisy, or off-manifold support, as in the
usual real-data case), one replaces the exact null space by the
\emph{small-singular-value subspace}: the trailing right-singular vectors of
$\Psi_T$, i.e.\ the directions the observed support constrains least. This gives
a general, representation-agnostic algorithm,
\[
\text{representation } \psi \;\to\; \Psi_T \;\to\; \text{(small-)singular-value subspace}
\;\to\; \Phi_C \;\to\; I\text{-optimal design},
\]
of which the polynomial-simplex basis is the analytic special case, and the
MACE-MP embedding-residual SVD (Appendix~\ref{sec:mp-design-appendix}) and the ESM
protein sketch (Appendix~\ref{sec:protein-design-appendix}) are concrete
instances. Proposition~\ref{prop:misspec} quantifies the cost of using such an
approximate subspace: the risk carries the irreducible term $\|h^\star_\perp\|^2$
measuring how much of the true invisible component $\Phi$ fails to span, the
quantity the real-data $\mathcal{N}_T$-overlap sweep (Appendix~\ref{app:main-tables})
and synthetic sweep (Appendix~\ref{app:synth-misspec}) probe empirically.

\section{Materials Project support counterfactual}
\label{app:mp-support}

\emph{Question and role.} The electrocatalysis experiment
(\S\ref{sec:real-setup}) asks whether \emph{adding} measurements
targeted to a support-invisible interaction space improves
prediction. Here we test the complementary intervention: when
$A+B$ labels \emph{are} available, does removing specifically that
joint support create a deficit that matched removal of other
training data does not? Because the representation is a frozen
MACE-MP encoder pretrained on Materials Project, this experiment
isolates the value of downstream property-label support
conditional on that representation; it is not an intervention on
representation pretraining. Together with \S\ref{sec:real-setup},
it probes missing support from both directions: add-then-fit
and delete-then-fit.

\emph{In one sentence.} For each element pair $(A,B)$ we train the same predictor
under five quantity-matched training-set removals and ask whether deleting
\emph{specifically} the $A{+}B$ copresence region causes a larger downstream loss
than deleting the same amount of other training data.

\subsection{Dataset and intervention}
\label{app:mp-support:setup}

We query the Materials Project API over a $21$-element alphabet
covering common cations (Li, Na, K, Mg, Ca, Ba, Fe, Mn, Co, Ni,
Cu, Zn, Ti, Al, Ga, Sn) and anions/metalloids (O, S, N, F, Cl),
enumerating all binary and ternary chemsys with $\leq 30$ sites
and $E_{\text{above hull}} \leq 0.2$ eV/atom (near-stable). This
yields $7467$ MP structures with cached DFT properties (formation
energy per atom, $E_{\text{above hull}}$, density, band gap) and
MACE-MP-medium invariant per-atom descriptors ($256$-dim
mean-pooled per structure, $L_2$-normalized). We keep the top-$30$
pairs by $|C_{AB}|$ subject to $|C_{AB}| \geq 40$ and $\geq 8$
single-element structures on each side. For each pair we
subsample $C_{AB}$ to a pool of size $\leq 300$, reserve a fixed
common evaluation pool $U_{AB} \subset C_{AB}$ ($50\%$ per seed),
and train ridge (or HGB) on the five conditions below; $15$ seeds
per pair for ridge, $8$ seeds for HGB.

Every condition trains on some subset of
$T_{AB} \cup (C_{AB} \setminus U_{AB})$ and is evaluated on the
same $U_{AB}$. Let $n := |C_{AB} \setminus U_{AB}|$. All three
matched-removal controls start from the \emph{full} condition and
delete $n$ samples from the training pool.
\begin{itemize}[leftmargin=1.6em,itemsep=1pt]
    \item \emph{full}: train on $T_{AB} \cup (C_{AB} \setminus U_{AB})$.
    \item \emph{zero}: train on $T_{AB}$ only (no joint labels).
    \item \emph{quantity-matched}: from \emph{full}, remove $n$
        uniformly-random $T$-side samples.
    \item \emph{covariate-nearest}: from \emph{full}, remove the
        $n$ $T$-side samples whose MACE-MP embeddings are closest
        to the mean $U_{AB}$-embedding.
    \item \emph{wrong-pair}: an \emph{eligible} pair is another
        pair in the top-$30$ set with
        $\{A', B'\} \cap \{A, B\} = \emptyset$ and
        $|C_{A'B'} \setminus C_{AB}| \ge n$; among eligibles, we
        select $(A', B')$ whose non-overlap size
        $|C_{A'B'} \setminus C_{AB}|$ is closest to $n$. From
        \emph{full}, remove $n$ samples drawn uniformly at random
        (per experiment seed) from
        $C_{A'B'} \setminus C_{AB}$, so target-pair joint labels
        are retained.
\end{itemize}

\subsection{Estimand and controls}
\label{app:mp-support:estimand}

For each pair, the primary estimand is the \emph{support gap}
\[
    \boxed{\; G_{AB} \;=\; \rho_{\text{full}} - \rho_{\text{zero}}, \quad
    \rho_{\text{cond}} \;=\; \rho_S(\hat y_{\text{cond}}, y)\big|_{U_{AB}}. \;}
\]
For each matched-removal control, define the analogue
\[
    \boxed{\; G^{\text{ctrl}}_{AB} \;=\; \rho_{\text{full}} - \rho_{\text{ctrl}}. \;}
\]
If joint support carries pair-specific information, $G_{AB} > 0$
and $G_{AB}$ exceeds $G^{\text{ctrl}}_{AB}$ under each matched
removal. Because both quantities subtract from the same
$\rho_{\text{full}}$,
\[
    G_{AB} - G^{\text{ctrl}}_{AB}
    \;=\; \rho_{\text{ctrl}} - \rho_{\text{zero}},
\]
so the primary paired contrast measures directly how much better
than \emph{zero-support} a matched-removal control is on $U_{AB}$
, i.e.\ whether removing all $A+B$ support hurts strictly more
than removing a matched amount or kind of other training data. The three controls target three distinct
confounds:
\begin{center}
\begin{tabular}{ll}
\toprule
Control & Alternative explanation probed \\
\midrule
quantity-matched & effect is due to smaller training size \\
covariate-nearest & effect is due to proximity to the target-region centroid in MACE-MP space \\
wrong-pair & effect is due to removing \emph{some} region, not $C_{AB}$ specifically \\
\bottomrule
\end{tabular}
\end{center}

\subsection{Primary support-intervention result}
\label{app:mp-support:primary}

Table~\ref{tab:app-mp-full} reports the five conditions on both
targets. On $E_{\text{above hull}}$ the paired contrast
$G_{AB} - G^{\text{ctrl}}_{AB}$ is positive on $87$--$90\%$ of
pairs under all three controls; band gap is directionally
consistent with smaller magnitude.

\begin{table}[!ht]
\centering\small
\caption{MP counterfactual support-ablation, full table
($30$ pairs, ridge on L2-normalized MACE-MP embeddings; $15$
seeds).}
\label{tab:app-mp-full}
\begin{tabular}{lcc}
\toprule
                                            & $E_{\text{above hull}}$ & Band gap \\
\midrule
$\rho_{\text{full}}$ (median)               & $+0.595$ & $+0.582$ \\
$\rho_{\text{zero}}$ (median)               & $+0.555$ & $+0.558$ \\
$\rho_{\text{match}}$ (random $T$)          & $+0.593$ & $+0.581$ \\
$\rho_{\text{match}}^{\text{near}}$ (nearest $T$ to $U_{AB}$) & $+0.598$ & $+0.574$ \\
$\rho_{\text{match}}^{\text{wrong}}$ (wrong-pair removal)     & $+0.592$ & $+0.580$ \\
\midrule
$G_{AB}$ median ($\%$ pairs $>0$)           & $+0.023$ ($90\%$) & $+0.006$ ($80\%$) \\
$G^{\text{match}}_{AB}$ (quantity)          & $+0.000$ ($50\%$) & $-0.000$ ($30\%$) \\
$G^{\text{match,near}}_{AB}$ (covariate)    & $+0.003$ ($63\%$) & $+0.001$ ($60\%$) \\
$G^{\text{match,wrong}}_{AB}$ (specificity) & $+0.000$ ($53\%$) & $-0.001$ ($27\%$) \\
Paired $G - G^{\text{match}}$ (quantity)    & $+0.023$ ($90\%$) & $+0.007$ ($80\%$) \\
Paired $G - G^{\text{match,near}}$ (covariate) & $+0.019$ ($90\%$) & $+0.005$ ($63\%$) \\
Paired $G - G^{\text{match,wrong}}$ (specificity) & $+0.021$ ($87\%$) & $+0.007$ ($80\%$) \\
\midrule
\multicolumn{3}{l}{\emph{$k$-recovery: all-pair median $\Delta\rho$ ($30$ pairs) $\;/\;$ subset $R(k)$ (defined below)}} \\
$k = 1$   &  $+0.0000 / +0.003$ & $+0.0000 / +0.000$ \\
$k = 2$   &  $+0.0001 / +0.005$ & $+0.0002 / +0.028$ \\
$k = 4$   &  $+0.0000 / +0.004$ & $+0.0002 / +0.057$ \\
$k = 8$   &  $+0.0013 / +0.051$ & $+0.0003 / +0.064$ \\
$k = 16$  &  $+0.0027 / +0.092$ & $+0.0002 / +0.153$ \\
$k = 32$  &  $+0.0057 / +0.258$ & $+0.0021 / +0.321$ \\
\bottomrule
\end{tabular}
\begin{flushleft}\footnotesize
Each $k$-recovery cell: all-pair ($30$-pair) median absolute
$\Delta\rho$ (primary summary) $/$ selected-subset relative recovery
$R(k) = \operatorname*{med}_{p:\,G_p \geq 0.02}
(\rho_{k,p} - \rho_{0,p})/G_p$ (median of the pairwise ratios, not a
ratio of medians).
$R(k)$ uses the pre-declared subset $G_{AB} \geq 0.02$ ($19/30$ pairs
on $E_{\text{above hull}}$; $9/30$ on band gap) and is
\emph{descriptive} (outcome-selected, noisy denominator).
\end{flushleft}
\end{table}

\subsection{Does restoring joint support repair the deficit?}
\label{app:mp-support:recovery}

Two related tests: adding joint labels to a zero-support base,
and swapping joint labels in for non-copresence labels at fixed
total size.

\paragraph{$k$-recovery (add joint labels).}
Table~\ref{tab:app-mp-full} reports the all-pair ($30$-pair) median
absolute gain
$\operatorname{med}_p(\rho_{k,p} - \rho_{0,p})$ alongside the
selected-subset relative recovery
$R(k) = \operatorname*{med}_{p:\,G_p \geq 0.02}
(\rho_{k,p} - \rho_{0,p})/G_p$ (the median is taken over the pairwise
ratios) on
the pre-declared subset $G_{AB} \geq 0.02$ (19/30 pairs on
$E_{\text{above hull}}$, 9/30 on band gap). The all-pair absolute
gain generally increases from $\approx 0$ at small $k$ to $+0.006$
on $E_{\text{above hull}}$ and $+0.002$ on band gap at $k = 32$, with
small nonmonotone steps at small $k$ (e.g.\ $k = 2 \to 4$ on
$E_{\text{above hull}}$, $0.0001 \to 0.0000$; $k = 8 \to 16$ on band
gap, $0.0003 \to 0.0002$); on
the broader set $G_{AB} > 0$ ($27/30$ and $24/30$ pairs) the
corresponding medians are $+0.006$ and $+0.003$, with relative
recovery $+0.281$ and $+0.298$. The selected-subset $R(k)$ reaches
$+0.258$ and $+0.321$ at $k = 32$ but, because the threshold is
applied to the same estimated $G_{AB}$ that appears in its
denominator, it is descriptive among empirically high-gap pairs; the
all-pair absolute $\Delta\rho$ is the safer summary and the estimand
we foreground.

\paragraph{Constant-training-size $k$-swap (relocate labels).}
Under a protocol that drops $k$ random $T$ samples to make room
for $k$ joint samples (so total training size is fixed at
$|T_{AB}|$), recovery is smaller but nonnegative and generally
increasing in $k$: absolute $\Delta \rho \in \{+0.000, +0.003, +0.006\}$ at
$k \in \{4, 16, 32\}$ on $E_{\text{above hull}}$ (and
$+0.000, +0.002, +0.002$ on band gap; the $k=4$ entry prints
$+0.000$, i.e.\ nonnegative rather than strictly positive).
Relocation still helps at the larger budgets
(the median point estimate is positive by $k = 16$, not merely an
artifact of increased training size), but the effect is muted, as expected when
useful $T$-side labels are given up.

\subsection{Does the effect depend on the learner?}
\label{app:mp-support:learner}

The support gap replicates under histogram gradient boosting
(HGB). On the matched $8$-seed subset used for the HGB comparison,
ridge $G_{AB}$-median is $+0.025$ ($93\%$ of pairs positive;
paired $G - G^{\text{match}}$ positive on $87\%$); HGB
$G_{AB}$-median is $+0.101$ ($97\%$; paired $100\%$). (The
headline ridge value $+0.023$ in
Table~\ref{tab:app-mp-full} uses $15$ seeds; the $+0.025$ here is
on the $8$-seed HGB-matched subset. Both statistics carry a
seed-level MAD of $\approx 0.02$--$0.03$.) Pair-level $G_{AB}$
correlates moderately across learners (Spearman $r = +0.46$,
Kendall's $W = 0.73$ on $n = 30$ pairs).

\paragraph{Per-pair decomposition of the ridge-to-HGB gap.} Split
the increment into
$\Delta^{\text{gain}}_p = \rho_{\text{full},p}^{\text{HGB}} -
\rho_{\text{full},p}^{\text{ridge}}$ and
$\Delta^{\text{extrap}}_p = \rho_{\text{zero},p}^{\text{ridge}} -
\rho_{\text{zero},p}^{\text{HGB}}$; by construction,
$G_{AB}^{\text{HGB}} - G_{AB}^{\text{ridge}} = \Delta^{\text{gain}}_p
+ \Delta^{\text{extrap}}_p$ exactly. Table~\ref{tab:app-hgb-decomp}
reports both terms.

\begin{table}[!ht]
\centering\small
\caption{HGB gap decomposition (per pair, median across $8$ seeds,
$30$ pairs).}
\label{tab:app-hgb-decomp}
\begin{tabular}{lcccc}
\toprule
Term & Median & IQR & Range & \% pairs $>0$ \\
\midrule
$\Delta^{\text{gain}}_p$
  & $+0.036$ & $[-0.013, +0.091]$ & $[-0.152, +0.287]$ & $70\%$ \\
$\Delta^{\text{extrap}}_p$
  & $+0.065$ & $[-0.020, +0.106]$ & $[-0.144, +0.303]$ & $67\%$ \\
$G_{AB}^{\text{HGB}} - G_{AB}^{\text{ridge}}$
  & $+0.066$ & $[+0.038, +0.145]$ & $[-0.012, +0.337]$ & $90\%$ \\
\bottomrule
\multicolumn{5}{l}{\footnotesize Seed-level MAD across $8$ seeds is $\approx 0.02$--$0.03$ for both learners under every condition;}\\
\multicolumn{5}{l}{\footnotesize $0/30$ pairs have $\rho_{\text{zero}} < 0$ under either learner; $1/30$ per learner ($2/30$ overall) has $\rho_{\text{zero}} < 0.3$.}
\end{tabular}
\end{table}

$11/30$ pairs have both $\Delta^{\text{gain}}_p > 0$ and
$\Delta^{\text{extrap}}_p > 0$ simultaneously; $12/30$ have
$\Delta^{\text{gain}}_p > \Delta^{\text{extrap}}_p$. Interpretation:
HGB both exploits supported joint data better than ridge (larger
$\rho_{\text{full}}$) and extrapolates worse when that support is
removed (smaller $\rho_{\text{zero}}$); the second term is the
larger contributor in the median.

\subsection{Dependence-aware robustness}
\label{app:mp-support:robustness}

The $30$ pair tasks share $17$ distinct elements, so pair rows are
not independent. Two dependence-aware checks address this.

\paragraph{Element-cluster bootstrap.}
Resample the $17$ elements with replacement ($1000$ replicates)
and weight each pair by the induced node/pigeonhole multiplicity
$\mathrm{freq}(A)\cdot\mathrm{freq}(B)$ (the same product-multiplicity
dyadic scheme used for electrocatalysis in
Tables~\ref{tab:stab-context} and~\ref{tab:zerolabel-cross}, rather
than a $\min(\cdot,\cdot)$ weighting), then take the weighted median
of each paired contrast. For $E_{\text{above hull}}$, the
$5$--$95\%$ CIs for $G_{AB} - G^{\text{ctrl}}_{AB}$ are strictly
above zero under all three controls: $[+0.014, +0.037]$ (quantity),
$[+0.008, +0.034]$ (covariate), $[+0.009, +0.040]$ (specificity).
For band gap the intervals are marginal, with lower bounds at
zero: $[+0.000, +0.026]$ (quantity),
$[+0.000, +0.021]$ (covariate), $[+0.000, +0.022]$ (specificity),
so the band-gap deficit is directional but not significant under
the dependence-aware bootstrap.

\paragraph{Leave-one-element-out (LOEO) sensitivity.}
For each of the $17$ elements, drop every pair containing it and
recompute the headline statistics on the remaining $15$--$29$
pairs. On $E_{\text{above hull}}$ every single-element deletion
leaves the main claim intact: $G_{AB}$ pair-median stays in
$[+0.019, +0.027]$ (baseline $+0.023$); $G_{AB}$-frac-positive in
$[80\%, 100\%]$ (baseline $90\%$); paired $G - G^{\text{ctrl}}$
frac-positive $\geq 80\%$ under all three controls in every fold.
The most stringent fold drops O (only $15$ pairs remain), where
$G_{AB}$-median is $+0.019$, frac-positive is $80\%$, and
$G - G^{\text{match,near}}$ median is $+0.012$ ($80\%$ pairs).
Band gap is more sensitive: pair-median $G_{AB}$ ranges
$[+0.001, +0.012]$, and $G - G^{\text{match,near}}$ has a
worst-fold median of $0.000$ ($53\%$ pairs; again the drop-O
fold).

\emph{Takeaway.} Removing labels specifically from the target
$A+B$ region produces a downstream prediction deficit that is not
reproduced by matched removal of the same number of training
examples, by removal of covariate-near examples, or by removal of
another compositional region. The effect persists across ridge
and HGB and under element-aware dependence checks, and
reintroducing joint labels partially repairs it. This establishes
a support-specific downstream effect under the frozen MACE-MP
representation. It does not establish that the true Materials
Project response lies in the polynomial invisible class of
\S\ref{sec:theory}, nor that MACE-MP itself lacks information
about the pair; rather, it is an external intervention showing
that relevant joint property labels carry information that
non-copresence labels do not substitute for. See
Appendix~\ref{sec:mp-design-appendix} for the accompanying MACE-MP
$\mathcal{N}_T$-$D$-opt design-selection experiment.
\clearpage

\section{Robustness, mechanism, and detail for the experimental-design results}
\label{app:main-tables}

This appendix collects the material omitted from \S\ref{sec:null-space-design}
for space, organized by the question each item answers:
\begin{itemize}[leftmargin=1.4em, itemsep=1pt, topsep=2pt]
\item \textbf{Robustness}: does the representation\,$\times$\,acquisition
factorial survive dependence-aware analysis? (provenance, per-system and
leave-one-element-out robustness, element-cluster bootstrap CIs).
\item \textbf{Mechanism}: is the $\mathcal{N}_T$ basis special relative to
random dimension-matched bases? (the random-6D null below; the
basis-misspecification sweep is in Appendix~\ref{app:synth-misspec}).
\item \textbf{Scaling}: does the predicted $\Theta(\sigma^2 r/m)$ rate appear
empirically? (the $r$-scaling validation of Proposition~\ref{prop:i-opt}).
\end{itemize}
The remaining tables give the full numbers behind the headline results (the
controlled synthetic design table, and the per-budget,
uncertainty-vs-identifiability, and pair-level electrocatalysis tables).

\paragraph{Status: offline counterfactual evaluation of prospective selection.}
The electrocatalysis study is not a live acquisition loop. All copresence labels
in $C_{AB}$ already exist; each acquisition rule is scored by the quality of the
subset it \emph{would} have queried, evaluated on a disjoint held-out pool
$U_{AB}$ it never observes. This is the standard offline (counterfactual)
protocol for comparing prospective experiment-selection rules on a fixed
dataset, a design analogue of off-policy evaluation. It isolates the selection
rule's value under a common evaluation target without conflating it with a fresh
data-collection budget; its limitation is that it re-uses already-measured
chemistries rather than probing genuinely unmeasured ones. The Materials Project
support-removal counterfactual (\S\ref{sec:mp-support-ablation}) and the
controlled synthetic sweeps address that complementary regime.

\paragraph{(i) Representation\,$\times$\,acquisition factorial: provenance and the
random-basis null.} The factorial of \S\ref{sec:real-setup}
(Table~\ref{tab:realdata-factorial}) uses one locked set of $U_{AB}$ splits,
seeds ($15$/pair), budgets, and background GP across all four cells and the
external baselines; every cell uses the same regularized Bayesian-linear model
($M_S = \lambda I + \Phi_S^\top\Phi_S$, $\lambda = 10^{-3}$), so pointwise
variance is $\phi(x)^\top M_S^{-1}\phi(x)$ (uncertainty sampling / $D$-opt) and
$I$-opt minimises $\mathrm{tr}(\Sigma_C M_S^{-1})$ on the same basis
(scripts \texttt{analysis\_realdata\_factorial.py},
\texttt{analysis\_realdata\_factorial\_robustness.py}). All three decomposed
effects survive per-system stratification and every leave-one-element-out fold,
with element-cluster bootstrap CIs excluding $1$ for the representation and
selection effects and spanning $1$ for the acquisition criterion. As a first
null for the representation claim, we compare the $\mathcal{N}_T$ basis against
$\ResultFacNullJ$ random dimension-matched ($r = 6$) bases drawn from the
ambient degree-$\leq 3$ non-$\mathcal{N}_T$ monomial library, running the
identical $I$-opt pipeline in each: at $m = 15$ the structural basis achieves
lower MSE than \emph{every} one of the $\ResultFacNullJ$ sampled random bases on
$\ResultFacNullFracStrong\%$ of pairs
(script \texttt{analysis\_realdata\_random\_null.py}).

\emph{Full-ambient control} (reviewer red-team; script
\texttt{analysis\_realdata\_fullspace\_null.py},
data \texttt{results/realdata\_fullspace\_null.csv}). The baselines above exclude
$\mathcal{N}_T$ directions by construction, which could bake the answer into the
comparison. We therefore draw $\ResultFacFullNullJ$ random $6$-dim bases from the
\emph{full} degree-$\leq 3$ ambient space and run the identical $I$-opt
pipeline. This ambient space \emph{can} contain interaction structure and in
fact partially overlaps $\mathcal{N}_T$ (median subspace overlap
$\ResultFacFullNullOverlapMed$, mean squared cosine of principal angles). Even so, the $\mathcal{N}_T$ basis attains lower
MSE than the median full-space random basis on $\ResultFacFullNullBeatPct\%$ of
pairs at both $m = 15$ and $m = 30$, so the advantage is specific to the
$\mathcal{N}_T$ geometry, not merely to having interaction features. Across the
random full-space bases, greater $\mathcal{N}_T$-overlap predicts lower
within-pair-normalized MSE (Spearman $\ResultFacFullNullRhoMFifteen$ at $m = 15$,
$p = 0.01$; $\ResultFacFullNullRhoMthirty$ at $m = 30$, $p = 10^{-7}$), the
real-data analogue of the synthetic overlap sweep of
Appendix~\ref{app:synth-misspec}, though the effect is modest over the limited
overlap range ($0.33$--$0.60$) that random full-space bases reach; the
high-overlap regime is attained only by $\mathcal{N}_T$ itself.

\begin{table}[!ht]
\centering\small
\caption{\textbf{Representation\,$\times$\,acquisition factorial on
electrocatalysis} (60 pairs $\times$ 15 seeds, locked splits; same regularized
Bayesian-linear model and $\mathcal{N}_T$/own-basis readout in all cells;
the three effects are plotted in Figure~\ref{fig:learning-curve}).
\emph{Top:} median per-pair held-out Spearman for the four cells at $m = 15$
(external Random / GP-variance baselines for context, $\mathcal{N}_T$ readout).
\emph{Bottom:} the three decomposed effects as median per-pair MSE ratios
(lower than $1$ favors the first term), with element-cluster $95\%$ CIs.}
\label{tab:realdata-factorial}
\begin{tabular}{lcc}
\toprule
\multicolumn{3}{l}{\emph{Cell Spearman at $m=15$ (higher better)}} \\
 & pointwise variance & integrated $I$-opt \\
\midrule
$\mathcal{N}_T$ representation & $\ResultFacNTvarSp$ & $\ResultFacNTioptSp$ \\
Generic-6D representation      & $\ResultFacGenvarSp$ & $\ResultFacGenioptSp$ \\
\midrule
\multicolumn{3}{l}{\emph{External baselines ($\mathcal{N}_T$ readout): Random $0.814$, GP-variance $0.816$}} \\
\bottomrule
\end{tabular}

\vspace{0.6em}
\begin{tabular}{lccc}
\toprule
Effect (MSE ratio, median per pair) & $m=15$ & $95\%$ CI & $m=30$ \\
\midrule
Representation ($\mathcal{N}_T$ / Generic) & $\ResultFacRepRatio$ ($\ResultFacRepWin\%$) & $[\ResultFacRepCIlo, \ResultFacRepCIhi]$ & $\ResultFacRepRatioMthirty$ \\
Selection (targeted / random)              & $\ResultFacSelRatio$ ($\ResultFacSelWin\%$) & $[\ResultFacSelCIlo, \ResultFacSelCIhi]$ & $\ResultFacSelRatioMthirty$ \\
Acquisition ($I$-opt / same-basis variance) & $\ResultFacAcqRatio$ & $[\ResultFacAcqCIlo, \ResultFacAcqCIhi]$ & $\ResultFacAcqRatioMthirty$ \\
\bottomrule
\end{tabular}
\end{table}

\begin{figure}[!ht]\centering
\includegraphics[width=0.95\linewidth]{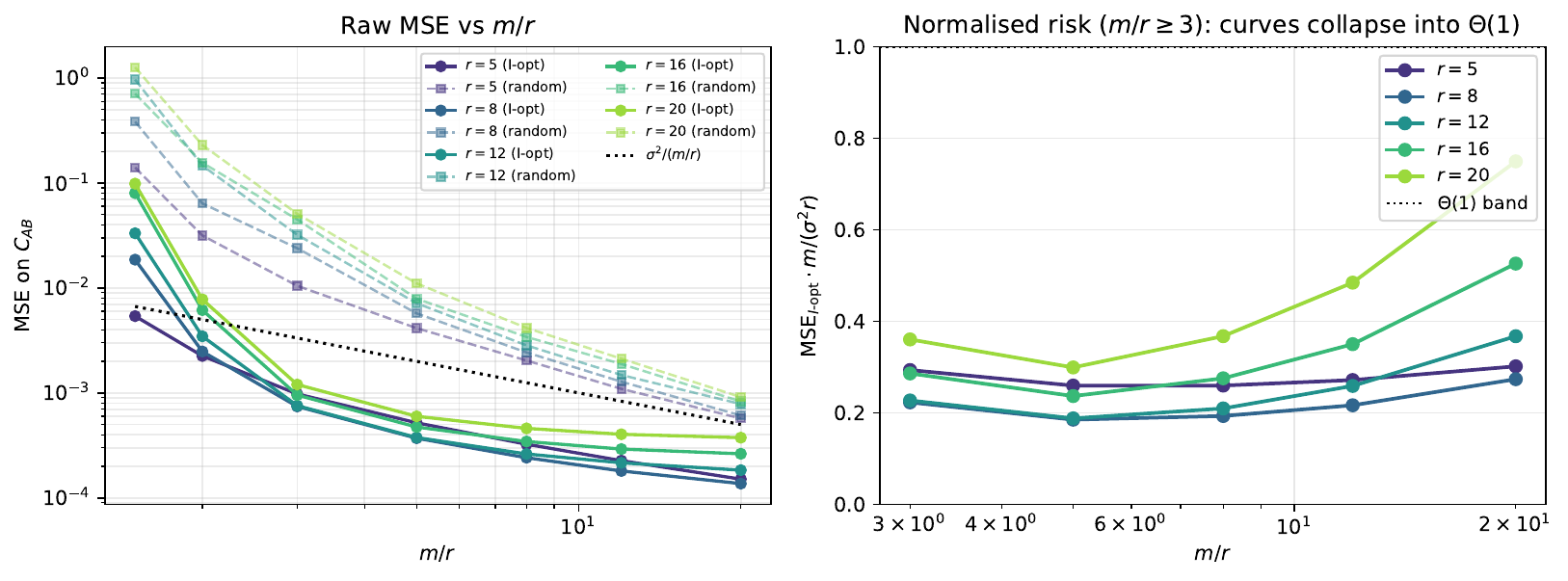}
\caption{\textbf{$\Theta(r/m)$ scaling of the $I$-optimal risk
(Proposition~\ref{prop:i-opt}).} Left: raw MSE (log--log) for the
$I$-optimal design (circles) vs random (squares) at $r \in \{5, 8,
12, 16, 20\}$; the dotted line marks $\sigma^2/(m/r)$. Right:
$\mathrm{MSE}_{I\text{-opt}} \cdot m / (\sigma^2 r)$ collapses across
$r$ into a bounded band above $m/r \approx 3$, matching the
theorem's prediction.}
\label{fig:r-scaling}
\end{figure}

\paragraph{$r$-scaling validation of Proposition~\ref{prop:i-opt}.}
Figure~\ref{fig:r-scaling} sweeps the invisible dimension $r$
through $d \in \{5, 8, 12, 16, 20\}$ at $D = 3$ (giving
$r = d$ by Lemma~\ref{lem:simplex-dim}) on the simplex-polynomial
synthetic and plots the
$I$-optimal MSE against $m/r$. The right panel normalises by
$m/(\sigma^2 r)$: for $m/r \geq 3$ all curves fall in the band
$[0.19, 0.75]$, consistent with
$\mathrm{MSE}_{I\text{-opt}} \propto \sigma^2 r/m$ across a factor of
$4$ in $r$. Ratios of random to $I$-optimal MSE reach $10$--$40\times$
at moderate budgets. This is consistent with the predicted scaling
over this finite range of $r$ and $m$; it is not a proof of
$\Theta$-asymptotics.

\paragraph{Synthetic simplex-aware D/I-optimal design (from \S\ref{sec:null-space-design}).}
\begin{table}[!ht]
\centering\small
\caption{Simplex-aware median OLS prediction MSE on $C$ under four
design strategies vs measurement budget $m$
($r_{\mathrm{simplex}} = \ResultNSDesignRsimplex$). Greedy
$D$-optimal and $I$-optimal each outperform random sampling from
$P_C$ across the just-past-invisible-dimension regime; the gap
narrows over the tested range as $m$ grows (random and optimal
designs can share a $1/m$ rate with different leading constants, so
this is a finite-range narrowing, not asymptotic equality). Table
entries are marginal medians across $80$ seeds. The headline
$3.5$--$4.1\times$ improvement is the fold-improvement of these
marginal medians ($\operatorname{med}_s \mathrm{MSE}_{\mathrm{random}}
/ \operatorname{med}_s \mathrm{MSE}_{\mathrm{design}}$; e.g.\
$3.95\times$ against $D$-opt at $m = 9$ and $4.14\times$ at $m = 16$).
The \emph{paired} per-seed median ratio
$\operatorname{med}_s(\mathrm{MSE}_{\mathrm{random},s}/
\mathrm{MSE}_{\mathrm{design},s})$ is a distinct but comparable
estimand ($2.35, 4.12, 4.49, 4.04, 3.40, 2.97\times$ against $D$-opt
at $m = 7, 9, 11, 16, 26, 46$). Maximin in input space does not
consistently help under simplex geometry; its input-space distance is
only weakly aligned with information in the null-space monomial
features.}
\label{tab:null-design}
\begin{tabular}{lcccc}
\toprule
$m$ (budget) & Random & Maximin & $D$-optimal & $I$-optimal \\
\midrule
$r + 1 = 7$     & $0.039$  & $0.022$ & $\mathbf{0.018}$ & $0.022$ \\
$r + 3 = 9$     & $0.023$  & $0.017$ & $\mathbf{0.006}$ & $0.006$ \\
$r + 5 = 11$    & $0.010$  & $0.012$ & $\mathbf{0.003}$ & $0.003$ \\
$r + 10 = 16$   & $0.006$  & $0.006$ & $\mathbf{0.002}$ & $0.002$ \\
$r + 20 = 26$   & $0.003$  & $0.003$ & $\mathbf{0.001}$ & $0.001$ \\
$r + 40 = 46$   & $0.001$  & $0.001$ & $\mathbf{0.0004}$ & $0.0006$ \\
\bottomrule
\end{tabular}
\end{table}

\paragraph{Per-budget Spearman on electrocatalysis (from
\S\ref{sec:real-setup}).}
\begin{table}[!ht]
\centering\small
\caption{Real-data null-space-aware design under the fixed-test
protocol. \emph{All rows here and in
Table~\ref{tab:uncertainty-vs-ident} come from one locked run}
(identical seeds, GP background, and $U_{AB}$ splits), so the Random
and GP columns match across the two tables. Rows are median per-pair
Spearman on the common held-out $U_{AB}$ (median across $15$ seeds
per pair, then across $60$ pairs); higher is better. Bold marks the
per-column maximum: on Spearman rank GP posterior variance leads at
small budgets ($m \le 12$) and $\mathcal{N}_T$-$D$-opt leads at
$m \ge 15$. Theorem-aligned $\mathcal{N}_T$-$I$-opt
(Prop.~\ref{prop:i-opt}) is reported in
Table~\ref{tab:uncertainty-vs-ident}, where it is tied with
$\mathcal{N}_T$-$D$-opt on Spearman but strictly better on MSE. The
``PCA-6 (matched eval)'' row scores the PCA-6 selection on its own
basis (rather than the $\mathcal{N}_T$ basis) and is the weakest
design, confirming the gain is not a generic dimension-matching
effect.}
\label{tab:null-design-real}
\begin{tabular}{lcccccccc}
\toprule
Design $\backslash m$
  & $8$ & $10$ & $12$ & $15$ & $18$ & $20$ & $25$ & $30$ \\
\midrule
Random                     & $0.800$ & $0.798$ & $0.808$ & $0.816$
                           & $0.814$ & $0.819$ & $0.820$ & $0.820$ \\
GP posterior variance      & $\mathbf{0.810}$ & $\mathbf{0.822}$ & $\mathbf{0.826}$ & $0.824$
                           & $0.827$ & $0.831$ & $0.832$ & $0.834$ \\
PCA-6 ($X_T$-fit, $\mathcal{N}_T$ eval)
                           & $0.798$ & $0.801$ & $0.802$ & $0.805$
                           & $0.806$ & $0.805$ & $0.816$ & $0.809$ \\
PCA-6 (matched eval)       & $0.728$ & $0.770$ & $0.792$ & $0.797$
                           & $0.800$ & $0.799$ & $0.804$ & $0.805$ \\
$D$-opt ($\mathcal{N}_T$)  & $0.807$ & $\mathbf{0.822}$ & $0.825$ & $\mathbf{0.830}$
                           & $\mathbf{0.829}$ & $\mathbf{0.833}$ & $\mathbf{0.834}$ & $\mathbf{0.838}$ \\
\bottomrule
\end{tabular}
\vspace{2pt}\\
\footnotesize Budgets start at $m = 8 > r_{\mathrm{simplex}}$;
$m \geq r = 6$ is necessary but not sufficient: identification
additionally requires the selected design to have full column rank,
$\mathrm{rank}(\Phi_S) = 6$, so that $\det(\Phi^\top \Phi) > 0$.
``$\mathcal{N}_T$ eval'' fits the residual on $\phi_{\mathcal{N}_T}$;
``matched eval'' fits it on the same PCA-6 basis used for selection.
PCA-6 is the support-informed generic 6-dim baseline: top-$6$ PCA
directions of the ambient non-$\mathcal{N}_T$ degree-$\leq 3$
polynomial features on $X_T$ (mean-centered on $X_T$; no copresence
labels). All methods share the candidate pool
$C_{AB} \setminus U_{AB}$ and per-(seed, budget) seed; the greedy
feature-space policies ($\mathcal{N}_T$-$D$-opt, PCA-6, and the
$\mathcal{N}_T$-$I$-opt of Table~\ref{tab:uncertainty-vs-ident})
share the same random $r$-point init, whereas Random and GP do not.
$\mathcal{N}_T$-$I$-opt is tied with $\mathcal{N}_T$-$D$-opt on
Spearman ($\Delta \in [-0.002, +0.001]$) but strictly better on
paired MSE and on $\mathrm{tr}(\Sigma_C M_S^{-1})$
(Table~\ref{tab:uncertainty-vs-ident}).
\end{table}

\paragraph{Uncertainty sampling vs identifiability-directed design (real data).}
Table~\ref{tab:uncertainty-vs-ident} reports the head-to-head between
GP posterior-variance sampling (model-uncertainty), $\mathcal{N}_T$-$I$-opt
(identifiability-directed), and random on all $60$ electrocatalysis
pairs under identical seeds, GP background, and $U_{AB}$ splits.
Reading paired median-of-per-pair-medians (design-vs-GP-variance on
the same seed and pair), $I$-opt reduces MSE by up to
$\sim 7\%$ (about $5$--$7\%$ for $m \geq 12$, smaller at
$m \leq 10$), improving MSE on $\ResultIoptGPvarMSEFrac\%$ of pairs
at $m = 15$, and is modestly better than GP-variance on
Spearman rank ($\ResultIoptGPvarSpearmanFrac\%$ of pairs, median
$\Delta = \ResultIoptGPvarSpearmanDelta$; the rank gap is small and
budget-dependent). On the theorem-aligned
quantities it dominates: it reduces $\mathrm{tr}(\Sigma_C M_S^{-1})$
by a paired factor of $1.5$--$2.2\times$ and increases
$\lambda_{\min}(\Phi_S^\top \Phi_S)$ by a paired factor of
$2.1$--$3.4\times$ ($M_S := \Phi_S^\top \Phi_S$) on every pair.
Median Jaccard overlap
$J_m(\mathrm{GP\text{-}var}, \mathcal{N}_T$-$I$-opt$)$ is
$0.05$--$0.20$: the two policies have low overlap (at $J=0.20$, equal-size sets still share $\approx 1/3$ of points).
For finite-pool selection, $\Sigma_C$ is replaced by the empirical
covariance $\widehat{\Sigma}_C = |C_{AB}|^{-1}\sum_{x \in C_{AB}}
\phi(x)\phi(x)^\top$ over the full copresence region (unlabeled
covariates only; no $y$-values on $U_{AB}$ are ever used). This is
transductive but strictly label-free and matches the reference
distribution in Prop.~\ref{prop:i-opt}. A sensitivity check
re-estimating $\widehat{\Sigma}_C$ on the selection pool only
($C_{AB}\setminus U_{AB}$, i.e.\ with $U_{AB}$ covariates completely
hidden) leaves the paired MSE essentially unchanged
(median-of-pair-medians ratio $0.996$--$1.020$ across
$m \in [8, 30]$; $46\%$--$54\%$ of pairs marginally lower with the
selection-pool covariance), so the result does not rely on
observing $U_{AB}$ covariates.

\begin{table}[t]
\centering\small
\caption{Real electrocatalysis: median per-pair (across $15$ seeds
then across $60$ pairs) under identical seeds, GP background, and
common held-out $U_{AB}$. Lower is better for MSE and
$\mathrm{tr}(\Sigma_C M_S^{-1})$; higher is better for Spearman and
$\lambda_{\min}$. $J_m$ is the median-across-pairs Jaccard between
GP-variance and $\mathcal{N}_T$-$I$-opt selection sets.}
\label{tab:uncertainty-vs-ident}
\begin{tabular}{lccccccc}
\toprule
$m$ & Design & Spearman & MSE & $\mathrm{tr}(\Sigma_C M_S^{-1})$
    & $\lambda_{\min}(\Phi_S^\top \Phi_S)$
    & $J_m$ \\
\midrule
$10$ & random         & $0.798$ & $5.10 \cdot 10^{-4}$ & $2.606$ & $7.3 \cdot 10^{-5}$ & \multirow{3}{*}{$0.05$} \\
     & GP variance    & $0.822$ & $4.96 \cdot 10^{-4}$ & $0.995$ & $1.9 \cdot 10^{-4}$ & \\
     & $I$-opt $\mathcal{N}_T$ & $\mathbf{0.824}$ & $\mathbf{4.82 \cdot 10^{-4}}$ & $\mathbf{0.498}$ & $\mathbf{4.6 \cdot 10^{-4}}$ & \\
\midrule
$15$ & random         & $0.816$ & $4.94 \cdot 10^{-4}$ & $0.891$ & $2.5 \cdot 10^{-4}$ & \multirow{3}{*}{$0.07$} \\
     & GP variance    & $0.824$ & $4.80 \cdot 10^{-4}$ & $0.417$ & $4.5 \cdot 10^{-4}$ & \\
     & $I$-opt $\mathcal{N}_T$ & $\mathbf{0.830}$ & $\mathbf{4.50 \cdot 10^{-4}}$ & $\mathbf{0.210}$ & $\mathbf{1.5 \cdot 10^{-3}}$ & \\
\midrule
$20$ & random         & $0.819$ & $4.70 \cdot 10^{-4}$ & $0.522$ & $4.6 \cdot 10^{-4}$ & \multirow{3}{*}{$0.11$} \\
     & GP variance    & $0.831$ & $4.68 \cdot 10^{-4}$ & $0.247$ & $8.2 \cdot 10^{-4}$ & \\
     & $I$-opt $\mathcal{N}_T$ & $0.831$ & $\mathbf{4.35 \cdot 10^{-4}}$ & $\mathbf{0.141}$ & $\mathbf{2.2 \cdot 10^{-3}}$ & \\
\midrule
$30$ & random         & $0.820$ & $4.52 \cdot 10^{-4}$ & $0.277$ & $9.4 \cdot 10^{-4}$ & \multirow{3}{*}{$0.20$} \\
     & GP variance    & $0.834$ & $4.46 \cdot 10^{-4}$ & $0.135$ & $1.6 \cdot 10^{-3}$ & \\
     & $I$-opt $\mathcal{N}_T$ & $\mathbf{0.838}$ & $\mathbf{4.26 \cdot 10^{-4}}$ & $\mathbf{0.091}$ & $\mathbf{3.4 \cdot 10^{-3}}$ & \\
\bottomrule
\end{tabular}
\vspace{2pt}\\
\footnotesize The theorem-aligned column
$\mathrm{tr}(\widehat\Sigma_C M_S^{-1})$ is the plug-in, model-based
I-optimal criterion of Prop.~\ref{prop:i-opt} (its exact minimax
risk up to $\sigma^2$ under the theorem's model; on real data $g$ is
estimated and $\phi$ is a structural prior, so this is the
model-based criterion, not the exact minimax risk of the real-data
problem). Entries are marginal medians
(median over seeds within a pair, then over the $60$ pairs). The
paired comparison (design vs GP-variance on the identical seed and
pair) tells a sharper story: at $m = 15$ the paired
$I/\mathrm{GPvar}$ median ratios are $\ResultIoptGPvarMSERatio$ (MSE),
$1/1.98$ ($\mathrm{tr}$), $3.29$ ($\lambda_{\min}$), with $I$-opt
improving $\ResultIoptGPvarMSEFrac\%$ / $100\%$ / $100\%$ of pairs
respectively; on \emph{Spearman rank} $I$-opt is modestly ahead
(winning $\ResultIoptGPvarSpearmanFrac\%$ of pairs, median
$\Delta = \ResultIoptGPvarSpearmanDelta$), so the design's benefit is
concentrated in
prediction error, consistent with the $L^2(P_C)$ (MSE) risk that
Prop.~\ref{prop:i-opt} governs. $J_m$: median-across-pairs Jaccard
between GP-variance and $I$-opt selection sets; observed
$0.05$--$0.20$ shows small overlap.
\end{table}

\begin{figure}[t]\centering
\includegraphics[width=0.95\linewidth]{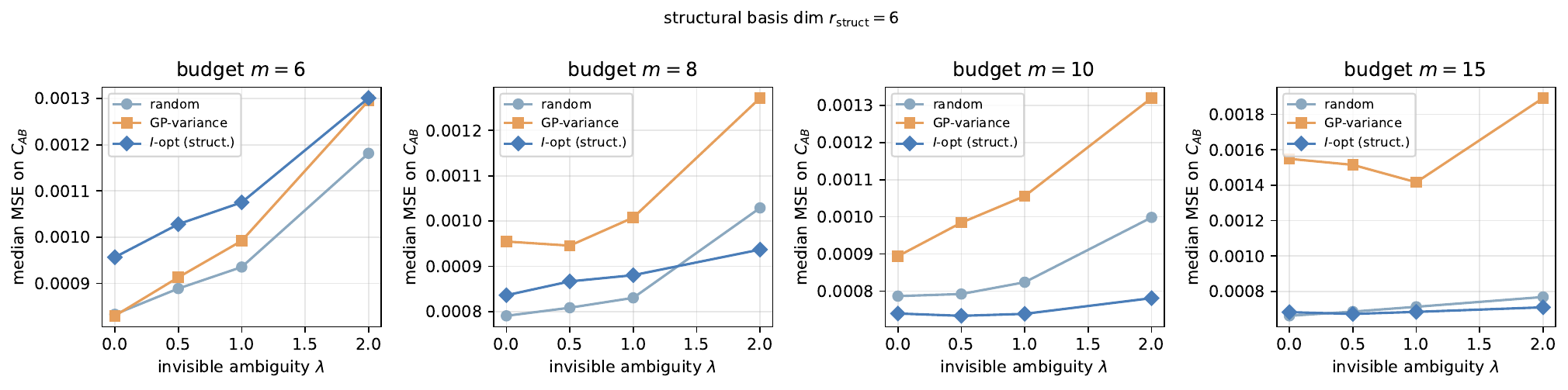}
\caption{\textbf{Synthetic decoupling: uncertainty vs identifiability.}
Simplex $d = 6$; we use a six-dimensional pair-gated \emph{structural}
basis ($r_{\mathrm{struct}} = 6$) made invisible by the $x_A = 0$
training support. This chosen basis need not span the full
degree-$\le 3$ polynomial restriction kernel of that single face
(which has dimension $\binom{7}{5} = 21$, larger than Lemma~7's
two-facet $r = 6$); $I$-optimality is computed on this chosen basis.
GP-variance is observed to increase with
$x_A$ for this kernel/design. Median MSE
on $C_{AB}$ across $20$ seeds vs invisible amplitude $\lambda$, at
four budgets $m$. $I$-opt on the structural basis (blue) is nearly flat in
$\lambda$; GP-variance (orange) degrades sharply because it stacks
selections into a rank-deficient corner of $C_{AB}$; random (grey)
sits between. Median Jaccard $J_m(\mathrm{GP\text{-}var}, I\text{-opt})
\leq 0.03$ at every $m$.}
\label{fig:synth-decoupling}
\end{figure}

\paragraph{Synthetic decoupling of $g$-uncertainty from
$\lambda$-ambiguity.} Figure~\ref{fig:synth-decoupling} isolates the
mechanism on a controlled synthetic where the T-support lies entirely
on the $A$-face ($x_A = 0$); for the kernel and design used here,
GP-variance is observed to increase with $x_A$ (an empirical property
of this simulation, not a consequence of the support geometry alone),
while the invisible signal
$h = \phi_{\mathcal{N}_T}^\top \theta$ (with $\theta$ drawn on the
unit sphere) has interior support on $C_{AB}$. The invisible amplitude
$\lambda$ is varied independently from the T-noise $\sigma_g$. At
tight budgets ($m \in \{8, 10, 15\}$ with $r = 6$), GP-variance
sampling selects a highly redundant set clustered at max-$x_A$; its
MSE is $2$--$3\times$ worse than $I$-opt and even underperforms random
at $m = 15$. Median Jaccard overlap
$J_m(\mathrm{GP\text{-}var}, I\text{-opt})$ is $0.00$--$0.03$ across
$m$: the two policies have very small median overlap. Crucially, the
$I$-opt advantage grows with $\lambda$ but is non-zero even at
$\lambda = 0$: GP-variance's degeneracy is a basis-conditioning failure
independent of the invisible amplitude.

\paragraph{Pair-level effects at $m = 15$ on electrocatalysis (from \S\ref{sec:real-setup}).}
\begin{table}[t]
\centering\small
\caption{Pair-level effects at the canonical budget $m = 15$: median
per-pair $\Delta$Spearman (design minus baseline) and fraction of
$60$ pairs with a positive effect; ranges are leave-one-element-out
sensitivity across the $24$ (system, element) drops. Per-system
fraction improved (out of $15$ pairs per system) shown in the last
column. All comparisons are pair-aggregated, not seed-level.}
\label{tab:null-design-effects}
\begin{tabular}{ll cc l}
\toprule
Contrast & Baseline & Median $\Delta$ & Pairs improved & Per-system \\
\midrule
$D$-opt ($\mathcal{N}_T$) $\;$vs.\;
  & random  & $+\ResultNSVThreeDoptNTvsRandomDelta$
  & $\ResultNSVThreeDoptNTvsRandomFrac\%$
  & A=$13$, B=$14$, C=$12$, D=$14$ \\
$D$-opt ($\mathcal{N}_T$) $\;$vs.\;
  & GP variance & $+\ResultNSVThreeNTvsGPvarDeltaFifteen$
  & $\ResultNSVThreeNTvsGPvarFracFifteen\%$
  & A=$\ResultNSVThreeNTvsGPvarSysA$, B=$\ResultNSVThreeNTvsGPvarSysB$,
    C=$\ResultNSVThreeNTvsGPvarSysC$, D=$\ResultNSVThreeNTvsGPvarSysD$ \\
$D$-opt ($\mathcal{N}_T$) $\;$vs.\;
  & PCA-6 ($X_T$-fit, $\mathcal{N}_T$ eval)
  & $+\ResultNSVFiveNTvsPCASixNTevalDelta$
  & $\ResultNSVFiveNTvsPCASixNTevalFrac\%$
  & A=$15$, B=$13$, C=$11$, D=$14$ \\
$D$-opt ($\mathcal{N}_T$) $\;$vs.\;
  & PCA-6 (matched eval)
  & $+\ResultNSVFiveNTvsPCASixMatchedDelta$
  & $\ResultNSVFiveNTvsPCASixMatchedFrac\%$
  & A=$15$, B=$15$, C=$15$, D=$14$ \\
\midrule
Nested-LOSO hybrid $\;$vs.\;
  & GP variance
  & $+\ResultNSVFiveHybridvsGPvarDelta$ & $\ResultNSVFiveHybridvsGPvarFrac\%$
  & A=$12$, B=$9$, C=$9$, D=$11$ \\
Nested-LOSO hybrid $\;$vs.\;
  & random & $+\ResultNSVFiveHybridvsRandomDelta$ & $\ResultNSVFiveHybridvsRandomFrac\%$
  & A=$13$, B=$11$, C=$12$, D=$13$ \\
Nested-LOSO hybrid $\;$vs.\;
  & $\mathcal{N}_T$-$D$-opt & $\ResultNSVFiveHybridvsNTDelta$
  & $\ResultNSVFiveHybridvsNTFrac\%$
  & A=$7$, B=$7$, C=$7$, D=$4$ \\
\bottomrule
\end{tabular}
\vspace{2pt}\\
\footnotesize All rows are computed from the single locked unified
run (same object as Tables~\ref{tab:null-design-real}--\ref{tab:uncertainty-vs-ident}
and Figure~\ref{fig:learning-curve}). LOEO sensitivity across the
$24$ (system, element) drops, all at $m = 15$; ranges are
descriptive deletion sensitivity, not confidence intervals.
$\mathcal{N}_T$ vs random $\Delta \in [+0.009, +0.011]$
(frac $\in [87\%, 91\%]$);
$\mathcal{N}_T$ vs GP variance $\Delta \in
[+\ResultNSVThreeLOEODeltaNTGPvarLo, +\ResultNSVThreeLOEODeltaNTGPvarHi]$
(frac $\in [\ResultNSVThreeLOEOFracNTGPvarLo\%, \ResultNSVThreeLOEOFracNTGPvarHi\%]$);
$\mathcal{N}_T$ vs PCA-6 $\Delta \in [+0.014, +0.016]$
(frac $\in [87\%, 91\%]$);
$\mathcal{N}_T$ vs PCA-6 matched-eval $\Delta \in [+0.024, +0.029]$
(frac $\in [98\%, 100\%]$);
nested-LOSO hybrid vs GP variance $\Delta \in [+0.002, +0.004]$
(frac $\in [65\%, 71\%]$); nested-LOSO hybrid vs $\mathcal{N}_T$-$D$-opt
is near-null across drops ($|\Delta| \leq 0.001$, frac in $[38\%, 45\%]$).
\end{table}

\let\oldappnullspaceredirect\relax
\phantomsection\label{app:nullspace-classes}

\FloatBarrier
\clearpage

\section{Proofs}
\label{app:proofs}

\emph{Purpose.} We give complete proofs of the theoretical results
in the order they appear in the main text. Each proof first states
the key argument and then supplies the minimal formal steps needed
to establish the claim.

\paragraph{Proof of Proposition~\ref{prop:missing-support} (from \S\ref{sec:theory}).}
\begin{proof}
Under $f_{+\lambda}$ and $f_{-\lambda}$ the marginal distribution of every
$(x_i, y_i)$ with $x_i \in T_{AB}$ coincides, because $h(x_i) = 0$. Hence the
joint distribution of $(D_T,\, \widehat{f}(D_T))$ is identical under the two
worlds. For any $\widehat{f}$ and $x \in C_{AB}$, the parallelogram identity
gives
\begin{align*}
\tfrac{1}{2}\!\left[(\widehat{f}(x) - f_{+\lambda}(x))^2 + (\widehat{f}(x) - f_{-\lambda}(x))^2\right]
= (\widehat{f}(x) - g(x))^2 + \lambda^2 h(x)^2
\geq \lambda^2 h(x)^2.
\end{align*}
Integrating against $P_C$ and using $\sup \geq \tfrac{1}{2}(\cdot + \cdot)$
yields \eqref{eq:prop1}. Gaussian noise plays no role here: the argument
uses only that the two worlds induce the same observation law on $T_{AB}$
(so $(D_T, \widehat{f}(D_T))$ has identical distribution under both), and
hence the bound holds for any noise model with that property.
\end{proof}

\paragraph{Proof of Corollary~\ref{cor:blind-spot} (from \S\ref{sec:theory}).}
\begin{proof}
\emph{Idea.} By the endpoint premise, $\mathcal{V}_{\mathcal{F}}$
contains both endpoints $g \pm \Lambda h$; a family $\mathcal{H}$ that excludes the
$h$-direction cannot see either of them, so its diameter is
insensitive to $\Lambda$ while the version-space diameter grows
linearly in $\Lambda$.

The first inequality
$\widehat{\Delta}_{\mathcal{H}} \le \Delta^{\mathrm{emp}}_{\mathcal{F}}$ is by
definition of the supremum, since $\mathcal{H} \subseteq
\mathcal{V}_{\mathcal{F}}$.

By the endpoint premise of the corollary, both $g + \Lambda h$ and
$g - \Lambda h$ belong to $\mathcal{V}_{\mathcal{F}}(D_T; \varepsilon)$
(automatic for the slab class $\mathcal{F}^{\mathrm{slab}}_\Lambda$,
a stated premise otherwise); because $h \equiv 0$ on $T_{AB}$ their
empirical training losses are equal. We do not invoke symmetry of the
admissible set about $g$, which fails for a global ball centred at
zero. Under the identical-$\tilde g$ assumption, every member of
$\mathcal{H}$ takes the form $f = \tilde g$ on $C_{AB}$, so
$\widehat{\Delta}_{\mathcal{H}} = 0$; the two branches
$g \pm \Lambda h \in \mathcal{V}_{\mathcal{F}}$ give
$(\Delta^{\mathrm{emp}}_{\mathcal{F}})^2 \geq 4\Lambda^2 \mathbb{E}_C[h^2]$.

For the general statement, assume every member of $\mathcal{H}$
excludes the $h$-direction. Then any difference $\hat f_{h_1} -
\hat f_{h_2}$ between ensemble members depends only on their
background continuations. Holding those continuations fixed while
increasing $\Lambda$ leaves $\widehat{\Delta}_{\mathcal{H}}$
unchanged, whereas the admissible class contains the pair
$g \pm \Lambda h$ and therefore
$\Delta^{\mathrm{emp}}_{\mathcal{F}} \geq 2\Lambda
\|h\|_{L^2(P_C)}$ grows at least linearly in $\Lambda$. Hence
$\widehat{\Delta}_{\mathcal{H}} / \Delta^{\mathrm{emp}}_{\mathcal{F}}$
can be made arbitrarily small.
\end{proof}

\paragraph{Proof of Proposition~\ref{prop:diameter} (from \S\ref{sec:theory}).}
\begin{proof}
For any $f_1 = f_0 + h_1$ and $f_2 = f_0 + h_2$ in $\mathcal{V}_T(f_0)$,
$\|f_1 - f_2\|^2_{L^2(P_C)} = \|h_1 - h_2\|^2_{L^2(P_C)}$. The
supremum of $\|h_1 - h_2\|_{L^2(P_C)}$ over $h_1, h_2 \in
\mathcal{N}_T \cap B_\Lambda$ is $2 \sup_h \|h\|_{L^2(P_C)}$,
attained at $h_1 = h^\star$ and $h_2 = -h^\star$ by symmetry of
$B_\Lambda$; compactness of $K_\Lambda$ is used only to attain this
supremum as a maximum, not for the equality of suprema. Squaring
gives the factor of $4$.
\emph{Identifiability (scaling step).} If $\|h\|_{L^2(P_C)} = 0$ for
every $h \in \mathcal{N}_T$ the diameter vanishes. Conversely,
suppose some $h \in \mathcal{N}_T$ has $\|h\|_{L^2(P_C)} > 0$. If
$h \notin B_\Lambda$, then because $\Lambda > 0$ we rescale to
$\widetilde h = t h$ with $0 < t \le \Lambda / \|h\|_{\mathcal{F}}$,
so that $\widetilde h \in K_\Lambda$ and
$\|\widetilde h\|_{L^2(P_C)} = t\|h\|_{L^2(P_C)} > 0$. The admissible
invisible direction $\widetilde h$ differs from zero on a set of
positive $P_C$-measure and places non-zero risk on $C_{AB}$.
\end{proof}

\paragraph{Proof of Proposition~\ref{prop:minimax} (from \S\ref{sec:theory}).}
\begin{proof}
Upper bound: taking $\widehat{f} = f_0$, against any adversarial
$f = f_0 + h \in \mathcal{V}_T(f_0)$ the squared risk is
$\|h\|^2 \leq \sup_h \|h\|^2$. Lower bound: for any $\widehat{f} = f_0
+ u$ with $u \in L^2(P_C)$ (the deviation of the estimator from the
centre; the symbol $u$ avoids collision with the background $g$),
take the pair $f_\pm = f_0 \pm h^\star$
where $h^\star \in \mathcal{N}_T \cap B_\Lambda$ achieves the
supremum (by compactness). The parallelogram identity gives
$\tfrac{1}{2}[\|\widehat{f} - f_+\|^2 + \|\widehat{f} - f_-\|^2]
= \|u\|^2 + \|h^\star\|^2$, so $\max_s \|\widehat{f} - f_s\|^2 \geq
\|h^\star\|^2$. Taking the infimum of the max-risk over
$\widehat{f}$ yields the matching lower bound.
\end{proof}

\paragraph{Proof of Proposition~\ref{prop:few-shot} (from \S\ref{sec:theory}).}
\begin{proof}
Under $\lambda_{\min}(\tfrac{1}{m}\Phi^\top\Phi) \geq \kappa$,
$\mathrm{Cov}(\widehat{\theta}_{\mathrm{OLS}} \mid X_C) =
\sigma^2 (\Phi^\top\Phi)^{-1}$ so
$\mathbb{E}_\varepsilon[\|\widehat{\theta}-\theta\|_2^2 \mid X_C]
= \sigma^2 \mathrm{tr}[(\Phi^\top\Phi)^{-1}]
\leq \sigma^2 r / (m\kappa)$, which is
\eqref{eq:prop2-param}. Then \eqref{eq:prop2-pred} follows from
$\|\widehat h_\theta - h_\theta\|^2_{L^2(P_C)} =
(\widehat\theta - \theta)^\top \Sigma_C (\widehat\theta - \theta)
\leq L\|\widehat\theta - \theta\|_2^2$ with
$L = \lambda_{\max}(\Sigma_C)$.
\end{proof}

\emph{Remark (high-probability random design).}
Proposition~\ref{prop:few-shot} is conditional on the well-conditioning event
$\mathcal{E}_\kappa = \{\lambda_{\min}(\tfrac{1}{m}\Phi^\top\Phi) \geq \kappa\}$.
When the $m$ design points are drawn i.i.d.\ from $P_C$ rather than fixed,
$\mathcal{E}_\kappa$ holds with high probability under mild conditions. Suppose
$\|\phi(x)\|_2^2 \leq B$ for every admissible $x$ and
$\mu := \lambda_{\min}(\Sigma_C) > 0$. The summands
$\phi(x_i)\phi(x_i)^\top$ are i.i.d.\ PSD matrices with
$\lambda_{\max} \leq B$ and mean $\Sigma_C$, so standard matrix Chernoff
concentration for sums of bounded i.i.d.\ PSD matrices gives, for any
$\varepsilon \in (0,1)$,
\[
\Pr\!\Big[\lambda_{\min}\big(\tfrac{1}{m}\Phi^\top\Phi\big) \leq (1-\varepsilon)\mu\Big]
\;\leq\; r\,\exp\!\Big(-c\,\varepsilon^2\,\tfrac{m\mu}{B}\Big),
\]
so with $\kappa = \tfrac{\mu}{2}$ (i.e.\ $\varepsilon = \tfrac12$), the event
$\mathcal{E}_\kappa$ holds with probability at least $1-\delta$ once
$m \gtrsim (B/\mu)\log(r/\delta)$. On this event Proposition~\ref{prop:few-shot}
applies verbatim, giving the random-design bound
$O(\sigma^2 r/m)$ up to the constant $\kappa = \mu/2$.

\paragraph{Proof of Proposition~\ref{prop:i-opt} (from \S\ref{sec:theory}).}
\begin{proof}
\emph{Idea.} Reduce arbitrary estimators to parameter estimators
by orthogonal projection onto $\mathcal{H}_\phi$; compute the
exact OLS risk; use a diffuse-Gaussian-prior Bayes bound to show
no estimator improves on it; minimize in the design; then bound
the minimum by AM-GM.

\emph{Reduction to $\mathcal{H}_\phi$.} The target
$h_\theta = \phi^\top \theta$ lies in the finite-dimensional
subspace $\mathcal{H}_\phi = \operatorname{span}\{\phi_1, \ldots,
\phi_r\} \subset L^2(P_C)$. For any candidate estimator $\hat h$,
let $\hat h_\parallel$ be its orthogonal $L^2(P_C)$-projection onto
$\mathcal{H}_\phi$. Then
$\|\hat h - h_\theta\|_{L^2(P_C)}^2 = \|\hat h_\parallel -
h_\theta\|_{L^2(P_C)}^2 + \|\hat h - \hat h_\parallel\|_{L^2(P_C)}^2
\geq \|\hat h_\parallel - h_\theta\|_{L^2(P_C)}^2$, so projection
weakly reduces risk. It therefore suffices to consider estimators
of the form $\hat h = \phi^\top \hat\theta$.

\emph{Step 1: exact OLS risk.} For fixed $S$ with
$M_S := \Phi_S^\top\Phi_S$ invertible,
$\widehat{\theta}_{\mathrm{OLS}} - \theta = M_S^{-1}\Phi_S^\top
\varepsilon$, so
\begin{equation}
R_{\mathrm{OLS}}(S)
\;=\; \mathbb{E}_\varepsilon\bigl[(\widehat{\theta}-\theta)^\top
\Sigma_C (\widehat{\theta}-\theta)\bigr]
\;=\; \sigma^2 \operatorname{tr}[\Sigma_C M_S^{-1}].
\label{eq:ols-risk}
\end{equation}

\emph{Step 2: minimax lower bound.} For any $\tau > 0$, place
a Gaussian prior $\theta \sim \mathcal{N}(0, \tau^2 I_r)$. The
Bayes-optimal estimator has posterior covariance
$V_\tau = (\sigma^{-2} M_S + \tau^{-2} I_r)^{-1}$ and Bayes risk
$\operatorname{tr}[\Sigma_C V_\tau]$. Since minimax risk is at
least the Bayes risk under any prior,
$R_{\mathrm{minimax}}(S) \geq \operatorname{tr}[\Sigma_C V_\tau]$
for every $\tau$. Taking $\tau \to \infty$,
$V_\tau \to \sigma^2 M_S^{-1}$, and the bound becomes
$\sigma^2 \operatorname{tr}[\Sigma_C M_S^{-1}]$, which OLS attains
by \eqref{eq:ols-risk}. Hence
\begin{equation}
R_{\mathrm{minimax}}(S)
\;=\; \sigma^2 \operatorname{tr}[\Sigma_C M_S^{-1}]
\end{equation}
exactly (over all estimators, by the reduction above; not only
unbiased ones).

\emph{Step 3: $I$-optimality.} Minimizing over the full-rank
design domain $\mathfrak{D}_m$ gives
\[
    S^\star \;\in\; \arg\min_{S \in \mathfrak{D}_m}
    \operatorname{tr}[\Sigma_C M_S^{-1}],
\]
i.e.\ the exact minimax measurement rule is $I$-optimal on the
invisible basis.

\emph{Step 4: $\Theta(\sigma^2 r/m)$ scaling.} \emph{Upper
bound.} By assumption there exists an admissible design
$\widetilde{S}$ with
$\lambda_{\min}\bigl(m^{-1} \Phi_{\widetilde{S}}^\top
\Phi_{\widetilde{S}}\bigr) \geq \kappa$. Since $S^\star$
minimizes the exact risk,
\[
    R(S^\star) \;\leq\; R(\widetilde{S}),
\]
and Proposition~\ref{prop:few-shot} applied to $\widetilde{S}$
gives $R(\widetilde{S}) \leq \sigma^2 L r/(m\kappa)$ with
$L = \lambda_{\max}(\Sigma_C)$; hence
$R(S^\star) \leq \sigma^2 L r/(m\kappa)$.

\emph{Lower bound.} Define
$A_S := \Sigma_C^{1/2} M_S^{-1} \Sigma_C^{1/2} \succ 0$; then
$\operatorname{tr}(A_S) = \operatorname{tr}(\Sigma_C M_S^{-1})$
and $\det(A_S) = \det(\Sigma_C)/\det(M_S)$. Applying AM-GM to the
positive eigenvalues of $A_S$,
\[
    \operatorname{tr}[\Sigma_C M_S^{-1}]
    \;=\; \operatorname{tr}(A_S)
    \;\geq\; r\,\det(A_S)^{1/r}
    \;=\; r\, \frac{\det(\Sigma_C)^{1/r}}{\det(M_S)^{1/r}}.
\]
Under the feature bound $\|\phi(x)\|_2^2 \leq B$,
$\operatorname{tr}(M_S) \leq mB$; a second AM-GM on the
eigenvalues of $M_S$ then gives
$\det(M_S)^{1/r} \leq \operatorname{tr}(M_S)/r \leq mB/r$.
Combining,
\[
    R(S) \;\geq\; \sigma^2 \frac{r^2 \det(\Sigma_C)^{1/r}}{m B},
\]
so $c_- = r \det(\Sigma_C)^{1/r} / B$ in
\eqref{eq:iopt-scaling}. Provided $B = O(r)$,
$\det(\Sigma_C)^{1/r}$ is bounded away from zero, and
$L/\kappa = O(1)$, the two constants remain dimension-independent
and $R(S^\star) = \Theta(\sigma^2 r/m)$.
\end{proof}

\emph{Remark (Bayesian analogue matching the regularized real-data criterion).}
Proposition~\ref{prop:i-opt} is the exact minimax result for an \emph{unregularized}
full-rank Gaussian linear model ($M_S = \Phi_S^\top\Phi_S$), whereas the
electrocatalysis and MACE experiments use the \emph{regularized} criterion
$M_S = \lambda I + \Phi_S^\top\Phi_S$. The latter has its own exact
\emph{Bayes}-optimal-design characterization: under the proper prior
$\theta \sim \mathcal{N}(0, \tau^2 I)$ and Gaussian noise, the posterior mean is
ridge regression and the Bayes prediction risk of a design $S$ is
\[
R_{\mathrm{Bayes}}(S)
\;=\; \mathrm{tr}\!\big[\Sigma_C\,(\sigma^{-2}\Phi_S^\top\Phi_S + \tau^{-2} I)^{-1}\big],
\]
minimized by the Bayesian $I$-optimal design; this is exactly the criterion the
experiments optimize, with $\lambda = \sigma^2/\tau^2$. Thus the unregularized
frequentist minimax rule (Prop.~\ref{prop:i-opt}) and the regularized rule used
empirically are the optimal designs for the improper-prior and proper-prior
versions of the same problem; we call the empirical criterion
\emph{theorem-aligned} rather than identical to Prop.~\ref{prop:i-opt}.

\begin{proposition}[Risk under a misspecified invisible basis; Eq.~(\ref{eq:misspec})]
\label{prop:misspec}
Fix a design/readout basis $\Phi$ spanning $\mathcal{H}_\Phi \subseteq L^2(P_C)$
with $L^2(P_C)$-orthogonal projection $\Pi_\Phi$, and split the true residual
$h^\star = \Pi_\Phi h^\star + h^\star_\perp$. For OLS on a fixed full-rank design
$S$ with $z_i = h^\star(x_i) + \varepsilon_i$,
$\varepsilon \sim \mathcal{N}(0, \sigma^2 I)$, the risk decomposes as in
Eq.~(\ref{eq:misspec}), with sample-aliasing term
$b_S(h^\star_\perp) = \big\|(\Phi_S^\top\Phi_S)^{-1}\Phi_S^\top (h^\star_\perp|_S)\big\|^2_{\Sigma_C} \geq 0$,
which vanishes when $h^\star_\perp|_S$ is orthogonal to the columns of $\Phi_S$;
when $\Phi$ spans $\mathcal{N}_T$ ($h^\star_\perp = 0$) it recovers
Proposition~\ref{prop:i-opt}.
\end{proposition}

\paragraph{Proof of Proposition~\ref{prop:misspec}.}
\begin{proof}
Since $\widehat h \in \mathcal{H}_\Phi = \operatorname{span}(\Phi)$ and
$h^\star_\perp \perp \mathcal{H}_\Phi$ in $L^2(P_C)$, Pythagoras gives
$\|\widehat h - h^\star\|^2_{L^2(P_C)} = \|\widehat h - \Pi_\Phi h^\star\|^2_{L^2(P_C)}
+ \|h^\star_\perp\|^2_{L^2(P_C)}$, which is deterministic in the second term.
Write $\Pi_\Phi h^\star = \Phi\theta^\star$. OLS on $S$ solves
$\widehat\theta = (\Phi_S^\top\Phi_S)^{-1}\Phi_S^\top z_S$ with
$z_S = \Phi_S\theta^\star + (h^\star_\perp|_S) + \varepsilon_S$ (the middle term
is the misspecification residual sampled on $S$). Hence
$\widehat\theta - \theta^\star = (\Phi_S^\top\Phi_S)^{-1}\Phi_S^\top
\big[(h^\star_\perp|_S) + \varepsilon_S\big]$, and because
$\|\widehat h - \Pi_\Phi h^\star\|^2_{L^2(P_C)}
= (\widehat\theta - \theta^\star)^\top \Sigma_C (\widehat\theta - \theta^\star)$
with $\Sigma_C = \mathbb{E}_{P_C}[\phi\phi^\top]$ (using
$\|\Phi v\|^2_{L^2(P_C)} = v^\top\Sigma_C v$), taking $\mathbb{E}_\varepsilon$
and using $\mathbb{E}[\varepsilon_S\varepsilon_S^\top] = \sigma^2 I$ splits the
expectation into the deterministic aliasing term
$b_S(h^\star_\perp) = \|(\Phi_S^\top\Phi_S)^{-1}\Phi_S^\top(h^\star_\perp|_S)\|^2_{\Sigma_C}$
and the noise term
$\sigma^2\,\mathrm{tr}[\Sigma_C(\Phi_S^\top\Phi_S)^{-1}]$. Summing the three pieces
gives \eqref{eq:misspec}. If $\Phi_S^\top(h^\star_\perp|_S) = 0$ then
$b_S = 0$; and if $h^\star_\perp = 0$ (the basis spans $\mathcal{N}_T$) the first
two terms vanish and the expression reduces to the exact risk of
Proposition~\ref{prop:i-opt}.
\end{proof}

\begin{lemma}[Simplex invisible dimension]
\label{lem:simplex-dim}
Let $\mathcal{P}_D$ be polynomials in $d$ variables of total degree
$\leq D$, restricted to the simplex
$\Delta^{d-1} = \{x \in \mathbb{R}^d_{\geq 0} : \sum_j x_j = 1\}$
; equivalently, identified modulo the polynomials in
$\mathcal{P}_D$ that vanish on $\Delta^{d-1}$, i.e.\ the quotient
vector space $\mathcal{P}_D / [(\ell) \cap \mathcal{P}_D]$ with
$\ell(x) = \sum_j x_j - 1$. For any distinct $A, B \in \{1,
\ldots, d\}$ and any $D \ge 2$, and with $T_{AB}$ the \emph{complete
geometric union} of the two coordinate facets $\{x_A = 0\} \cup
\{x_B = 0\}$,
\begin{equation}
\dim \mathcal{N}_{T_{AB}}(\mathcal{P}_D)\big|_{\Delta^{d-1}}
\;=\; \binom{D + d - 3}{d - 1}.
\label{eq:simplex-dim}
\end{equation}
For any $T_{\mathrm{obs}} \subseteq T_{AB}^{\mathrm{geom}}$, including a
strict or finite subset, these pair-gated polynomials remain in
$\mathcal{N}_{T_{\mathrm{obs}}}$ (a canonical structural subspace of the
kernel), but they need not span it.
\end{lemma}

\begin{proof}
\emph{Idea.} The ambient invisible space has dimension
$\binom{D-2+d}{d}$; polynomials differing by a multiple of
$\ell(x) = \sum_j x_j - 1$ agree on the simplex, so subtract the
dimension of the intersection.

The ambient null space is $\operatorname{span}\{x^\alpha :
\alpha_A \geq 1, \alpha_B \geq 1, |\alpha| \leq D\}$, which by
the bijection $\alpha = \beta + e_A + e_B$ has dimension
$\binom{D-2+d}{d}$ (equation~\ref{eq:null-poly-dim}). The linear
form $\ell$ vanishes on $\Delta^{d-1}$; we claim
\begin{equation}
(\ell) \cap \mathcal{N}_{T_{AB}}(\mathcal{P}_D)
\;=\; \ell \cdot \mathcal{N}_{T_{AB}}(\mathcal{P}_{D-1}).
\label{eq:lem7-intersection}
\end{equation}
Indeed, since $\ell$ is coprime to $x_A x_B$ in the polynomial
ring, divisibility of $\ell q$ by $x_A x_B$ implies divisibility
of $q$ by $x_A x_B$; hence any $\ell q \in \mathcal{N}_{T_{AB}}$
has $q \in \mathcal{N}_{T_{AB}}(\mathcal{P}_{D-1})$, establishing
\eqref{eq:lem7-intersection}. The right-hand side has dimension
$\binom{D-3+d}{d}$, and the quotient dimension is
$\binom{D-2+d}{d} - \binom{D-3+d}{d} = \binom{D+d-3}{d-1}$
by Pascal's identity.

\emph{Surjectivity (the image is the full restriction kernel).}
Writing $L_D = (\ell) \cap \mathcal{P}_D$ and
$\mathcal{N}_D^{\mathrm{amb}} = x_A x_B \, \mathcal{P}_{D-2}$, the
restriction to the simplex identifies
$\mathcal{N}_D^{\Delta} \cong (\mathcal{N}_D^{\mathrm{amb}} + L_D)/L_D
\cong \mathcal{N}_D^{\mathrm{amb}} / (\mathcal{N}_D^{\mathrm{amb}}
\cap L_D)$ (an isomorphism of quotient spaces, not a literal
equality). We must check that \emph{every} polynomial vanishing on
both facets of $\Delta^{d-1}$ has a representative divisible by
$x_A x_B$, so that $\mathcal{N}_D^{\mathrm{amb}}$ surjects onto the
kernel. For $d \ge 3$, eliminate a coordinate $c \notin \{A, B\}$
via $x_c = 1 - \sum_{j \ne c} x_j$; the chart polynomial vanishes on
the relatively open pieces $\{x_A = 0\}$ and $\{x_B = 0\}$ of the
chart, and a polynomial vanishing on a relatively open subset of a
coordinate hyperplane is divisible by that coordinate, so it is
divisible by $x_A x_B$. For $d = 2$, set $t = x_A$, $x_B = 1 - t$;
a univariate polynomial vanishing at $t = 0$ and $t = 1$ is
divisible by $t(1-t) = x_A x_B$. In both cases the divisible
representative lies in $\mathcal{N}_D^{\mathrm{amb}}$, establishing
surjectivity; \eqref{eq:lem7-intersection} then gives the count.
(On the simplex, only the equivalence \emph{class} is divisible by
$x_A x_B$: not every representative is.)

For $d \ge 3$, the same chart shows the kernel is
$\{x_A x_B \cdot q(x_{-c}) : \deg q \leq D-2\}$, whose
dimension is that of degree-$\leq D-2$ polynomials in $d - 1$
variables, giving $\binom{D+d-3}{d-1}$; take the convention
$\binom{d-1}{d} = 0$ so the formula holds at $D = 2$.
\end{proof}
\clearpage

\section{Is Pair Difficulty a Stable Empirical Object?}
\label{app:stability}

\emph{Question and role.} Section~\ref{sec:theory} distinguished
realized learner error $D_m$ from structural ambiguity
$\Delta_{\mathcal{F}}$. Before using a learner-aggregated
difficulty $D_{\mathrm{cons}}$ as the target of the zero-label
analyses in Appendix~\ref{app:zerolabel}, we ask whether pair
difficulty is reproducible enough to support such an aggregate at
all. We test three progressively stronger forms of stability:
across learners, across random subsamples of the same copresence
region, and across genuinely different compositional contexts.
Materials are the primary demonstration; a partial protein
replication is reported in Appendix~\ref{app:crossdomain}.

\begin{figure}[!ht]\centering
\includegraphics[width=0.95\linewidth]{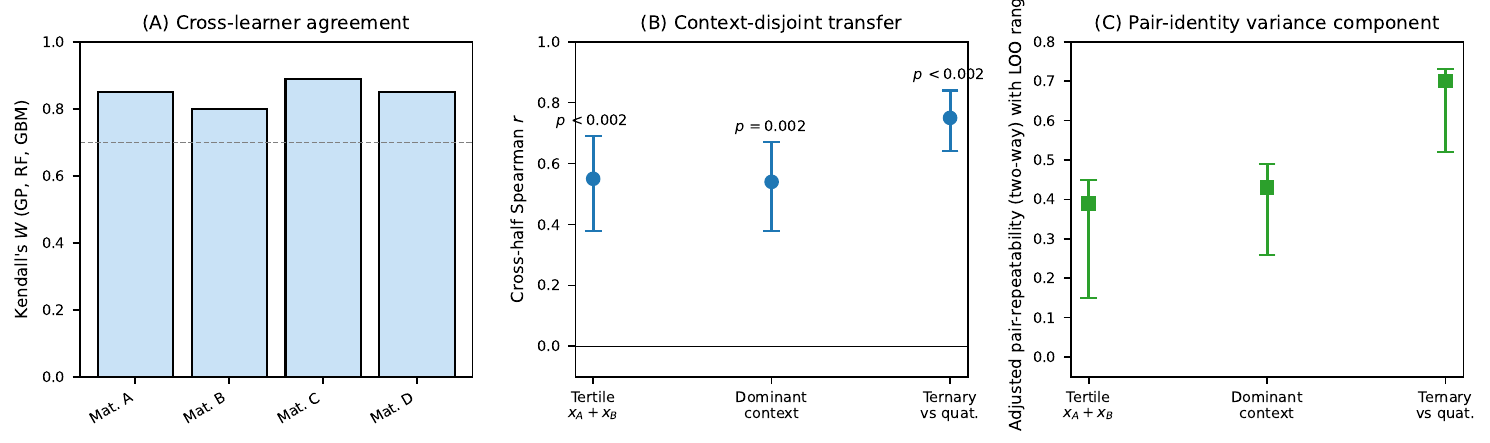}
\caption{\textbf{Pair difficulty is a stable empirical object
(electrocatalysis).}
(A) Cross-learner Kendall's $W$ across GP, RF, and GBM per system:
$W \ge 0.80$ in every system.
(B) Cross-context Spearman correlation across disjoint context
splits, with node-permutation tests. All node-permutation
$p \le 0.002$ ($500$ permutations; add-one Monte Carlo convention
$p = (b+1)/(500+1)$, whose minimum is $1/501 \approx 0.002$).
(C) Intraclass correlation (two-way, absorbing system and context
fixed effects) with leave-one-element-out sensitivity range: pair
identity accounts for a majority of the variance under the
compositional-order split, a substantial minority under the
third-element split, and is weak under a concentration-tertile
split.}
\label{fig:stability}
\end{figure}

\paragraph{Per-learner difficulty.}
For each pair $(A, B)$ in a system, we train a learner $m$ on the
non-copresence set $T_{AB}$ and evaluate its predictions on the
copresence set $C_{AB}$. Learner-specific realized difficulty is
measured by a rank error and a value error:
\begin{equation*}
    D_m^{\mathrm{rank}}(A, B) \;=\; 1 - \rho_S(y_C, \widehat{f}_m(x_C)),
    \qquad
    D_m^{\mathrm{value}}(A, B) \;=\; \frac{\mathbb{E}_C[(y_C - \widehat{f}_m(x_C))^2]}
                                              {\mathrm{Var}_C(y_C)}.
\end{equation*}
Sections~\ref{sec:stab-learner}--\ref{sec:stab-context} report
per-learner quantities; $D_{\mathrm{cons}}$ is defined at the end
of \S\ref{sec:stab-learner} once cross-learner agreement is
established.

\subsection{Across learners: can we justify an aggregate?}
\label{sec:stab-learner}

We fit three learners with distinct functional biases (a
Mat\'ern-$5/2$ Gaussian process, a random forest, and
gradient-boosted trees) on $T_{AB}$ and rank pairs within each
system by $D_m^{\mathrm{rank}}$. Cross-learner rank agreement is
reported as Kendall's coefficient of concordance $W$ across the
three rankings.

The pooled concordance across all $60$ materials pairs is
$W = \ResultAMaterialsKendallW$, and \emph{importantly every system
separately} has $W \geq \ResultAKendallSysB$: pooling is not
carrying the headline. In the protein panel, agreement is more
heterogeneous across substrates; the full protein breakdown is
reported in Appendix~\ref{app:crossdomain}.

\paragraph{Definition of $D_{\mathrm{cons}}$ (used from here on).}
Given the strong within-system concordance, we form a
learner-aggregated per-pair difficulty by within-system rank
aggregation. Let $P_s$ be the number of pairs in system $s$ and
let $\operatorname{rank}_{s, m}(A, B)$ denote the position of
$D_m^{\mathrm{rank}}(A, B)$ in the sorted list of pair
difficulties for learner $m$ (with $1 = $ easiest and $P_s = $
hardest; ties broken by $D_m^{\mathrm{value}}$). Define the
normalized rank
\[
    r_{s, m}(A, B)
    \;:=\; \frac{\operatorname{rank}_{s, m}(A, B) - 1}{P_s - 1}
    \;\in\; [0, 1].
\]
Then
\[
    \boxed{\;
    D_{\mathrm{cons}}(A, B; s)
    \;=\; \frac{1}{M} \sum_{m=1}^M r_{s, m}(A, B),
    \;}
\]
i.e.\ the mean of the within-system rank across the $M$ learners
in the panel. All zero-label and sparse-label analyses in
Appendices~\ref{app:zerolabel} and~\ref{app:calibration} use this
$D_{\mathrm{cons}}$ as the target. (Kendall's $W$ from
\S\ref{sec:stab-learner} is not $D_{\mathrm{cons}}$; it is the
concordance test that justifies forming the aggregate.)

\subsection{Across samples: is the ranking sample-reproducible?}
\label{sec:stab-splithalf}

For each pair we split its copresence set $C_{AB}$ into two
disjoint random halves $C^{(1)}$ and $C^{(2)}$ across $50$ outer
seeds, compute $D_m(C^{(1)})$ and $D_m(C^{(2)})$ from cached
predictions, and correlate them across pairs.

For each system $s$, seed $t$, and learner $m$ we compute the
Spearman rank correlation $\rho_S^{(s, t, m)}$ between
$D_m^{\mathrm{rank}}(C^{(1)})$ and $D_m^{\mathrm{rank}}(C^{(2)})$
across the $15$ pairs in that system. The reported statistic is
the median of these $\{ \rho_S^{(s, t, m)} \}$ over the
$4 \text{ systems} \times 50 \text{ seeds} \times 3
\text{ learners}$ cells; it equals
$\ResultBMaterialsMedianR$, with a positive correlation in every
one of the $50$ seeds for every system and every learner. These
are dependent resamples of the same copresence set, not
independent replications, so the statistic measures
sample-reproducibility, not independent-sample generalization.

\subsection{Across contexts: does the ranking survive
distribution change?}
\label{sec:stab-context}

Random split-halves share the same mixture of third-element and
compositional-order contexts. A stronger test is whether $D_m$
transfers across \emph{disjoint} context distributions. We use
three context-disjoint schemes on materials:
(a) tertile split of $x_A + x_B$ (bottom third vs top third, middle
dropped): the sharpest concentration separation;
(b) parity of the dominant non-$(A, B)$ element:
third-element-context change with concentration approximately
matched;
(c) ternary vs quaternary compositions: compositional-order change.

\begin{table}[!ht]
\centering\small
\caption{Cross-context Spearman correlations of pair-level
$D_m^{\mathrm{rank}}$ (GP row) with dependence-aware bootstrap
intervals and node-aware element-permutation tests. Materials
$n = 60$ pairs. Bootstrap: $2{,}000$ element-cluster (dyadic)
resamples (resample elements within system; each pair enters with
multiplicity $\mathrm{count}(a)\,\mathrm{count}(b)$); node
permutation: $500$ within-system element relabellings with the
add-one convention $p = (b+1)/(500+1)$ (minimum $1/501 \approx
0.002$). RF and GBM give the same qualitative ordering. The node-permutation
$p$ is a \emph{sensitivity} statistic: its exactness requires a sign-symmetry
null on the element effects, and with few distinct elements it is anticonservative
relative to the dyadic bootstrap; we read the bootstrap interval as primary.}
\label{tab:stab-context}
\begin{tabular}{lccc}
\toprule
Split & Bootstrap median $r$ & $90\%$ pctile interval (dyadic) & Node-perm $p$ \\
\midrule
Tertile $x_A + x_B$         & $+\ResultBTertileNodePermR$ & $[+0.23, +0.76]$ & $\le 0.002$ \\
Dominant-context parity     & $+\ResultBContextDomNodePermR$ & $[+0.22, +0.74]$ & $\le 0.002$ \\
Ternary vs quaternary       & $+\ResultBRegimeNodePermR$ & $[+0.55, +0.88]$ & $\le 0.002$ \\
\bottomrule
\end{tabular}
\end{table}

All three splits have dependence-aware (element-cluster) bootstrap
CIs strictly above zero. The most demanding split (tertile of
$x_A + x_B$) gives GP bootstrap $r = +0.55$, dyadic CI
$[+0.23, +0.76]$; the regime split is $[+0.55, +0.88]$. Under this
dyadic scheme the correlation is positive in every resample for the
regime split and in $\ge 99\%$ of resamples for the tertile and
dominant-context splits (a naive pair-row bootstrap gives $\sim
1.5$--$2\times$ narrower CIs and $100\%$ positivity, but overstates
precision by treating element-sharing pairs as independent).

\subsection{Variance decomposition: pair identity vs context}
\label{sec:stab-icc}

A high cross-half correlation does not by itself say how much of
the observed variance in $D_m$ is a genuine pair-level effect. We
fit a model that absorbs system and context \emph{fixed} effects and
places a random effect on within-system pair identity, on the tidy
long form
\[
    D_{m, s, AB, c}
    \;=\; \mu + \alpha_s + \gamma_c + u_{s, AB} + \varepsilon_{s, AB, c},
\]
where $(s, AB)$ indexes the (system, pair), $c$ indexes the two
context halves, $\alpha_s$ and $\gamma_c$ are system and context
fixed effects, $u_{s, AB} \sim \mathcal{N}(0, \sigma_u^2)$ is the
random effect of within-system pair identity, and
$\varepsilon_{s, AB, c} \sim \mathcal{N}(0, \sigma_\varepsilon^2)$
is the residual. The reported statistic is the \emph{adjusted
pair-repeatability variance fraction}
\[
    \boxed{\;
    \mathrm{ICC}^{\mathrm{adj}}_{\mathrm{pair}}
    \;=\; \mathrm{VPC}_{\mathrm{pair}\mid s,c}
    \;=\; \frac{\sigma_u^2}{\sigma_u^2 + \sigma_\varepsilon^2}
    \quad\text{(after absorbing }\alpha_s, \gamma_c\text{)}.
    \;}
\]
This is a \emph{conditional variance-partition coefficient}
($\mathrm{VPC}_{\mathrm{pair}\mid s,c}$). The random effect is
within-system pair identity, not a universal element-pair constant
across catalyst substrates; absorbing $\alpha_s$ prevents
between-system mean differences from inflating this fraction. A low
$\mathrm{VPC}$ is consistent with context modulation, but its
residual also absorbs finite-test noise and pair-by-context
variation, so it is not by itself a direct measure of modulation
strength.

\begin{table}[!ht]
\centering\small
\caption{Adjusted pair-repeatability variance fraction (two-way, system and context fixed effects absorbed) point estimate with
leave-one-element-out sensitivity range for pair-level
$D_m^{\mathrm{rank}}$ across two disjoint context halves. Splits
vary third-element context (dominant parity), compositional order
(ternary vs quaternary), or the pair's own concentration (tertile).
GP row shown; RF and GBM patterns are identical.}
\label{tab:stab-icc}
\begin{tabular}{lcc}
\toprule
Split                       & $\mathrm{ICC}^{\mathrm{adj}}_{\mathrm{pair}}$ point & Leave-one-element-out sensitivity range \\
\midrule
Tertile $x_A + x_B$         & $\ResultBICCTertileAnova$
                            & $[\ResultBICCTertileJackLo, \ResultBICCTertileJackHi]$ \\
Dominant-context parity     & $\ResultBICCContextDomAnova$
                            & $[\ResultBICCContextDomJackLo, \ResultBICCContextDomJackHi]$ \\
Ternary vs quaternary       & $\mathbf{\ResultBICCRegimeAnova}$
                            & $[\ResultBICCRegimeJackLo, \ResultBICCRegimeJackHi]$ \\
\bottomrule
\end{tabular}
\begin{flushleft}\footnotesize
Estimator: closed-form ANOVA-style random-effects ICC grouping the
two context-half observations per $(s, AB)$, computed on residuals
after absorbing \emph{system and context main effects} as fixed
effects (two-way model $D = \mu + \alpha_s + \gamma_c + u_{s, AB} +
\varepsilon$), with negative-variance clipping at zero. This
corrects a one-way $\mathrm{ICC}(1,1)$, which credits between-system
mean differences to pair identity and is upward-biased (one-way
dominant-context point $0.54$ vs two-way $\ResultBICCContextDomAnova$).
Ranges from a leave-one-element-out sensitivity: for each of the
eleven elements in $\bigcup_s \operatorname{elems}(s)$, drop all
pairs containing that element and recompute (11 folds); these are
descriptive deletion ranges, not confidence intervals.
\end{flushleft}
\end{table}

Once system and context main effects are absorbed, pair identity
accounts for a majority of the variance in $D_m$ only under the
compositional-order (ternary-vs-quaternary) split
($\ResultBICCRegimeAnova$, leave-one-element-out
$[\ResultBICCRegimeJackLo, \ResultBICCRegimeJackHi]$, entirely above
one half). Under the dominant-context split it is a substantial but
\emph{minority} component ($\ResultBICCContextDomAnova$,
$[\ResultBICCContextDomJackLo, \ResultBICCContextDomJackHi]$, below
one half across all element-deletions); the earlier one-way estimate
of $0.54$ was inflated by between-system differences. Under the
sharpest concentration split (tertile) the pair-identity share is
lower still ($\ResultBICCTertileAnova$). The safe conclusion is that
pair repeatability is a majority component across compositional-order
splits, a substantial minority across third-element splits, and low
across concentration tertiles; the reported brackets are
leave-one-element-out (LOEO) sensitivity ranges, not confidence
intervals or a jackknife standard error.

\emph{Takeaway.} On electrocatalysis, compositional-OOD pair
difficulty is not merely a learner- or sample-specific artifact.
Pair rankings are strongly concordant across model classes ($W_s
\ge 0.80$ in every system), highly reproducible across disjoint
random test halves, and remain positively associated across
disjoint third-element and compositional-order contexts. The ICC
analysis therefore cautions against interpreting
$D_{\mathrm{cons}}(A, B; s)$ as an unconditional intrinsic
property of an element pair: pair repeatability is a majority
component across compositional-order splits, a substantial
minority across third-element splits, and lowest across concentration
tertiles. A low tertile VPC is consistent with concentration
modulating difficulty, but the residual there also contains
measurement/test noise, so quantifying the modulation itself would
need a separate model rather than reading it off the VPC. The downstream analyses in
Appendices~\ref{app:zerolabel} and~\ref{app:calibration} use the
system-specific aggregate $D_{\mathrm{cons}}(A, B; s)$ over each
system's observed copresence distribution. A partial cross-domain
replication on the retained protein DMS landscapes is collected
in Appendix~\ref{app:crossdomain}, which also documents the
exclusion of A4-Seuma for insufficient pair copresence.
\clearpage

\section{Zero-Label Detection Fails on Real Materials}
\label{app:zerolabel}

\emph{Question and role.} Appendix~\ref{app:stability} established
that the materials pair-difficulty target $D_{\mathrm{cons}}(A, B; s)$
is stable enough to route on. Can any zero-label signal (element
descriptors, uncertainty scores, or ensemble disagreement)
predict it? A natural hope is that such signals could identify
difficult pairs without requiring joint measurements. The purpose
of this appendix is to test that hypothesis under
dependence-aware, leakage-aware, and out-of-system evaluation.
The narrative arc is \emph{apparent signal $\rightarrow$ remove
endogenous overlap $\rightarrow$ test true transfer}: an
apparently strong association survives dependence-aware inference
but collapses when target and predictor share no model families,
and further fails under leave-one-system-out transfer.

\emph{Zero-label protocol.} Throughout this appendix, ``zero-label''
means that no response values $y$ from $C_{AB}$ are observed;
unlabeled compositions and covariates in $C_{AB}$ may be used to
construct a diagnostic. This matches the transductive-covariate
regime used elsewhere in the paper (\S\ref{sec:real-setup},
Appendix~\ref{app:mp-support}).

\emph{How to read this appendix.} Each result below leads with its headline
finding and interpretation; the supporting statistical machinery (dyadic
bootstrap, Fisher-$z$ references, permutation tests, denominator definitions) and
scope caveats follow rather than precede the conclusion.

\begin{figure}[!ht]\centering
\includegraphics[width=0.95\linewidth]{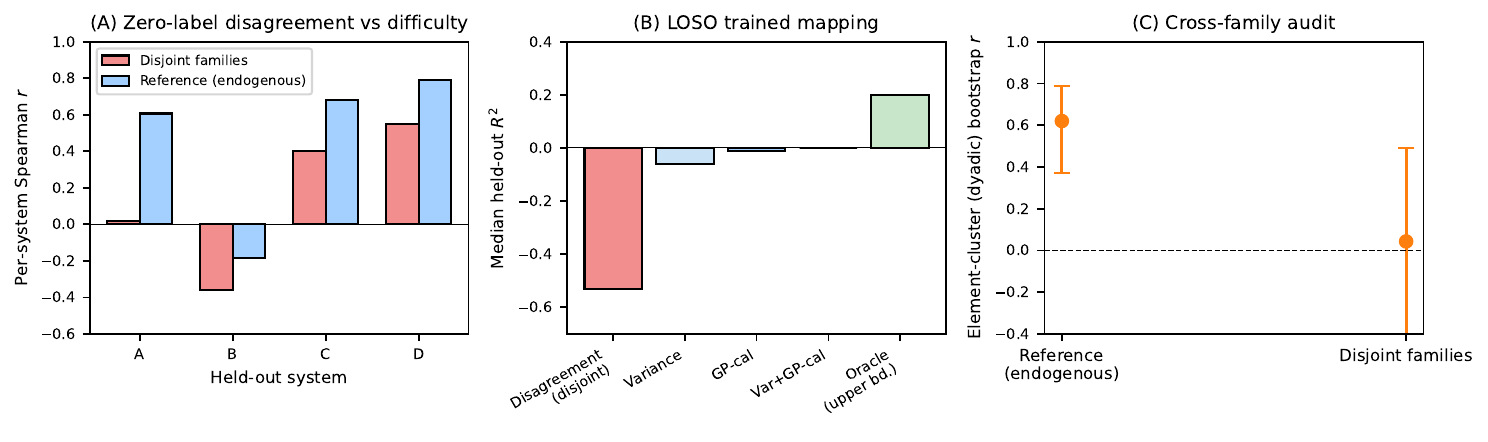}
\caption{\textbf{Zero-label disagreement fails under transfer;
sparse calibration provides a reference comparison.}
(A) Per-system Spearman between ensemble disagreement
$V_{\mathcal{H}}^{\mathrm{avg,\,rank}}$ and consensus pair
difficulty $D_{\mathrm{cons}}$: red = disjoint families, blue =
endogenous reference. Removing model-family overlap substantially
weakens the association and reveals strong between-system
heterogeneity, including a negative association on system B.
(B) LOSO trained-mapping median held-out $R^2$: zero-label
disagreement is worse than the held-out-system mean predictor
(negative standard test-set $R^2$). The
sparse-label comparators shown here reference the leakage-safe
matched-target analysis in
Appendix~\ref{sec:cal-loso-results} rather than a direct
comparison on this appendix's $D_{\mathrm{cons}}$ target.
(C) Cross-family audit under the element-cluster (dyadic) bootstrap
of Table~\ref{tab:zerolabel-cross}: the endogenous reference sits at
$r = 0.62$ [0.37, 0.79]; the disjoint-family configuration
collapses to $r = 0.043$ [$-0.42$, 0.49].}
\label{fig:zerolabel}
\end{figure}

\subsection{The evaluation ladder}
\label{app:zerolabel-ladder}

We use ``cross-system transfer'' to mean three increasingly
stringent objects. Every later result in this appendix is one of
these three and no other; the labels are used verbatim.

\begin{itemize}[leftmargin=1.6em,itemsep=1pt]
    \item \emph{Per-system association.} For each system $s$
        separately, compute
        $\rho_S(V(s), D_{\mathrm{cons}}(s))$ over that system's
        pairs. Aggregate vian element-cluster (dyadic) bootstrap or
        Fisher-$z$ pool over the four per-system correlations. No
        mapping is learned. This is descriptive: it says whether
        the signal tracks difficulty within known systems.
    \item \emph{Disjoint-family association.} As above, but the
        model families used to construct the \emph{target}
        $D_{\mathrm{cons}}$ and the \emph{predictor} $V$ share no
        members. This removes endogenous target--predictor overlap
        while keeping the same descriptive statistic.
    \item \emph{LOSO trained transfer.} Fit a mapping
        $D_{\mathrm{cons}} \sim V$ on the pairs of three systems
        and predict the fourth. Report held-out $R^2$ and
        Spearman. This is a strictly stronger claim than either
        association: it asks whether a signal-to-difficulty rule
        learned on other systems predicts a genuinely unseen system.
\end{itemize}

Only the third certifies transfer; the first is descriptive; the
second removes the model-family confound that would otherwise
inflate the first.

\subsection{Simple pair descriptors do not transfer}
\label{app:zerolabel-physical}

We first ask whether pair identity alone (via a compact set of
$31$ physical pair descriptors) transfers as a zero-label
difficulty predictor. The features are atomic number, Pauling
electronegativity, atomic radius, group, period, valence electrons,
$d$-electron count, transition-metal and lanthanide indicators, and
their pairwise differences, sums, and products. They carry pair
identity but deliberately \emph{omit} the catalyst substrate,
third-element context, and concentration regime that
Appendix~\ref{app:stability} identified as material modulators of
$D_{\mathrm{cons}}$.

Under LOSO trained transfer, ridge regression on the $31$-feature
representation gives median held-out $R^2 = -3.25$ across the four
systems (catastrophic failure on system~D, $R^2 \approx -30$);
gradient boosting on the same features gives $R^2 = -2.06$. Adding
these features to GP-calibration does not improve on GP-calibration
alone. This is a narrow negative: \emph{simple element identity
alone does not provide a transferable difficulty router}. It does
not rule out richer, context-aware physical routers.

\subsection{Standard uncertainty and support-geometry scores
do not transfer as routers}
\label{app:zerolabel-baselines}

The natural next attempts are ordinary out-of-distribution and
uncertainty signals: mean Mahalanobis distance of $C_{AB}$ from
$T_{AB}$; mean posterior variance of a Mat\'ern-$5/2$ GP fit on
$T_{AB}$; median nearest-training distance; the support-geometry
median $x_A x_B$ on $C_{AB}$; the third-element Shannon entropy on
$C_{AB}$; and the number of non-zero composition components.

Under both per-system association and LOSO trained transfer, none
of these signals produces an element-cluster (dyadic) bootstrap CI on the
pooled association strictly above zero, and none yields a
positive held-out $R^2$ under LOSO. The scoped conclusion is that
\emph{these signals do not provide a transferable pair-level
routing signal on this benchmark}. This is narrower than a general
failure claim: GP posterior variance is a useful \emph{within-pair
acquisition} signal in \S\ref{sec:real-setup} (choosing which
compositions to measure inside a chosen pair), which is a
different decision problem from pair-level routing (choosing
\emph{which pairs to measure at all}).

\subsection{Naïve disagreement appears strongly predictive}
\label{app:zerolabel-D}

The theoretically motivated attempt is Corollary~\ref{cor:blind-spot}'s
object: the pairwise disagreement of a training-compatible
ensemble on $C_{AB}$. We use a seven-model calibrated library (
Mat\'ern GP, RBF GP, RF, GBM, MLP, additive-kernel GP,
polynomial-ridge) with an $80/20$ fit/validation split of
$T_{AB}$, one-standard-error retention on validation MSE, and
affine calibration on $T$-validation predictions. The primary
statistic is $V_{\mathcal{H}}^{\mathrm{avg,\,rank}}$ (see
Appendix~\ref{app:three-quantities}).

\paragraph{The apparently strong result.}
Na\"ively pooling all $60$ material-pair rows gives
\[
    \rho_S\bigl(V_{\mathcal{H}}^{\mathrm{avg,\,rank}},\;
    D_{\mathrm{cons}}\bigr)
    \;\approx\; \ResultDPoolEndog
    \qquad (n = 60).
\]
Because pairs within a system share elements, this pooled-row
calculation does not provide valid independent-row inference.
Under an element-cluster (dyadic) bootstrap, however, the endogenous
reference effect size remains
\[
    r = \ResultDPoolEndogBootMed
    \quad [\ResultDPoolEndogBootLo,\, \ResultDPoolEndogBootHi].
\]
So the story is not \emph{``the naïve $r$ collapses under
dependence-aware inference.''} Row dependence mainly affects the
na\"ive significance calculation, not the effect-size point
estimate. The signal is heterogeneous, however: systems A, C, D
show per-system $r > 0.60$, while system B is already
qualitatively negative
($r_B = \ResultDLOSOSysBOrig$, $p = 0.52$, $n = 15$).

At this point a reasonable analyst would provisionally conclude
that ensemble disagreement predicts pair difficulty.

\subsection{Removing endogenous model overlap collapses the
signal}
\label{app:zerolabel-cross}

The reference association is potentially inflated because
$D_{\mathrm{cons}}$ (built from GP, RF, GBM as the learner panel
in Appendix~\ref{sec:stab-learner}) shares model family with three
of the seven ensemble members. We partition the library into two
disjoint groups:
\[
    \mathcal{G}_1 = \{\text{Mat\'ern GP},\ \text{RF},\ \text{GBM}\},
    \qquad
    \mathcal{G}_2 = \{\text{RBF GP},\ \text{additive GP},\ \text{MLP},\ \text{polynomial ridge}\},
\]
and, for each pair, compute the target
$D_{\mathrm{cons}}^{(\mathcal{G}_1)}$ from the retained members of
$\mathcal{G}_1$ and the predictor
$V_{\mathcal{H}}^{\mathrm{avg,\,rank}}(\mathcal{G}_2)$ from the
retained members of $\mathcal{G}_2$. Target and predictor now
share no model families. All inference is under the same
element-cluster (dyadic) bootstrap and node-aware element-permutation
protocol.

\begin{table}[!ht]
\centering\small
\caption{Cross-family sensitivity of the
$V_{\mathcal{H}}^{\mathrm{avg,\,rank}}$--$D_{\mathrm{cons}}$
association. Bootstrap median and $5$--$95\%$ CI over $2{,}000$
element-cluster (dyadic) resamples that resample elements within
system and admit each pair with multiplicity
$\mathrm{count}(a)\,\mathrm{count}(b)$, preserving the dyadic
dependence among element pairs; Fisher-$z$ pool over the four
per-system Spearman correlations (a conventional non-dyadic
reference, not a dependence-aware interval: it treats the
within-system pair rows as independent); node-aware permutation over
element identities ($500$ within-system relabellings). The
dependence-aware intervals are $\sim 1.5$--$2\times$ wider than a
naive pair-row bootstrap but preserve every sign and
above-/spanning-zero verdict.}
\label{tab:zerolabel-cross}
\begin{tabular}{lccc}
\toprule
Configuration & Bootstrap $r$ [$90\%$ pctile] & Fisher-$z$ pool [95\% CI] & Node-perm $p$ \\
\midrule
Reference (endogenous, $\mathcal{G}_1 \cup \mathcal{G}_2$)
    & $\ResultDPoolEndogBootMed$ [$\ResultDPoolEndogBootLo$, $\ResultDPoolEndogBootHi$]
    & $\ResultDPoolEndogFisherR$ [$\ResultDPoolEndogFisherLo$, $\ResultDPoolEndogFisherHi$]
    & $\le 0.002$ \\
Disjoint families ($\mathcal{G}_2 \rightarrow \mathcal{G}_1$)
    & $\ResultDCrossDirOneBootMed$ [$\ResultDCrossDirOneBootLo$, $\ResultDCrossDirOneBootHi$]
    & $\ResultDCrossDirOneFisherR$ [$\ResultDCrossDirOneFisherLo$, $\ResultDCrossDirOneFisherHi$]
    & $\ResultDCrossDirOneNodePerm$ \\
\bottomrule
\end{tabular}
\end{table}

The contrast is
\[
    \boxed{\;
    \rho \;\approx\; \ResultDPoolEndogBootMed
    \quad
    \xrightarrow[\text{same inference}]{\text{remove family overlap}}
    \quad
    \rho \;\approx\; \ResultDCrossDirOneBootMed.
    \;}
\]
The apparent association is therefore not robust to removing
model-family overlap. Under disjoint families the point estimate is
near zero ($\rho \approx \ResultDCrossDirOneBootMed$) but with
\emph{wide} uncertainty: the element-cluster bootstrap $5$--$95\%$
interval $[\ResultDCrossDirOneBootLo, \ResultDCrossDirOneBootHi]$
spans zero and admits substantial effects in either direction, and
the node-permutation test is non-significant
($p = \ResultDCrossDirOneNodePerm$). A confidence interval spanning
zero and a non-significant permutation test do not \emph{establish} a
near-zero effect, only that this zero-label router is not
demonstrably useful here. (The Fisher-$z$ pool
$\ResultDCrossDirOneFisherR$ is a conventional non-dyadic reference,
treating the within-system pair rows as independent; inference is
taken from the node bootstrap and permutation.) The two family groups
share no identical model specification but both contain Gaussian
processes, so they are not disjoint at the broad model-family level;
the held-out transfer failure below (\S\ref{app:zerolabel-loso}) is
the stronger evidence for this router's limited utility. System~B
remains qualitatively negative in both configurations
($r_B = \ResultDCrossDirOneSysB$ under disjoint families).

\subsection{Out-of-system prediction fails}
\label{app:zerolabel-loso}

The strongest test escalates from association to actual transfer:
fit the linear mapping
$D_{\mathrm{cons}} \sim V_{\mathcal{H}}^{\mathrm{avg,\,rank}}$ on
three systems and predict $D_{\mathrm{cons}}$ on the held-out
fourth.

\emph{Result.} Under disjoint-family LOSO, zero-label disagreement gives median
held-out $R^2 = \ResultDLOSODisjointRsq$ and Spearman
$\rho = \ResultDLOSODisjointSp$.
\emph{Interpretation.} Zero-label disagreement does not transfer across systems:
it predicts held-out pair difficulty worse than the held-out system's own mean.
\emph{Caveat.} This establishes the failure of \emph{this router} across systems,
not the chemical cause of any single system's behavior; the $R^2$ denominator
convention and the coefficient sign-flip driving the large negative values are
detailed below.

\begin{table}[!ht]
\centering\small
\caption{LOSO trained transfer: fit
$D_{\mathrm{cons}} \sim V_{\mathcal{H}}^{\mathrm{avg,\,rank}}$ on
three systems, predict the fourth. Median across four held-out
systems.}
\label{tab:zerolabel-loso}
\begin{tabular}{lcc}
\toprule
Configuration & Median held-out $R^2$ & Median held-out Spearman \\
\midrule
Endogenous (all-vs-all)             & $+\ResultDLOSOEndogRsq$    & $+\ResultDLOSOEndogSp$ \\
Disjoint families ($\mathcal{G}_2 \rightarrow \mathcal{G}_1$) & $\ResultDLOSODisjointRsq$  & $\ResultDLOSODisjointSp$ \\
\bottomrule
\end{tabular}
\end{table}

Under disjoint-family LOSO,
\[
    \boxed{\;
    R^2_{\mathrm{LOSO}} \;=\; \ResultDLOSODisjointRsq,
    \qquad
    \rho_{S,\mathrm{LOSO}} \;=\; \ResultDLOSODisjointSp,
    \;}
\]
i.e.\ \emph{worse than the held-out-system mean predictor} (the
standard test-set $R^2 = 1 - \sum_{i}(y_i - \widehat y_i)^2 /
\sum_i (y_i - \bar y_{\mathrm{test}})^2$ uses the held-out-set mean
$\bar y_{\mathrm{test}}$ in its denominator, so a negative value
means worse than that mean; individual held-out
$R^2$ ranges from $-0.07$ on system A to $-5.93$ on system D,
driven by the linear mapping's coefficient sign flipping across
systems). Even the endogenous configuration (with maximal
shared model-family variance) has median held-out $R^2$ only
$+\ResultDLOSOEndogRsq$, catastrophically bad on system~B
($R^2 = \ResultDLOSOEndogSysBRsq$).

\paragraph{Comparison with sparse-label calibration.}
Under the same LOSO trained-transfer protocol,
Section~\ref{sec:cal-loso} reports that a small number of
copresence labels used for pair-region variance and GP-calibration
recovers to near-zero calibrated $R^2$ with positive held-out
Spearman:

A matched-target head-to-head comparison of zero-label
disagreement against the sparse-label probes on the same
leakage-safe target $D_{\mathrm{test}}^{\mathrm{hold}}$ is
deferred to Appendix~\ref{app:calibration}
(\S\ref{sec:cal-loso-results}), where every row of the comparison
is fit to the identical held-out target under the identical
$50$-draw LOSO protocol.

\emph{Takeaway.} Pair difficulty is reproducible enough to define
a routing target (Appendix~\ref{app:stability}), but none of the
zero-label signals tested here provides reliable cross-system
prediction of that target. Ensemble disagreement is particularly
instructive: an apparently strong association survives
dependence-aware inference when target and predictor share model
families, but collapses when those families are separated and
fails under true leave-one-system-out transfer. Thus
finite-ensemble disagreement is not a reliable zero-label router
for $D_{\mathrm{cons}}$ on these materials systems. This
empirical result is consistent with
Corollary~\ref{cor:blind-spot}, which shows why finite-model
disagreement need not reveal admissible ambiguity;
Appendix~\ref{app:synth} isolates that mechanism under controlled
conditions and does not identify the chemical cause of system~B's
behaviour on real materials.
\clearpage

\section{Synthetic Witness for the Shared Blind-Spot Mechanism}
\label{app:synth}

\emph{Question.} Can a collection of plausible fitted models
remain nearly unchanged as the set of observationally compatible
worlds becomes arbitrarily wider?

\emph{Role.} Appendix~\ref{app:zerolabel} showed that zero-label
ensemble disagreement fails as a transferable difficulty router on
real materials, but that experiment does not establish why. Here
we isolate the shared-blind-spot mechanism of
Corollary~\ref{cor:blind-spot} in a controlled setting where the
admissible ambiguity is known analytically. The synthetic does
\emph{not} claim to reproduce real chemistry; it isolates the
theoretical claim.

\subsection{Controlled construction}
\label{app:synth-setup}

We use the same observationally-equivalent (OE) worlds construction
as \S\ref{sec:theory}. The true world is
$f^\star(x) = g(x) + \lambda^\star h(x)$; training data lie on
$T = \{x_A x_B = 0\}$; the copresence region is
$C = \{x_A x_B > 0\}$. The admissible class radius $\Lambda$
determines what interactions are consistent with the observed
support and the class norm bound.

We run two clean sub-experiments:
\begin{itemize}[leftmargin=1.6em,itemsep=1pt]
    \item \emph{Realized-error experiment
    (\S\ref{app:synth-difficulty})}: vary $\lambda^\star \in \{0,
    0.25, 0.5, 1, 2, 3\}$. The identity
    $D_C(\lambda^\star) = D_C(0) + a\lambda^\star +
    b(\lambda^\star)^2$ is a fixed-predictor decomposition
    verified against $b = \mathbb{E}_C[h^2]$.
    \item \emph{Ambiguity experiment
    (\S\ref{app:synth-ambiguity})}: fix $\lambda^\star = 0$ (truth
    inside every admissible class) and sweep
    $\Lambda \in \{0.25, 0.5, 1, 2, 3\}$. Only the admissible
    ambiguity changes; training data and $f^\star$ are held fixed.
    This isolates whether finite ensembles track $\Delta^{\mathrm{orc}}$.
\end{itemize}
Three ensembles $\mathcal{H}$ are tested throughout:
\begin{itemize}[leftmargin=1.6em,itemsep=1pt]
    \item \textbf{Restricted}: additive-only members that cannot
        represent $h$ (ridge, additive-kernel GP with varying
        length-scales, standardised ridge).
    \item \textbf{Rich-naïve}: standard learners that can
        represent interactions but default to smooth extrapolation
        (RF, GBM, MLP, RBF-GP, polynomial-kernel ridge).
    \item \textbf{Covering}: the one-dimensional OE class
        \[
            \mathcal{H}_{\mathrm{cover}}(\Lambda)
            \;=\; \{\, g - \Lambda h,\; g,\; g + \Lambda h \,\},
        \]
        a three-point, diameter-attaining subset of the fixed-$g$
        slice $\mathcal{S}_\Lambda(g,h) = \{g + \lambda h :
        |\lambda| \le \Lambda\}$ (not the full continuum, but it
        realizes the slice value-space diameter). The covering ensemble
        uses knowledge of $h$; it is a controlled witness, not a
        practical algorithm.
\end{itemize}
For the ambiguity experiment (\S\ref{app:synth-ambiguity}), for
each seed we sample $D_T$, the copresence pool $C$, and every
model random seed \emph{once}, then hold them fixed across the
$\Lambda$-grid. Only the covering branches change with $\Lambda$;
$f^\star = g$ (since $\lambda^\star = 0$) and the fitted
restricted/rich-naïve predictors do not depend on $\Lambda$ at
all. Each cell is repeated across eight seeds with $n_T = 200$
and $n_C = 100$.

\begin{figure}[!ht]\centering
\includegraphics[width=0.72\linewidth]{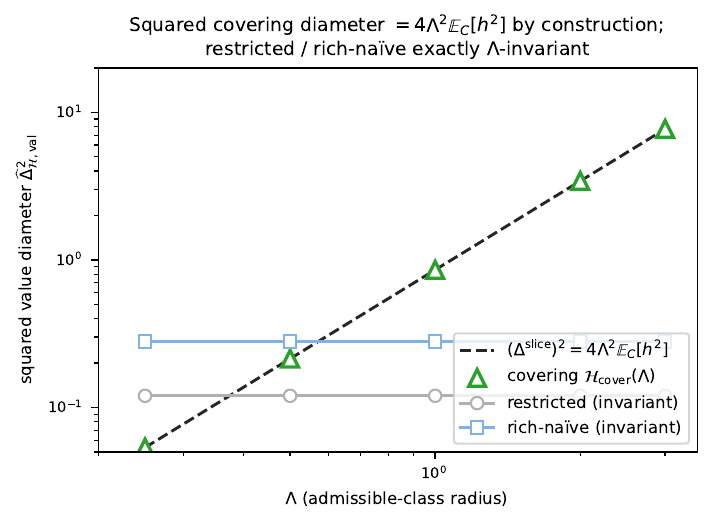}
\caption{\textbf{Shared blind-spot mechanism, synthetic witness
(fixed-data protocol).}
Squared value diameter
$\widehat{\Delta}^{2}_{\mathcal{H},\mathrm{val}}$ vs admissible-class
radius $\Lambda$ on log-log scale, with $\lambda^\star = 0$ and
$(D_T, C, \text{model seeds})$ drawn once per seed and held fixed
across the $\Lambda$-grid.
\textbf{Covering}: $\mathcal{H}_{\mathrm{cover}}(\Lambda) =
\{g - \Lambda h,\, g,\, g + \Lambda h\}$; its maximum pairwise
squared distance equals the fixed-$g$ slice diameter
$\Delta^{\mathrm{slice}}(\Lambda)^2 = 4\Lambda^{2}
\mathbb{E}_{C}[h^{2}]$ (dashed line; this equals
$(\Delta^{\mathrm{orc}})^2$ here only under the coefficient-compatible
normalization $\|h\|_{\mathcal{F}} = 1$ of the one-dimensional
$\mathcal{N}_T$, as stated in the text) as a construction
identity.
\textbf{Restricted / rich-naïve}: since these fitted predictors
do not depend on $\Lambda$ under the fixed-data protocol, their
squared diameter is exactly $\Lambda$-invariant. The two
horizontal levels shown are illustrative seed-representative
constants; the $\Lambda$-invariance is by construction.}
\label{fig:synth-mechanism}
\end{figure}

\subsection{Hidden interactions increase realized error}
\label{app:synth-difficulty}

Because $h \equiv 0$ on $T$, the training data are identical for
every $\lambda^\star$: the fitted predictor $\widehat{f}_T$ does
not depend on $\lambda^\star$. For a fixed $\widehat{f}_T$ the
excess $L^2(P_C)$-error decomposes exactly as
\begin{equation}
    D_C(\lambda^\star) - D_C(0)
    \;=\; a\,\lambda^\star + b\,(\lambda^\star)^2,
    \qquad
    a \;=\; 2\,\mathbb{E}_C\!\bigl[(g - \widehat{f}_T)\,h\bigr],
    \quad
    b \;=\; \mathbb{E}_C[h^2].
    \label{eq:excess-quadratic}
\end{equation}
Over the tested nonnegative $\lambda^\star$ range, increasing the
hidden-interaction amplitude increases realized prediction error
while the training data and fitted predictor remain unchanged
(same $D_T \Rightarrow$ same $\widehat{f}_T$; $D_C(\lambda^\star)
- D_C(0) = a\lambda^\star + b(\lambda^\star)^2$ with $b =
\mathbb{E}_C[h^2] > 0$). Monotonic increase for all
$\lambda^\star \ge 0$ requires $a \ge 0$ (in general the minimum sits
at $\lambda^\star = -a/(2b)$); the fitted seed-level $a$'s are
positive here (Table~\ref{tab:synth-quadratic}), so the observed
curves increase over the tested range. This is the ``agreement while
wrong'' side of Corollary~\ref{cor:blind-spot}.

\begin{table}[t]
\centering\small
\caption{Seed-level fit of the exact decomposition
$D_C(\lambda^\star) - D_C(0) = a \lambda^\star + b (\lambda^\star)^2$.
For each of $8$ seeds, $(a, b)$ are fitted jointly over the six
$\lambda^\star$ values. The table reports the median and $5$--$95\%$
range across the eight seed-level coefficient estimates.
Analytical prediction: $b = \mathbb{E}_C[h^2] = \ResultEAnalyticalEhSq$
for $q = 1$ and $X_A, X_B \sim \mathrm{Uniform}(0.3, 1)$. The linear
coefficient $a$ is a fixed-predictor cross-term reflecting the
correlation between the learner's background error $(g - \widehat{f}_T)$
and the hidden direction $h$ on $C$; positive $a$ suffices to explain
observed log-log slopes below $2$ in $|\lambda^\star|$ across the tested
finite amplitude range.}
\label{tab:synth-quadratic}
\begin{tabular}{lcc}
\toprule
Family      & Linear $a$ (median, $90\%$ pctile) & Quadratic $b$ (median, $90\%$ pctile) \\
\midrule
Restricted  & $\ResultEQuadRestrictedA$ $[\ResultEQuadRestrictedALo, \ResultEQuadRestrictedAHi]$
            & $\ResultEQuadRestrictedB$ $[\ResultEQuadRestrictedBLo, \ResultEQuadRestrictedBHi]$ \\
Rich-na\"ive & $\ResultEQuadRichNaiveA$ $[\ResultEQuadRichNaiveALo, \ResultEQuadRichNaiveAHi]$
            & $\ResultEQuadRichNaiveB$ $[\ResultEQuadRichNaiveBLo, \ResultEQuadRichNaiveBHi]$ \\
\bottomrule
\end{tabular}
\end{table}

Table~\ref{tab:synth-quadratic} shows the fit. The quadratic coefficient
$b$ matches the analytical $\mathbb{E}_C[h^2] = \ResultEAnalyticalEhSq$
to two decimals in both families (median $b = \ResultEQuadRestrictedB$
and $\ResultEQuadRichNaiveB$). The linear coefficient $a$ is positive in
both families (median $\ResultEQuadRestrictedA$, $\ResultEQuadRichNaiveA$)
: the learner's fixed background error is correlated with $h$ on $C$.
The exact quadratic decomposition of \eqref{eq:excess-quadratic} is
therefore fully consistent with the observed excess-error curves; the
log-log slope in $|\lambda^\star|$ is not the theoretically relevant
statistic.

\subsection{Finite ensembles can fail to track admissible
ambiguity}
\label{app:synth-ambiguity}

Under the fixed-data protocol of \S\ref{app:synth-setup}, with
$\lambda^\star = 0$ and every seed-level draw held fixed across
the $\Lambda$-grid, the truth sits inside every admissible class
and only the admissible ambiguity varies. The fixed-$g$
one-dimensional slice
$\mathcal{S}_\Lambda(g,h) = \{g + \lambda h : |\lambda| \le \Lambda\}$
has $L^2(P_C)$ diameter, computed directly,
\[
    \Delta^{\mathrm{slice}}(\Lambda)
    \;=\; \operatorname{diam}_{L^2(P_C)} \mathcal{S}_\Lambda(g,h)
    \;=\; 2\Lambda \sqrt{\mathbb{E}_C[h^2]},
    \qquad
    \Delta^{\mathrm{slice}}(\Lambda)^2
    \;=\; 4\Lambda^2\,\mathbb{E}_C[h^2],
\]
attained by the two extreme branches $g \pm \Lambda h$. In this
synthetic the admissible class is the linear span
$\mathcal{F} = \mathcal{G} \oplus \operatorname{span}(h)$ with
$\mathcal{N}_T = \operatorname{span}(h)$ one-dimensional, and we scale
the class norm so that $\|h\|_{\mathcal{F}} = 1$ (making $B_\Lambda$,
the radius-$\Lambda$ $\mathcal{F}$-ball, restrict the invisible
coefficient to $|\lambda| \le \Lambda$ on the line
$\operatorname{span}(h)$). Only under this coefficient-compatible
normalization is the local invisible ball
$\mathcal{N}_T \cap B_\Lambda$ exactly the segment
$\{\lambda h : |\lambda| \le \Lambda\}$, so that the slice diameter
coincides with the oracle-fibre diameter of Prop.~\ref{prop:diameter},
$\Delta^{\mathrm{slice}}(\Lambda) = \Delta^{\mathrm{orc}}(\Lambda)$; a
different scaling ($\|h\|_{\mathcal{F}} = c$) would rescale the ball's
coefficient cutoff to $|\lambda| \le \Lambda/c$ and the two diameters
would differ by $1/c$. For a higher-dimensional $\mathcal{N}_T$ the
slice diameter lower-bounds $\Delta^{\mathrm{orc}}$ provided the slice
perturbations $\{\lambda h\}$ lie in $B_\Lambda$.

Two direct consequences of the construction pin the entire result.

\paragraph{Covering ensemble attains the slice diameter.}
For $\mathcal{H}_{\mathrm{cover}}(\Lambda) = \{g - \Lambda h,\, g,\, g + \Lambda h\}$
the maximum pairwise squared distance is attained by the two
extremes and equals
\[
    \widehat{\Delta}^{2}_{\mathcal{H}_{\mathrm{cover}}, \mathrm{val}}
    \;=\; \|(g + \Lambda h) - (g - \Lambda h)\|_{L^2(P_C)}^2
    \;=\; 4\Lambda^2\,\mathbb{E}_C[h^2]
    \;=\; \Delta^{\mathrm{slice}}(\Lambda)^2
    \;=\; (\Delta^{\mathrm{orc}})^2(\Lambda),
\]
the last equality holding because $\mathcal{N}_T$ is one-dimensional
\emph{and} normalized ($\|h\|_{\mathcal{F}} = 1$) here, so the slice is
the full oracle fibre.
Empirically, the ratio between the sampled covering squared
diameter and this analytical value is $\ResultEDeltaRatio$ across
the $40$ (seed, $\Lambda$) cells, confirming that the
implementation and the analytical expression agree.

\paragraph{Restricted / rich-naïve ensembles are exactly
$\Lambda$-invariant.} Under the fixed-data protocol,
$D_T$, $C$, and every model random seed are drawn once and held
fixed across $\Lambda$. Restricted and rich-naïve members take
only $D_T$ and their own random seed as input: their fitted
predictors $\hat f_h$ therefore do not depend on $\Lambda$ at all.
Consequently the empirical squared diameter
$\widehat{\Delta}^{2}_{\mathcal{H}, \mathrm{val}}
= \max_{h_1, h_2} \|\hat f_{h_1} - \hat f_{h_2}\|_{L^2(P_C)}^2$
is a fixed function of the seed, exactly invariant in $\Lambda$;
its log-log slope in $\Lambda$ is $0$ by construction.

\paragraph{Summary of the mechanism.}
Putting the two consequences together:
\[
    \boxed{\;
    \Lambda \uparrow
    \;\Longrightarrow\;
    (\Delta^{\mathrm{orc}})^2 \uparrow \text{ at slope } 2,
    \quad \text{but} \quad
    \widehat{\Delta}^{2}_{\mathcal{H}, \mathrm{val}}
    \text{ stays fixed for restricted / rich-naïve}.
    \;}
\]
Figure~\ref{fig:synth-mechanism} plots the three curves against
the analytical $(\Delta^{\mathrm{orc}})^2 = 4\Lambda^2 \mathbb{E}_C[h^2]$
line: covering coincides with the analytical line; restricted and
rich-naïve are horizontal. The restricted and rich-naïve learners are
not shown to lie in Corollary~\ref{cor:blind-spot}'s named admissible
version space, so this experiment \emph{illustrates} the shared
blind-spot mechanism rather than literally instantiating the
corollary.

\subsection{Rank-space analogue}
\label{app:synth-rank}

Since the scientific endpoint is discovery ranking, we repeat the
audit with pairwise ordering disagreement, a rank-space analogue
of the practical Spearman disagreement
$V_{\mathcal{H}}^{\mathrm{avg,\,rank}}$ used in
Appendix~\ref{app:zerolabel}:
\[
    d_{\mathrm{ord}}(f_m, f_{m'})
    \;=\; \Pr_{x, x' \sim P_C \otimes P_C}
    \bigl[\, \operatorname{sign}(f_m(x) - f_m(x'))
    \neq \operatorname{sign}(f_{m'}(x) - f_{m'}(x')) \,\bigr],
\]
$V_{\mathcal{H}}^{\mathrm{avg,\,ord}}$ as the pairwise average and
$\widehat{\Delta}^{\mathrm{ord}}_{\mathcal{H}} =
\max_{h_1, h_2 \in \mathcal{H}} d_{\mathrm{ord}}(\hat f_{h_1}, \hat f_{h_2})$
as the finite-ensemble ordering diameter. Both are bounded in
$[0, 1]$; since the diameter is the maximum pairwise ordering
disagreement, $V_{\mathcal{H}}^{\mathrm{avg,\,ord}} \le
\widehat{\Delta}^{\mathrm{ord}}_{\mathcal{H}}$.

Under the fixed-data protocol every fitted member is identical
across $\Lambda$, so for restricted and rich-naïve families
every pairwise ordering disagreement is identical across $\Lambda$;
both $V_{\mathcal{H}}^{\mathrm{avg,\,ord}}$ and
$\widehat{\Delta}^{\mathrm{ord}}_{\mathcal{H}}$ are therefore
exactly invariant. For the covering ensemble the maximum member-pair
ordering diameter is attained by the two extreme OE branches
$g \pm \Lambda h$ and equals the analytical ordering diameter of the
fixed-$g$ slice, written $\Delta^{\mathrm{orc},\mathrm{ord}}(\Lambda)$
(the slice ordering diameter, which coincides with the oracle
fibre's because $\mathcal{N}_T$ is one-dimensional and normalized
$\|h\|_{\mathcal{F}} = 1$ here), so the
covering diameter provably grows with $\Lambda$.

\begin{table}[!ht]
\centering\small
\caption{Dependence of the rank-space disagreement statistics on
$\Lambda$ under the fixed-data protocol. For restricted and
rich-naïve families the fitted members do not depend on
$\Lambda$, so \emph{both} $V_{\mathcal{H}}^{\mathrm{avg,\,ord}}$
and $\widehat{\Delta}^{\mathrm{ord}}_{\mathcal{H}}$ are exactly
$\Lambda$-invariant. For the covering ensemble the diameter
$\widehat{\Delta}^{\mathrm{ord}}_{\mathcal{H}}$ equals the
analytical $\Delta^{\mathrm{orc},\mathrm{ord}}(\Lambda)$ by
construction and grows sublinearly with $\Lambda$ (bounded metric);
the covering average $V_{\mathcal{H}}^{\mathrm{avg,\,ord}}$
tracks the same growth. The covering-ensemble numerical slope of
$V_{\mathcal{H}}^{\mathrm{avg,\,ord}}$ is
$\ResultEVordSlopeCovering$
$[\ResultEVordSlopeCoveringLo, \ResultEVordSlopeCoveringHi]$ over
the tested $\Lambda$-grid.}
\label{tab:synth-rank}
\begin{tabular}{lcc}
\toprule
Family
    & $V_{\mathcal{H}}^{\mathrm{avg,\,ord}}$ vs $\Lambda$
    & $\widehat{\Delta}^{\mathrm{ord}}_{\mathcal{H}}$ vs $\Lambda$ \\
\midrule
Restricted    & exactly invariant & exactly invariant \\
Rich-naïve    & exactly invariant & exactly invariant \\
Covering      & grows sublinearly & $= \Delta^{\mathrm{orc},\mathrm{ord}}(\Lambda)$ (by construction) \\
\bottomrule
\end{tabular}
\end{table}

\emph{Takeaway.} Empirical ensemble
disagreement reflects the portion of the admissible variation
represented by the chosen ensemble; it can therefore remain nearly
unchanged while the full admissible ambiguity grows. This is
exactly the mechanism Corollary~\ref{cor:blind-spot} permits.

\emph{What this does not prove.} This synthetic establishes the
mechanism permitted by Corollary~\ref{cor:blind-spot}. It does not
establish that the real-material failure in
Appendix~\ref{app:zerolabel} is caused by the same hidden
direction, nor does it identify the chemical form of that
direction.

\subsection{Basis-misspecification sweep: overlap with the true invisible space}
\label{app:synth-misspec}

The real-data experiment (\S\ref{sec:real-setup}) uses a structural subspace of
the empirical kernel, not necessarily its full kernel, so we validate the
mechanism directly: does prediction improve as the design basis overlaps more
with the \emph{true} invisible space? On a $D=6$ simplex with a known true
invisible subspace $\Phi_\star$ ($r=6$ pair-gated monomials vanishing on the
$A$-face support) and an orthogonal nuisance library $\Psi$, we build design
bases with $k$ correct directions ($k$ of $\Phi_\star$ plus $r-k$ of $\Psi$),
run the same regularized $I$-opt pipeline (select, fit, predict in that basis),
and measure held-out MSE ($30$ seeds; script
\texttt{analysis\_synth\_basis\_misspec.py}, data
\texttt{results/synth\_basis\_misspec.csv}).

Prediction improves \emph{monotonically} with $k$: at $m = 15$ the fully
misspecified basis ($k = 0$) has $\ResultSynthMisspecRatio\times$ the MSE of the
correctly specified basis ($k = r$), decreasing at every intermediate $k$. A
continuous principal-angle variant $\Phi_\alpha = \cos\alpha\,\Phi_\star +
\sin\alpha\,\Psi$ shows the benefit is also \emph{sensitive to angular
misspecification}: a $15^\circ$ rotation already forfeits most of the advantage.
So more overlap between the proposed and true invisible subspaces yields better
few-shot prediction, and the advantage can be sharp in the alignment, directly
validating the representation effect that dominates the electrocatalysis
factorial (Table~\ref{tab:realdata-factorial}), rather than only showing that one
handcrafted basis works.

\section{Sparse Joint Labels Enable Pair-Level Routing}
\label{app:calibration}

\emph{Question and role.} Proposition~\ref{prop:few-shot} gives a
stylised recovery condition: known invisible basis $\phi$, known
background $g$, well-conditioned copresence design, OLS. We test a
weaker operational consequence: given a pair $(A, B)$, can $k$
random copresence labels estimate held-out pair difficulty well
enough to route future measurement, without claiming to recover
$h$ or to close a version-space diameter?
Appendix~\ref{app:zerolabel} showed that zero-label ensemble
disagreement fails at this task; here we ask what a small number
of joint labels can and cannot buy.

\subsection{Held-out target}
\label{sec:cal-target}

For each system $s$, each pair $(A, B)$, and each draw $t$,
sample
\[
    \mathcal{S}_{AB}(s, t) \;\subset\; C_{AB}, \qquad
    |\mathcal{S}_{AB}| = k,
\]
uniformly at random. The \emph{diagnostic} is computed only from
labels in $\mathcal{S}_{AB}$; the \emph{target} is pair difficulty
on the untouched remaining region
\[
    C_{AB}^{\mathrm{hold}}(s, t)
    \;=\; C_{AB} \setminus \mathcal{S}_{AB}(s, t).
\]
We adopt a leakage-safe single-model target defined by a
Mat\'ern-$5/2$ GP $\hat f_{\mathrm{Mat}, T}$ (matching one member of
$\mathcal{G}_1$; see \S\ref{sec:cal-loso}):
\[
    D_{\mathrm{test}}^{\mathrm{hold}}(A, B; s, t)
    \;:=\; 1 - \rho_S\!\left(\, y,\; \hat f_{\mathrm{Mat}, T} \,\right)
    \Big|_{C_{AB}^{\mathrm{hold}}(s, t)}.
\]
The leakage boundary is thus algebraic:
$y(\mathcal{S}_{AB}) \to$ probe;
$y(C_{AB} \setminus \mathcal{S}_{AB}) \to$ target.

\subsection{Two probes on materials}
\label{sec:cal-probes}

We define two primary materials probes on the calibration set
$\mathcal{S}_{AB}$; the singles-based probe used for proteins is
deferred to \S\ref{sec:cal-crossdomain}.

\paragraph{Response-only variance (no compositional information).}
\[
    s_{\mathrm{var}}(\mathcal{S}_{AB})
    \;=\; \mathrm{Var}_{x \in \mathcal{S}_{AB}}[\, y(x) \,].
\]
This uses only the labels in $\mathcal{S}_{AB}$; the compositions
$x_i$ are not consulted at all. It is the ``the labels themselves do the work'' baseline.

\paragraph{Model-aware GP calibration.}
Let $\hat f_{\mathrm{RBF}, T}$ be an RBF-kernel GP fit on $T_{AB}$
(a $\mathcal{G}_2$ member, disjoint from the target family
$\mathcal{G}_1$ used in \S\ref{sec:cal-loso}). Define
\[
    s_{\mathrm{GPcal}}(\mathcal{S}_{AB})
    \;=\; \frac{1}{|\mathcal{S}_{AB}|}
    \sum_{x \in \mathcal{S}_{AB}}
    \bigl(\, y(x) - \hat f_{\mathrm{RBF}, T}(x) \,\bigr)^2.
\]
This is the mean squared residual of the disjoint-family GP
background on the calibration points: a natural per-pair
mismatch statistic.

Trivial calibration-set summaries (range, IQR, standard deviation,
MAD) are reported as controls in the supplementary tables; none
beats $s_{\mathrm{var}}$ or $s_{\mathrm{GPcal}}$ on the held-out
Spearman.

\subsection{Leakage-safe LOSO protocol}
\label{sec:cal-loso}

We evaluate each probe under the same protocol as
Appendix~\ref{app:zerolabel-loso}. Partition the model library
into disjoint families
\[
    \mathcal{G}_1 = \{\text{Mat\'ern GP},\ \text{RF},\ \text{GBM}\},
    \qquad
    \mathcal{G}_2 = \{\text{RBF GP},\ \text{additive GP},\ \text{MLP},\ \text{polynomial ridge}\},
\]
with the target $D_{\mathrm{test}}^{\mathrm{hold}}$ defined by the
$\mathcal{G}_1$ Mat\'ern GP $\hat f_{\mathrm{Mat}, T}$ as in
\S\ref{sec:cal-target}. For GP-calibration the fitted background
$\hat f_{\mathrm{RBF}, T}$ is drawn from $\mathcal{G}_2$ (the
RBF-kernel GP), so target and probe share no model families;
$s_{\mathrm{GPcal}}$ from \S\ref{sec:cal-probes} thus reads
\[
    s_{\mathrm{GPcal}}(\mathcal{S}_{AB})
    \;=\; \frac{1}{|\mathcal{S}_{AB}|}
    \sum_{x \in \mathcal{S}_{AB}}
    \bigl(\, y(x) - \hat f_{\mathrm{RBF}, T}(x) \,\bigr)^2.
\]
Response-only variance uses no fitted model on the probe side.
The combined \emph{variance $+$ GP-cal} predictor is the two-feature
LOSO regression
$D_{\mathrm{test}}^{\mathrm{hold}} \sim \beta_0 + \beta_1
s_{\mathrm{var}} + \beta_2 s_{\mathrm{GPcal}}$, with $\beta$ fit
on the pairs of the three training systems only.

For each of $50$ draws of $\mathcal{S}_{AB}$ per pair, fit a
linear mapping $D_{\mathrm{test}}^{\mathrm{hold}} \sim
s_{\mathrm{probe}}$ on the pairs of three systems and predict
$D_{\mathrm{test}}^{\mathrm{hold}}$ on the held-out fourth.
Report the median held-out $R^2$ and Spearman across the four
systems and $50$ draws, at $k = 20$.

\subsection{Held-out results: rank transfers, calibration does not}
\label{sec:cal-loso-results}

A matched-target head-to-head against
$D_{\mathrm{test}}^{\mathrm{hold}}$ under the same LOSO protocol
(each row predicts the same held-out target from
\S\ref{sec:cal-target}; the zero-label row uses the disjoint
$\mathcal{G}_2$ ensemble disagreement
$V_{\mathcal{H}}^{\mathrm{avg,\,rank}}(\mathcal{G}_2)$ from
Appendix~\ref{app:zerolabel-cross} as its predictor):
\[
    \begin{array}{lcc}
        \toprule
        \text{Signal} & R^2_{\mathrm{LOSO}} & \rho_{S,\mathrm{LOSO}} \\
        \midrule
        \text{Zero-label disagreement}    & \ResultDLOSOMatchedRsq & \ResultDLOSOMatchedSp \\
        \text{Response variance ($k = 20$)}   & \ResultLOSOVarianceOnly & +\ResultLOSOVarianceOnlySpMed \\
        \text{Model-aware GP-cal ($k = 20$)}  & \ResultLOSOGPcalOnly & +\ResultLOSOGPcalOnlySpMed \\
        \text{Variance $+$ GP-cal ($k = 20$)} & \ResultLOSOVarianceGPcal & +\ResultLOSOVarianceGPcalSpMed \\
        \bottomrule
    \end{array}
\]
The interpretation:
\[
    \boxed{\;
    R^2_{\mathrm{LOSO}} \approx 0,
    \quad \rho_{S,\mathrm{LOSO}} \approx +\ResultLOSOVarianceOnlySpMed:
    \text{ sparse labels transfer pair ordering, not absolute difficulty scale.}
    \;}
\]
Two consequences deserve emphasis.
The near-zero held-out $R^2$ is informative: the absolute
difficulty scale does not transfer across systems, whereas
relative pair ordering does.
Under the leakage-safe protocol, response-only variance and
model-aware GP-cal have nearly equal held-out Spearman point
estimates ($+\ResultLOSOVarianceOnlySpMed$ vs
$+\ResultLOSOGPcalOnlySpMed$; we report point estimates without a
paired equivalence margin, so this is ``nearly equal,'' not
statistical equivalence). Most of the transferable rank
signal is already present in the labels' own dispersion on
$\mathcal{S}_{AB}$; adding a model-aware residual over a fitted
RBF-GP background does not measurably improve the rank point
estimate over the labels themselves at $k = 20$ (point estimates are
close under this protocol; no equivalence test was performed). This fits the paper's thesis: joint
labels reveal pair difficulty without needing an assumed invisible
basis.

\subsection{Three decisions: routing $\ne$ within-pair acquisition
$\ne$ global pair-map allocation}
\label{sec:cal-selective}

\begin{figure}[!ht]\centering
\includegraphics[width=0.98\linewidth]{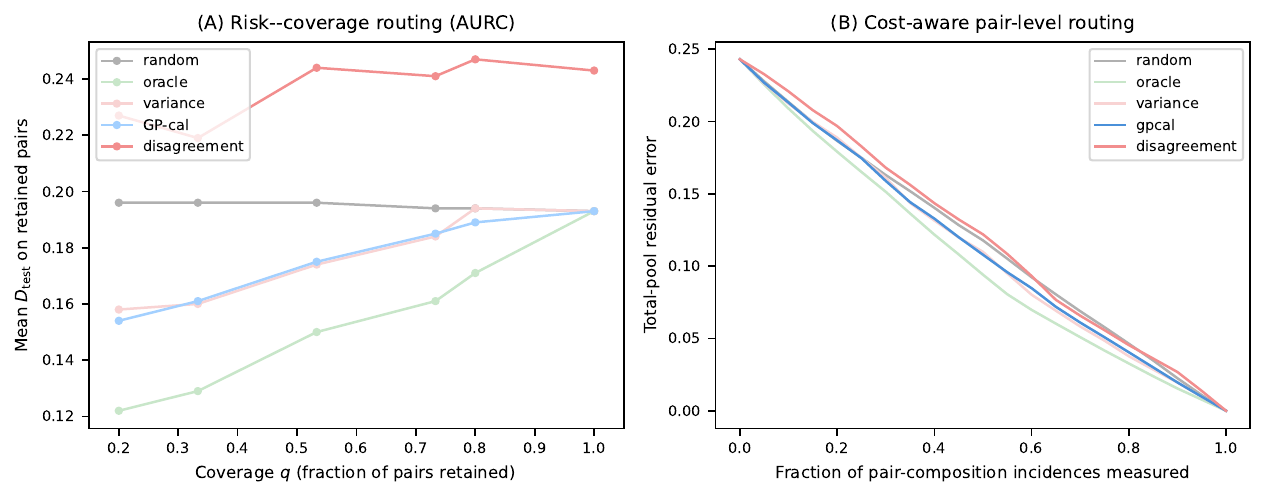}
\caption{\textbf{Sparse-label pair-level routing.}
(A) Risk--coverage: retain the fraction $q$ of pairs with the
lowest predicted $D$, and report the mean untouched-region
difficulty on the retained set. Oracle (green) uses the true
$D_{\mathrm{test}}^{\mathrm{hold}}$. Both sparse-label signals
(variance, GP-cal) close a substantial fraction of the
random-to-oracle gap; zero-label disagreement (red) is worse
than random. (B) Cost-aware routing: budget axis is the fraction of
\emph{pair-composition incidences} $\sum_{\mathrm{routed}}|C_{AB}| /
\sum_{\mathrm{all}}|C_{AB}|$ vs total-pool residual error
$E(b) = B_{\text{pool}}^{-1} \sum_{(AB,x) \notin R_b} e_{AB,x}$ with
$B_{\text{pool}} = \sum_{AB}|C_{AB}| = 39{,}413$ the fixed incidence
count (routed incidences contribute $0$, so $E(1) = 0$; this is a
total-pool sum over a fixed denominator, not a conditional mean over
the shrinking un-routed set);
response variance and GP-cal are essentially tied at the median.
Because a ternary composition belongs to $3$ pairs and a quaternary
to $6$, this sum exceeds the number of unique compositions by
$\approx 4.7\times$ on this pool, so the axis measures per-pair
measurement budget, not distinct compositions.}
\label{fig:routing}
\end{figure}

Pair difficulty is used in three distinct downstream decisions.
The rewritten Appendix conflates none of them:
\begin{itemize}[leftmargin=1.6em,itemsep=1pt]
    \item \emph{Pair routing:} which pairs require more
        measurement? Predict vs measure at pair granularity;
        driven by $D_{\mathrm{test}}^{\mathrm{hold}}$ ranking.
    \item \emph{Within-pair acquisition:} which
        $x \in C_{AB}$ should we measure given a chosen pair?
        Addressed in \S\ref{sec:real-setup} by
        identifiability-directed design on $\mathcal{N}_T$.
    \item \emph{Global pair-map allocation:} given a finite
        total measurement budget, how do we distribute it across
        pairs to find globally best catalysts? Different
        objective, since high-FOM catalysts can concentrate in
        difficult pairs.
\end{itemize}

\paragraph{Routing: sparse-label signals close roughly half of
the random-to-oracle gap.}
On a held-out system, we fit the LOSO probe$\to$difficulty
mapping (\S\ref{sec:cal-loso}) on the other three systems to
score each held-out pair by predicted difficulty
$\widehat D(A, B)$. With $P = 15$ pairs per system, let
$\mathcal{P}_q$ be the $\lceil q P \rceil$ pairs with the lowest
$\widehat D$ (retained); the remaining $P - \lceil q P \rceil$
are routed to measurement. The \emph{realized} coverage is
$q^\star = \lceil 15q \rceil / 15 \in
\{0.200, 0.333, 0.533, 0.733, 0.800, 1.000\}$, and the routing risk is
\[
    R(q^\star) \;=\; \frac{1}{|\mathcal{P}_q|}
    \sum_{(A, B) \in \mathcal{P}_q}
    D_{\mathrm{test}}^{\mathrm{hold}}(A, B).
\]
We report the \emph{unnormalized partial AURC}
$\mathrm{pAURC}_{[0.2,1]} = \sum_i \tfrac{q^\star_{i+1}-q^\star_i}{2}
[R(q^\star_i) + R(q^\star_{i+1})]$, i.e.\ the trapezoidal area under
the median $R(q^\star)$ curve of Figure~\ref{fig:routing}A on the
realized grid (area under the pointwise-median curve, aggregated
across $50$ draws $\times$ four held-out systems; not the median of
per-draw areas). Under the disjoint-family protocol, this
$\mathrm{pAURC}$ is $\ResultAURCRandom$ (random) and drops to
$\ResultAURCVariance$ (variance) and $\ResultAURCGPcal$ (GP-cal),
with $\ResultAURCOracle$ (oracle). Variance and GP-cal close roughly two-fifths to one-half of the
random-to-oracle gap ($\ResultAURCVarianceGapClosed\%$ and
$\ResultAURCGPcalGapClosed\%$ respectively on unrounded pAURC; the
displayed $3$-decimal scalars give about $42$--$46\%$). Their point
estimates are close under this protocol (no equivalence test was
performed), and combining them ($\ResultAURCVarGPcal$) changes the
point estimate by only $0.001$--$0.002$ relative to either signal
alone; practical and statistical significance of that difference were
not assessed (no equivalence margin specified).
Zero-label ensemble disagreement is worse than random
($\ResultAURCDisagreement$).

Figure~\ref{fig:routing}B is an \emph{idealized} cost-aware
routing proxy. We charge budget in \emph{pair-composition
incidences} $B_{\text{pool}} = \sum_{AB} |C_{AB}|$; on this pool
$B_{\text{pool}} = 39{,}413$, whereas the number of \emph{unique}
multi-element compositions is
$B_{\text{unique}} = |\bigcup_{AB} C_{AB}| = 8{,}458$, a ratio of
$4.66\times$ (a composition of order $n$ contributes $\binom{n}{2}$
incidences). The incidence axis therefore overstates the physical
number of experiments, since one composition can inform several
pairs; the union cost $B_{\text{unique}}(R) = |\bigcup_{AB \in R}
C_{AB}|$ is the physical count. At budget-fraction
$b \in [0, 1]$, the $\lceil b \cdot B_{\text{pool}} \rceil$
incidences with the highest $\widehat D$ are routed and assigned
zero error; the remaining use GP predictions. The $y$-axis is the
total-pool risk
\[
    E(b) = \frac{1}{B_{\text{pool}}}
    \sum_{(AB, x) \notin R_b} e_{AB, x},
    \qquad
    B_{\text{pool}} = \sum_{AB} |C_{AB}|,
\]
where $e_{AB, x} = (\widehat f(x) - y(x))^2$ is the squared GP error
on incidence $(AB, x)$ and the fixed denominator is the total
incidence count $B_{\text{pool}} = 39{,}413$ (routed incidences
contribute $0$, so $E(1) = 0$ by construction). This is a total-pool
sum over a fixed denominator, not a conditional mean over the
shrinking un-routed set, and is a retrospective proxy for the routing
benefit, not a prospective acquisition simulation charged in unique
compositions.

\paragraph{Global pair-map allocation: the routing signal is not
an acquisition score.}
Here top-$5\%$ recall is
$\mathrm{Recall@}5\% = |M \cap T_{5\%}| / |T_{5\%}|$, where
$T_{5\%}$ is the true top-$5\%$-FOM set of compositions in the
held-out system (median $|T_{5\%}| \approx 106$; ties broken by
composition index) and $M$ the set of \emph{measured} compositions.
The budget is \emph{capped} at $B_{\max} = k_{\text{unif}} \cdot
15 = 300$ pair-composition incidences: after a uniform $k_0 = 5$-per-pair
initialization ($75$ incidences), the remaining $225$ are allocated by the
policy: the uniform baseline spreads them evenly, while adaptive
policies give the predicted-easy half of pairs the extra budget and
leave the predicted-hard half at $k_0$. The realized per-policy
incidence count $B_{\mathrm{policy}}$ is $300$ for uniform but $299$
for the adaptive policies (the integer rule below drops one), so the
budgets are near-equal but not literally identical; we therefore
report realized costs rather than call them equal-cost.
\emph{Budget accounting (a caveat).} The budget is counted in
\emph{pair-composition incidences}: a composition participating in
several pairs (a ternary or quaternary lying in multiple $C_{AB}$) is
charged once per pair it is measured for, whereas the recall set $M$
counts \emph{unique} compositions. \emph{Allocation rule.} The $15$
pairs are split at the median predicted difficulty into
$n_{\mathrm{easy}} = \lfloor 15/2 \rfloor = 7$ ``easy'' pairs and $8$
``hard'' pairs; the $225$ post-init incidences are divided as
$k_{\mathrm{easy}} = k_0 + \lfloor 225/7 \rfloor = 37$ per easy pair,
hard pairs staying at $k_0 = 5$, giving
$7 \times 37 + 8 \times 5 = 299$ incidences (the floor drops one, so
adaptive policies realize $B_{\mathrm{policy}} = 299$ vs uniform's
$300$; a prespecified deterministic tie rule could assign the leftover
incidence to restore exactly $300$, which we leave to a follow-up).
\emph{Unique-composition cost.} Because policies concentrate budget on
different pairs, they can select different overlaps and hence different
numbers of unique measured compositions $|M|$ at the same incidence
budget. Deduplicating measured compositions as selections arrive, the
mean unique count is $B_{\mathrm{unique}} = |M| \in [279.7, 281.4]$
across the seven policies ($93.6$--$94.0\%$ of each policy's incidence
budget: uniform $281.4/300 = 93.8\%$; adaptive-GP-cal
$280.6/299 = 93.9\%$), a spread below $2$ compositions. \emph{This
does not by itself prove the ordering survives an exactly-fixed unique
budget}: with a top-$5\%$ set of $\approx 106$ compositions, one extra
top-set hit shifts Recall@$5\%$ by $1/106 \approx 0.0094$, comparable
to the $0.010$ uniform-vs-GP-cal gap, so neither the $1$-incidence
shortfall nor the range of \emph{average} unique costs can be dismissed
on magnitude: either could move a policy by one hit. We therefore
confine the conclusion below to this \emph{near-equal incidence}
setting (cap $300$; realized $299$/$300$) and report the unique counts
only as a descriptive check; a rerun at exactly matched incidence
\emph{and} unique-composition budget (deterministic tie rule plus
deduplication as selections arrive) is left as the cleaner test.
We test whether the same predicted-difficulty signal can steer an
adaptive budget toward predicted-easy pairs on the pair-map
top-$5\%$-FOM recall problem, under the same disjoint-family
protocol (target GP = Mat\'ern, probe GP = RBF). Under the capped
budget ($B_{\max} = 300$, realized $299$/$300$) with a $k_0 = 5$ init
per pair, adaptive-GP-cal top-$5\%$
recall is $+0.132$ vs uniform $+0.142$ (paired Wilcoxon
signed-rank on $n = 120$ seed$\times$system differences, median
paired $\Delta = 0.000$, $p = 0.75$, n.s.); adaptive-variance is
\emph{worse} at $+0.127$ ($\Delta = -0.009$, $p = 4 \times 10^{-4}$);
the combined variance$+$GP-cal policy is $+0.132$
($\Delta = -0.009$, $p = 1.5 \times 10^{-3}$, worse than uniform).
Adaptive-oracle (using the true $D_{\mathrm{test}}$ ranking to
allocate) is essentially tied with uniform at $+0.142$
($\Delta = 0.000$, $p = 0.05$).
These $p$-values quantify seed-level replication conditional on the
four fixed systems (30 seeds each), not population inference across
systems, of which there are only four. Treating the system as the
unit of analysis, the effect is heterogeneous:
adaptive-variance is nonpositive in all four systems (per-system mean
$\Delta$ from $-0.002$ to $-0.031$), whereas the oracle effect is
positive in two systems and negative in two. With only four systems
these system-level comparisons are
\emph{descriptive} (a nominal unadjusted one-sample $t(3)$ on the
four per-system effects gives $p = 0.12$ for variance, $0.60$ for
oracle, and $0.25$ for variance$+$GP-cal, but $n = 4$ and no
multiplicity plan make these indicative, not confirmatory; even the
smallest exact two-sided sign-test value achievable with four
same-direction effects is $2/2^4 = 0.125$). All adaptive policies correctly
reduce weighted mean prediction difficulty on the un-routed pool
(from $+0.192$ under uniform to $+0.156$ under oracle), so the
difficulty rankings themselves are useful. What the experiment
therefore shows is that \emph{difficulty-based allocation, even with
oracle difficulty, reduces prediction risk while not improving (and
often worsening) top-FOM recall}, so pair-difficulty minimization
and extreme-property discovery are misaligned objectives on this
benchmark (\emph{adaptive variance} is nonpositive in all four
systems; the other policies, especially oracle, are heterogeneous,
with the oracle effect positive in two systems and negative in two;
with only four systems and no multiplicity plan, population-level
evidence is inconclusive).
This is consistent with top-FOM
candidates being disproportionately located in harder pair
regions. Per-pair routing and global-budget pair-map allocation
are distinct objectives, and the same statistic (even an
oracle statistic) need not serve both.

\subsection{Cross-domain extension on proteins}
\label{sec:proteins-additive}
\label{sec:cal-crossdomain}

Protein DMS landscapes admit a canonical singles basis, so a
natural additive-residual probe
\[
    s_{\mathrm{Eadd}}(\mathcal{S}_{AB})
    \;=\; \tfrac{1}{|\mathcal{S}_{AB}|}
    \sum_{x \in \mathcal{S}_{AB}}
    (y(x) - \hat y_{\mathrm{singles}}(x))^2,
    \qquad
    \hat y_{\mathrm{singles}}(i{:}a, j{:}b)
    \;=\; y(i{:}a) + y(j{:}b) - y_{\mathrm{WT}},
\]
is available in that domain but has no canonical materials
analogue. The singles sum is additive only on the assay's reported
(log-enrichment / additive-fitness) scale, on which $y(\cdot)$ and
$y_{\mathrm{WT}}$ are defined; we use $y$ verbatim on that
transformed scale. On the retained protein panel
(Appendix~\ref{app:crossdomain}), $s_{\mathrm{Eadd}}$ reaches
pooled within-benchmark Spearman $\rho_S \approx 0.75$ against
$D_{\mathrm{cons}}$, competitive with $s_{\mathrm{GPcal}}$ at the
same $k$. The protein panel is not large enough to support the
analogous cross-domain LOSO transfer test used in
\S\ref{sec:cal-loso}, so we present this as a complementary
within-benchmark extension in a domain with a canonical singles-based
additive reference, not as a full replication of the materials
result.

\emph{Takeaway.} At $k = 20$ (roughly $3\%$ of a typical
$|C_{AB}| \approx 656$-composition pool), sparse joint labels
provide a useful transferable routing signal on the
electrocatalysis benchmark, closing roughly half of the
random-to-oracle risk--coverage AURC gap while remaining poorly
calibrated in absolute difficulty. Under the leakage-safe
protocol, GP calibration alone has a rank point estimate close to
response variance ($\rho_S = +0.49$ vs $+0.48$; point estimates
close, no equivalence test); combining the two modestly
increases held-out rank correlation to $+0.54$ but yields no
measurable additional risk--coverage improvement over either signal
alone (point estimates close; no equivalence margin specified). The signal
supports pair routing but is not a substitute for within-pair
acquisition (\S\ref{sec:real-setup}) or a global pair-map
allocation rule; those are distinct decisions and the same
statistic need not serve all three. Detailed per-system tables,
sample-complexity sweeps, and the online pair-map experiment are
in the supplementary code and cached results.

\section{Cross-Domain Synthesis}
\label{app:crossdomain}

This appendix synthesises what the materials and protein experiments do
and do not jointly establish. It is deliberately not a further experiment
section: the underlying results live in \S\S\ref{app:stability}--\ref{app:calibration}
(materials) and Table~\ref{tab:crossdomain-protein-breakdown} (proteins),
and we do not restate their numbers here beyond what is needed to draw
the boundary of the claim.

\paragraph{Shared structure.}
Three themes recur across both substrates, at different evidential
strength. First, the missing-support geometry itself: in
electrocatalysis, pair copresence is rare relative to the full simplex;
in DMS panels, double mutants at any two positions are rare relative to
the full doubles combinatorial. The framework of
\S\ref{sec:theory} names this common structural fact rather than a
species-specific accident. Second, pair difficulty is a reproducible
empirical object above the copresence gate: cross-learner concordance
$W \ge 0.80$ in every catalyst system and
$W_{\mathrm{pool}} = \ResultAProteinsKendallW$ across the
retained DMS panel (Table~\ref{tab:crossdomain-protein-breakdown}).
Third, in both domains, sparse joint observations inside a pair's
copresence region carry information about that pair's difficulty; only
materials currently establishes out-of-system transfer of that signal
to the untouched region ($D_{\mathrm{test}}^{\mathrm{hold}}$, LOSO
$\rho_S \approx +0.48$--$+0.54$; Appendix~\ref{app:calibration}), while
proteins are established on the within-benchmark pool
($E_{\mathrm{add}}$ singles-derived probe, pooled
$\rho_S = \ResultProbeEepiProteinsSpearman$;
\S\ref{sec:proteins-additive}). These are two different experiments,
and the paper does not claim they are the same.

\paragraph{Domain-specific realization.}
Materials lack a canonical singles-derived additive reference: the
degree-$\le 3$ null-space basis used in \S\ref{sec:real-setup} is
therefore a structural prior rather than a known ground-truth
interaction basis. Direct GP calibration on $\mathcal{S}_{AB}$
sidesteps the basis specification; at $k = 20$, GP calibration alone
has a rank point estimate close to response variance
($\rho_S = +0.49$ vs $+0.48$; no equivalence test), while combining
the two modestly increases held-out rank correlation to $+0.54$ but
yields no measurable additional risk--coverage improvement over
either signal alone. Proteins do have a canonical basis: the singles-fitness sum is
a physically meaningful additive reference, and the additive-residual
probe matches or exceeds direct GP calibration on the pooled
within-benchmark association. The same \emph{diagnostic} principle
(sparse joint observations reveal information about pair difficulty)
is supported in both domains; decision-level pair-map routing is
demonstrated only on materials.

\paragraph{Per-DMS breakdown of pair-difficulty stability.}
The stratified macro-summary is $0.77$, the equal-$n$ mean of the three
per-DMS concordances $\ResultAKendallGBOne$ (GB1),
$\ResultAKendallGRBTwo$ (GRB2), and $\ResultAKendallYAPOne$
(YAP1); we treat this as the headline protein
concordance. The \emph{pooled} value
$W_{\mathrm{pool}} = \ResultAProteinsKendallW$ is a \emph{different
estimand}: it ranks all $120$ cross-assay position-pairs together
(40 per DMS) with average-rank tie handling before computing a single
Kendall's $W$ across $\{$GP, RF, GBM$\}$. Global ranking across
assays of differing difficulty scale depresses it relative to the
stratified mean, so we retain $\ResultAProteinsKendallW$ only with
this global-ranking preprocessing and tie rule stated.
Table~\ref{tab:crossdomain-protein-breakdown} reports the per-DMS
numbers.

\begin{table}[!ht]
\centering\small
\caption{Per-DMS cross-learner concordance on the retained protein
DMS panel (40 position-pairs per DMS, 120 total). Kendall's $W$ is
computed separately per DMS across $\{$GP, RF, GBM$\}$ on the pairs
above the $n_{\mathrm{copresence}} \ge 25$ gate; the pooled
panel-level value (all 120 pairs) is
$W_{\mathrm{pool}} = \ResultAProteinsKendallW$.}
\label{tab:crossdomain-protein-breakdown}
\begin{tabular}{lcl}
\toprule
DMS & Cross-learner $W$ & Note \\
\midrule
SPG1-GB1 (Olson) & $\ResultAKendallGBOne$ & weaker learner agreement \\
GRB2 (Faure) & $\ResultAKendallGRBTwo$ & materials-scale learner agreement \\
YAP1 \citep{araya2012fundamental} & $\ResultAKendallYAPOne$ & learner-stable; rank split-half $r \approx \ResultBYAPWeakR$ \\
A4-Seuma & excluded & per-position-pair copresence below $n_{\mathrm{copresence}} \ge 25$ gate \\
\bottomrule
\end{tabular}
\end{table}

Substrate-level asymmetry (GB1's weaker learner agreement; YAP1's
learner-stability but weaker rank split-half reproducibility) is a
\emph{plausible} explanation for why $W_{\mathrm{pool}} =
\ResultAProteinsKendallW$ sits between materials-scale strength
($W \ge 0.80$ in every catalyst system) and near-null; global ranking
across assays of differing scale and this substrate asymmetry are not
separately identified by the two summaries ($0.77$ and $0.69$) alone.
Descriptively, in the protein panel GP disagrees with the tree
ensembles more strongly than the trees disagree with each other, so
the reduction is a substrate-level pattern rather than a global
instability.

\begin{table}[t]
\centering\small
\caption{Summary of cross-domain differences. The evaluation level is
deliberately made explicit in each cell: some rows compare LOSO (out-of-system)
transfer on materials against within-benchmark pooling on proteins,
which are not the same test. The missing-support geometry is shared;
zero-label disagreement transfer, concentration modulation, and the
LOSO cost-aware routing evaluation are established only on materials.}
\label{tab:crossdomain}
\begin{tabular}{@{}p{4.3cm}p{4.6cm}p{4.6cm}@{}}
\toprule
\textbf{Property}                             & \textbf{Materials (electrocatalysis)}                 & \textbf{Proteins (DMS)} \\
\midrule
Missing support                                & pair copresence rare                                    & pair copresence rare \\
Cross-learner $W$                              & $W \ge 0.80$ in every system (pooled $\ResultAMaterialsKendallW$)  & per-DMS $\ResultAKendallGBOne / \ResultAKendallGRBTwo / \ResultAKendallYAPOne$; pooled panel-level $\ResultAProteinsKendallW$ \\
Zero-label disagreement (LOSO)                 & fails (system~B negative)                              & no analogous LOSO run \\
Natural additive basis                         & none canonical                                          & singles-fitness sum \\
Sparse-label probe (evaluation level)          & variance / GP-cal / var$+$GP-cal, LOSO to $D_{\mathrm{test}}^{\mathrm{hold}}$, $\rho_S = +0.48 / +0.49 / +0.54$ & $E_{\mathrm{add}}$ (singles-sum residual), within-benchmark pool, $\rho_S = \ResultProbeEepiProteinsSpearman$ \\
Concentration modulation                       & consistent with a low tertile VPC ($= \ResultBICCTertileAnova$, LOEO $[\ResultBICCTertileJackLo, \ResultBICCTertileJackHi]$) & not measured \\
\bottomrule
\end{tabular}
\end{table}

\paragraph{Unequal strength of evidence.}
The materials evidence is stronger along three axes that matter for
the paper's central claims. Cross-system generalisation is tested by
LOSO across four independently synthesised electrocatalyst panels
(A, B, C, D), so both the zero-label failure (\S\ref{app:zerolabel})
and the sparse-label recovery (\S\ref{app:calibration}) are checked
out-of-distribution in the strongest sense available on this
substrate. Concentration modulation is measurable in composition
space (low tertile VPC $= \ResultBICCTertileAnova$, LOEO
$[\ResultBICCTertileJackLo, \ResultBICCTertileJackHi]$) and cannot be
formulated on discrete DMS panels. The pair-level routing experiment
(Figure~\ref{fig:routing}, \S\ref{sec:cal-selective}) also runs only
on materials. On the protein side, all analyses are within-benchmark;
DMS heterogeneity is substantial
($W = \ResultAKendallGBOne$ on GB1,
$\ResultAKendallGRBTwo$ on GRB2, $\ResultAKendallYAPOne$ on YAP1); and
the substrate panel is too small to run LOSO across it. Where
materials give four out-of-system transfers, proteins give three
within-benchmark replications.

\paragraph{What has not generalized.}
Three specific results are established on only one domain and should
not be read as jointly claimed. The zero-label disagreement failure
(negative on system B; \S\ref{app:zerolabel}) is a materials-only
finding; nothing in the protein panel tests it. Concentration
modulation is likewise materials-only. The pair-level routing and
global pair-map allocation experiments
(Appendix~\ref{app:calibration}, \S\ref{sec:cal-selective}) are
materials-only; the analogous within-pair acquisition experiment is
reported in \S\ref{sec:real-setup}. No analogous protein
routing or allocation at comparable scale has been run.

\paragraph{Boundary of the claim.}
Zero-label failure, that missing joint support cannot be diagnosed
reliably from non-copresence evidence alone, is established only on
the four materials panels. The theoretical non-implication is
domain-general, but the empirical failure is not yet cross-domain: in
proteins the current evidence shows only that sparse joint labels
correlate with within-benchmark difficulty, and it does not test
whether a protein zero-label diagnostic could succeed. What is shared
across substrates is the positive lesson that obtaining even sparse
joint observations can reveal pair-level difficulty. Which statistic
should then summarize
those observations is substrate-specific (response variance and GP
calibration on materials, singles-sum residual on proteins), and how
those observations should be translated into a routing or allocation
policy is a further question that the materials experiments answer
partially (routing yes; adaptive global allocation no) and the protein
experiments do not test. We do not claim a universal zero-label
detector, a universal pair-difficulty statistic, or a universal
routing policy. Nor do we claim the results extend to biological
substrates whose fitness landscape topology differs substantially
from the retained DMS panel (heavy assay-floor clamping, narrower
fitness ranges, or single-pair epistasis topology), nor to material
families whose composition space does not admit a copresence gate at
the pair level (single-element endmembers, small stoichiometry
counts).


\section{Materials Project design-selection appendix (MACE-MP)}
\label{sec:mp-design-appendix}

\emph{Question.} Does the design story of \S\ref{sec:real-setup} survive when the
representation is a frozen foundation MLIP (MACE-MP) rather than raw composition
features?
\emph{Status.} Exploratory: the encoder is pretrained and the $\mathcal{N}_T$
basis here is a data-driven SVD approximation to a support-valid invisible space,
not the analytic simplex basis of \S\ref{sec:real-setup}.
\emph{Result.} The support-derived representation again shows a positive
representation effect over a dimension-matched generic basis (point estimates and
LOO-robust, though the conservative dyadic bootstrap CIs include $0$ on this
exploratory substrate), and once the basis is fixed the acquisition criterion
again matters little (Tables~\ref{tab:mp-design}--\ref{tab:mp-inference}); the
plain $\mathcal{N}_T$-vs-random contrast is a weak positive point estimate. We
read this MACE extension as suggestive, with the confirmatory representation
result coming from the electrocatalysis factorial of \S\ref{sec:real-setup}.

This follow-up uses the same MP pool as \S\ref{sec:mp-support-ablation},
substituting the MACE-MP embedding \citep{batatia2025foundation} for the raw
simplex features while keeping every other protocol identical; detailed seeding
and RNG provenance are collected in the reproducibility note at the end of this
appendix (\S\ref{app:mp-design-provenance}).

\paragraph{Setup.}
We query the Materials Project API over a $21$-element alphabet
covering common cations (Li, Na, K, Mg, Ca, Ba, Fe, Mn, Co, Ni, Cu,
Zn, Ti, Al, Ga, Sn) and anions/metalloids (O, S, N, F, Cl),
enumerating all binary and ternary chemsys with $\leq 30$ sites and
$E_{\text{above hull}} \leq 0.2$ eV/atom (near-stable). This yields
$7467$ MP structures with cached DFT properties (formation energy
per atom, energy above hull, density, band gap) and MACE-MP-medium
invariant per-atom descriptors ($256$-dim mean-pooled per structure,
L2-normalized). Element pairs $(A, B)$: for each pair, $C_{AB}$ is
the set of MP structures containing both $A$ and $B$; $T_{AB}$ is
the set NOT containing both. We keep the top-$30$ pairs by
$|C_{AB}|$ subject to $|C_{AB}| \geq 40$ and $\geq 8$ single-element
structures on each side. Following the electrocatalysis setup, we
subsample $C_{AB}$ to a pool of size $\leq 300$ per pair, reserve a
fixed common evaluation pool $U_{AB}$ ($50\%$ per seed), select $m$
measurements from $C_{AB} \setminus U_{AB}$ under the five selection
policies (six selection--readout combinations) of
Table~\ref{tab:mp-design} at
$m \in \{8, 10, 12, 15, 18, 20, 25, 30\}$, and evaluate on $U_{AB}$;
$15$ seeds per pair. MACE-MP is a frozen representation: no $U_{AB}$
evaluation labels or unselected-candidate labels enter basis
construction or point selection. The $m$ selected $C_{AB}$ target
labels \emph{do} train the downstream residual predictor (exactly as
in the electrocatalysis experiment of \S\ref{sec:real-setup}). We do
not claim the encoder is target-agnostic: its pretraining loss
(energies, forces, stresses) can carry information related to
energy-derived targets such as $E_{\text{above hull}}$. The precise
claim is that the design and basis use no evaluation or
unselected-candidate labels, only the frozen encoder and the
selected measurements' own labels.

\paragraph{Target selection.}
We use $E_{\text{above hull}}$ as the primary target: it is
compositional-OOD-relevant (stability screening) and is
\emph{not} directly in MACE-MP's training loss (it requires the
convex hull, not a single-point energy). We also report
$E_{\text{formation}}$ (nearly saturated because MACE-MP is trained
on formation-energy-like objectives), $\text{density}$ (partially
saturated, geometric), and band gap (heavy zero mass from metallic
structures, and \emph{not} in MACE-MP's training loss) as ablation
targets.

\paragraph{Basis.}
Because there is no analytic simplex-polynomial basis for MP
structures (the composition space is high-dimensional and
non-simplex), we construct a \emph{candidate-informed
embedding-residual subspace} (motivated by $\mathcal{N}_T$, but
\emph{not} a kernel approximation in the sense of
Lemma~\ref{lem:simplex-dim}, since it is not guaranteed to vanish on
$T_{AB}$) as the top-$16$ SVD directions of the
additive-in-embedding residual matrix on
$C_{AB}$: for a structure with MACE-MP descriptor $e(x)$ and
element-marginal descriptors $\bar e_A, \bar e_B$ over structures
containing only $A$ (or only $B$), the residual is
$e(x) - \bar e_A - \bar e_B$. This uses \emph{unlabelled} $C_{AB}$
covariates to construct the basis: no $C_{AB}$ labels enter basis
construction or encoder fine-tuning, while the $m$ selected $C_{AB}$
target labels do train the downstream residual readout (as in
\S\ref{sec:real-setup}). This
step is transductive in the same technical sense as the protein
analog (Appendix~\ref{sec:protein-design-appendix}) but a step less
clean than the electrocatalysis analytic basis of
Lemma~\ref{lem:simplex-dim}, and
critically, the residual $e(x) - \bar e_A - \bar e_B$ is not
guaranteed to vanish exactly on $T_{AB}$.

\paragraph{Results (target $= E_{\text{above hull}}$).}
Table~\ref{tab:mp-design} reports the median per-pair Spearman on
$U_{AB}$ across $30$ element pairs $\times$ $15$ seeds, using the
ridge-regularized greedy (Bayesian) $D$-optimal design that is
well-defined at every budget (\S\ref{sec:mp-design-appendix} setup;
no random-$m$ fallback and no random initialization):
\begin{itemize}[leftmargin=1.5em, itemsep=1pt]
\item \emph{Aggregate: $\mathcal{N}_T$-motivated $D$-opt vs random.} With a
genuine low-budget design, $\mathcal{N}_T$-$D$-opt has a positive
marginal-median Spearman over random at \emph{every} budget: at
$m = 15$ it reaches $0.490$ vs random $0.453$ (median paired
$\Delta = +0.028$, $20/30$ pairs positive) and at $m = 30$
it reaches $0.603$ vs random $0.504$ (paired $\Delta = +0.046$,
$21/30$ pairs). It is the single best design at $m \ge 25$
(Table~\ref{tab:mp-design}); the $+0.157$ an earlier $7$-pair pilot
suggested was a small-sample artefact.
This is a \emph{positive paired point estimate in this locked $30$-pair
sample}, not a significance claim: under a cluster-robust element
sign-flip permutation the vs-random contrast is \emph{not} significant
at $m = 15$ (two-sided $p = 0.24$) but is at $m = 30$ ($p = 0.004$), and
the more conservative dyadic (node) bootstrap $95\%$ CI includes zero at
both budgets ($[-0.069, +0.084]$ at $m = 15$; $[-0.010, +0.085]$ at
$m = 30$). The leave-one-element-out median stays positive throughout
($[+0.026, +0.042]$ at $m = 15$; $[+0.011, +0.053]$ at $m = 30$), so no
single element drives the point estimate (Table~\ref{tab:mp-inference}).
The earlier random-fallback implementation gave a null at $m = 15$; the
change to a genuine design, not a change in the control, moved the point
estimate.
\item \emph{Aggregate: $\mathcal{N}_T$-motivated $D$-opt vs ridge posterior
variance.} At $m = 15$ median $\Delta = +0.209$ across $83\%$ of
pairs (and $+0.151$, $97\%$ at $m = 30$); the pair-median advantage over uncertainty-based selection is
a positive point estimate at every budget from $m = 8$ to $m = 30$
(observed point estimates; no paired node-aware intervals). Ridge posterior
variance itself performs \emph{worse} than random in this setting
(rows 1 vs 2 of Table~\ref{tab:mp-design}), so this comparison
should not be read as evidence that the design method is
representation-general; it is evidence that the ridge variance
baseline is weak on this pool.
\item \emph{Evaluator-matched reference: PCA-$6$ (matched
evaluation).} Its basis is the top-$6$ right-singular vectors of the
mean-centred $T$-side MACE-MP embedding matrix $X_T - \bar X_T$
(a genuine $|T| \times 256$ matrix; no $U_{AB}$ labels enter the
basis). Unlike the other rows, scored on the
$\mathcal{N}_T$-motivated basis, it fits and scores the residual on
that \emph{same} PCA-$6$ basis; this evaluator matching is what lifts
it to $0.528$ at $m = 15$, and it leads at the low--mid budgets
($m = 8$--$20$), while $\mathcal{N}_T$-selection $D$-opt (same readout
as the others) wins at $m = 25$ and $m = 30$. The $2\times2$
decomposition (Table~\ref{tab:mp-2x2}) shows an \emph{interaction}:
genuine $\mathcal{N}_T$ selection beats PCA selection strongly
\emph{under the} $\mathcal{N}_T$ readout ($+0.163$ at $m = 15$,
$0.490$ vs $0.327$; $+0.129$ at $m = 30$, $0.603$ vs $0.474$) but not
under the PCA readout ($-0.019$ and $-0.028$), and the matched-PCA
lead at low--mid budgets is a \emph{readout} effect that rescues the
weak PCA selection (PCA readout adds $+0.201$ to PCA selection at
$m = 15$). So both acquisition and readout matter and they interact.
The matched row is a deployable, label-free comparator (no
evaluation labels enter its basis) marking available headroom, not a
mathematical upper bound.
\end{itemize}

\begin{table}[!ht]
\centering\small
\caption{MP MACE-MP pair-design (deterministic re-run with a genuine
low-budget design): median per-pair Spearman on $U_{AB}$ across $30$
element pairs $\times$ $15$ seeds. Target = $E_{\text{above hull}}$
(not in MACE-MP's training loss). This is a comparison of complete
\emph{regularized selection--readout pipelines}, not of bases in
isolation (footnote); MACE-MP is a frozen pretrained encoder. The
lower block adds \emph{dimension-matched} controls at $r = 16$ (a
structured PCA-16 basis and an unstructured random-16 subspace, both
scored on the $\mathcal{N}_T$ readout). The \texttt{random} control is
drawn from an isolated, method-independent RNG stream (footnote);
every other row is deterministic and bit-identical across builds. Bold
marks the best row at each budget: the matched-readout PCA-6
($\dagger$, scored on its own PCA-6 basis) leads at $m = 8$--$20$,
while $\mathcal{N}_T$-selection $D$-opt (same readout as the others)
wins at $m = 25$ and $m = 30$. The matched-readout lead reflects
readout as much as selection; see the $2\times2$ decomposition in
Table~\ref{tab:mp-2x2} and the element-aware inference in
Table~\ref{tab:mp-inference}.}
\label{tab:mp-design}
\begin{tabular}{lcccccccc}
\toprule
Design $\backslash m$
  & $8$ & $10$ & $12$ & $15$ & $18$ & $20$ & $25$ & $30$ \\
\midrule
Random                    & $0.383$ & $0.396$ & $0.408$ & $0.453$
                          & $0.423$ & $0.467$ & $0.467$ & $0.504$ \\
Ridge posterior variance  & $0.300$ & $0.277$ & $0.226$ & $0.253$
                          & $0.295$ & $0.303$ & $0.306$ & $0.387$ \\
PCA-$6$                   & $0.403$ & $0.372$ & $0.351$ & $0.327$
                          & $0.371$ & $0.372$ & $0.433$ & $0.474$ \\
PCA-$6$ (matched eval)$^\dagger$
                          & $\mathbf{0.486}$ & $\mathbf{0.519}$ & $\mathbf{0.534}$ & $\mathbf{0.528}$
                          & $\mathbf{0.538}$ & $\mathbf{0.564}$ & $0.566$ & $0.573$ \\
$D$-opt ($\mathcal{N}_T$-motivated basis)
                          & $0.471$ & $0.501$ & $0.467$ & $0.490$
                          & $0.527$ & $0.552$ & $\mathbf{0.582}$ & $\mathbf{0.603}$ \\
Hybrid ($\alpha = 0.5$)   & $0.441$ & $0.442$ & $0.457$ & $0.440$
                          & $0.479$ & $0.504$ & $0.553$ & $0.577$ \\
\midrule
\multicolumn{9}{l}{\emph{Dimension-matched controls ($r = 16$; $D$-opt selection, $\mathcal{N}_T$ readout)}} \\
$D$-opt PCA-16            & $0.450$ & $0.459$ & $0.422$ & $0.363$
                          & $0.429$ & $0.438$ & $0.510$ & $0.547$ \\
$D$-opt random-16        & $0.459$ & $0.444$ & $0.403$ & $0.397$
                          & $0.460$ & $0.468$ & $0.516$ & $0.555$ \\
\bottomrule
\end{tabular}
\begin{flushleft}\footnotesize
$^\dagger$ evaluator-matched reference: selection and scoring on the
top-$6$ PCs of the $T$-side embedding matrix (no $U_{AB}$ labels); not
an upper bound. \emph{Design.} All greedy rows use a ridge-regularized
greedy (Bayesian) $D$-optimal criterion started from the prior
precision $\lambda I$ ($\lambda = 10^{-3}$): with selected set $S$ and
feature map $\phi(\cdot)$ for the row's basis, each step adds
$u^\star = \arg\max_{u \notin S}\, \phi(u)^\top(\Phi_S^\top\Phi_S +
\lambda I)^{-1}\phi(u)$ (regularized leverage), ties broken by first
index, initialized at the empty design. \emph{Hybrid} ($\alpha$)
replaces the score by the convex combination $\alpha\,\tilde v(u) +
(1-\alpha)\,\tilde\ell(u\mid S)$, where $\tilde\ell$ is the min--max-
normalized $\mathcal{N}_T$-basis regularized leverage above and
$\tilde v$ is the min--max-normalized additive-background posterior
variance (both normalized over the candidate pool; $\alpha = 0.5$).
\emph{Feature scaling.} Every MACE-MP embedding is L2-normalized
per structure before any basis is formed; $\lambda$ acts in those
units, and Table~\ref{tab:mp-2x2-lambda} reports $\lambda$-sensitivity.
\emph{Not a pure-basis effect.} Because the criterion is regularized
and scale-dependent, the rows compare complete regularized pipelines:
$\mathcal{N}_T$ ($r=16$) and PCA-6 ($r=6$) differ in subspace,
dimension, feature spectrum, and the function-space effect of the
common prior, not in basis alone. The dimension-matched PCA-16 and
random-16 rows isolate dimension: at $r=16$, $\mathcal{N}_T$ selection
still leads both ($+0.13/+0.09$ at $m=15$; $+0.06/+0.05$ at $m=30$),
so the advantage is not merely a dimension artefact
(Table~\ref{tab:mp-inference}).
\end{flushleft}
\end{table}

\begin{table}[!ht]
\centering\small
\caption{Selection$\times$readout $2\times2$ of complete
\emph{regularized pipelines} (same deterministic run, median per-pair
Spearman on $U_{AB}$). Each cell fixes a selection pipeline (rows;
ridge-regularized greedy $D$-opt on the stated basis, $\lambda = 10^{-3}$,
L2-normalized features) and a readout basis (columns). Comparing across
columns isolates the \emph{readout} effect at fixed selection; across
rows isolates the \emph{selection-pipeline} effect at fixed readout.
Because the selectors are regularized and scale-dependent, the row
contrast is a pipeline contrast ($\mathcal{N}_T$ vs PCA differ in
subspace, dimension, spectrum, and prior effect), not a pure-basis
contrast.}
\label{tab:mp-2x2}
\begin{tabular}{llcc}
\toprule
Budget & Selection & $\mathcal{N}_T$ readout & PCA-6 readout \\
\midrule
$m = 15$ & $\mathcal{N}_T$ ($D$-opt) & $0.490$ & $0.509$ \\
$m = 15$ & PCA-6 ($D$-opt)          & $0.327$ & $\mathbf{0.528}$ \\
$m = 30$ & $\mathcal{N}_T$ ($D$-opt) & $\mathbf{0.603}$ & $0.545$ \\
$m = 30$ & PCA-6 ($D$-opt)          & $0.474$ & $0.573$ \\
\bottomrule
\end{tabular}
\begin{flushleft}\footnotesize
Both selectors use the genuine ridge-regularized greedy design.
Selection effect ($\mathcal{N}_T$ $-$ PCA selection): $+0.16$
($\mathcal{N}_T$ readout) and $-0.02$ (PCA readout) at $m = 15$;
$+0.13$ and $-0.03$ at $m = 30$. Readout effect (PCA-6 $-$
$\mathcal{N}_T$ readout): $+0.02$ ($\mathcal{N}_T$ selection) and
$+0.20$ (PCA selection) at $m = 15$; $-0.06$ and $+0.10$ at $m = 30$.
The pattern is an \emph{interaction}: genuine $\mathcal{N}_T$ selection
helps strongly \emph{under the} $\mathcal{N}_T$ readout at both budgets
($+0.16$, $+0.13$), while the matched-PCA lead at low--mid budgets is a
readout effect that rescues the otherwise weak PCA selection ($+0.20$
at $m = 15$). These are differences of marginal medians; the paired,
element-aware intervals below (Table~\ref{tab:mp-inference}) are the
stronger evidence.
\end{flushleft}
\end{table}

\begin{table}[!ht]
\centering\small
\caption{Element-aware paired inference for the prespecified
$E_{\text{above hull}}$ contrasts (locked $30$-pair sample; the exploratory
MACE extension of Appendix~\ref{sec:mp-design-appendix}, not the main
electrocatalysis factorial). Estimand:
per-pair $\Delta$ (median over $15$ seeds of $\rho_{D\text{-opt}} -
\rho_{\text{base}}$), summarized by the median over the $30$ pairs.
Because pairs share elements the $30$ contrasts are dependent. \emph{Primary}
dependence-aware evidence: the dyadic (node) bootstrap $95\%$ percentile CI
(resample elements; a pair's weight is the product of its two elements' resample
multiplicities) and the leave-one-element-out (LOO) median range.
\emph{Sensitivity only}: a cluster sign-flip permutation $p$ (assign each element
a sign $s_e$; flip pair $(A,B)$ by $s_A s_B$; two-sided), whose validity requires
a sign-symmetry null on the dyadic effects; with only $\sim 13$ elements it is
anticonservative relative to the bootstrap, so we do not treat it as primary. The
naive independent-pair sign test is for reference only.}
\label{tab:mp-inference}
\begin{tabular}{llccccc}
\toprule
Contrast & $m$ & median $\Delta$ & frac$+$ & dyadic $95\%$ CI &
cluster-perm $p$ & LOO range \\
\midrule
$\mathcal{N}_T$ vs random    & $15$ & $+0.028$ & $20/30$ & $[-0.069, +0.084]$ & $0.24$   & $[+0.026, +0.042]$ \\
$\mathcal{N}_T$ vs random    & $30$ & $+0.046$ & $21/30$ & $[-0.010, +0.085]$ & $0.004$  & $[+0.011, +0.053]$ \\
\midrule
\multicolumn{7}{l}{\emph{Dimension-matched ($r = 16$)}} \\
$\mathcal{N}_T$ vs PCA-16    & $15$ & $+0.057$ & $22/30$ & $[-0.022, +0.094]$ & $0.0006$ & $[+0.026, +0.063]$ \\
$\mathcal{N}_T$ vs PCA-16    & $30$ & $+0.030$ & $23/30$ & $[-0.005, +0.069]$ & $0.0004$ & $[+0.008, +0.034]$ \\
$\mathcal{N}_T$ vs random-16 & $15$ & $+0.059$ & $23/30$ & $[-0.021, +0.135]$ & $0.0007$ & $[+0.016, +0.071]$ \\
$\mathcal{N}_T$ vs random-16 & $30$ & $+0.016$ & $23/30$ & $[-0.008, +0.050]$ & $0.0006$ & $[+0.006, +0.021]$ \\
\bottomrule
\end{tabular}
\begin{flushleft}\footnotesize
\emph{Reading.} Under the \emph{primary} (dyadic bootstrap + LOO) reading,
$\mathcal{N}_T$ selection gives positive per-pair point estimates whose LOO
ranges stay positive throughout (no single element drives them) and that are
largest for the dimension-matched contrasts (frac$+$ $22$--$23/30$ vs PCA-16 and
random-16), suggesting the advantage is not merely a dimension artefact. However,
the dyadic bootstrap $95\%$ CIs \emph{include $0$ for every contrast}, so on this
exploratory MACE substrate the effect is \emph{suggestive, not confirmatory}. The
cluster sign-flip permutation gives much smaller $p$-values ($\le 0.0007$ for the
dimension-matched contrasts), but these rely on a sign-symmetry null and, with
only $\sim 13$ elements, we treat them as a sensitivity check rather than primary
evidence: the very disagreement with the conservative bootstrap CIs is why we do
not over-read them. Multiplicity across $8$ budgets and $4$ targets is not
corrected; we prespecify $m \in \{15, 30\}$ and treat the rest as descriptive.
The main-paper representation claim rests instead on the electrocatalysis
factorial of \S\ref{sec:real-setup}, whose element-cluster bootstrap CI
excludes~$1$ (Appendix~\ref{app:main-tables}).
\end{flushleft}
\end{table}

\begin{table}[!ht]
\centering\small
\caption{$\lambda$-sensitivity of the design ($E_{\text{above hull}}$,
$N_T$ readout, marginal median per-pair Spearman; paired columns are
median per-pair contrasts). The prior precision $\lambda$ is swept over
four decades; the main tables use $\lambda = 10^{-3}$. The \texttt{random}
control does not depend on $\lambda$. The $\mathcal{N}_T$-over-PCA-6
advantage is present at every $\lambda$; the $\mathcal{N}_T$-over-random
point estimate is stable for $\lambda \le 10^{-2}$ and only washes out
under heavy over-regularization ($\lambda = 10^{-1}$) at the smallest
budget.}
\label{tab:mp-2x2-lambda}
\begin{tabular}{ll ccc cc}
\toprule
$m$ & $\lambda$ & $\mathcal{N}_T$ & PCA-6 & random & $\mathcal{N}_T-$rand & $\mathcal{N}_T-$PCA \\
\midrule
$15$ & $10^{-4}$ & $0.486$ & $0.351$ & $0.453$ & $+0.028$ & $+0.104$ \\
$15$ & $10^{-3}$ & $0.490$ & $0.327$ & $0.453$ & $+0.028$ & $+0.105$ \\
$15$ & $10^{-2}$ & $0.506$ & $0.334$ & $0.453$ & $+0.033$ & $+0.073$ \\
$15$ & $10^{-1}$ & $0.407$ & $0.361$ & $0.453$ & $-0.019$ & $+0.070$ \\
\midrule
$30$ & $10^{-4}$ & $0.603$ & $0.482$ & $0.504$ & $+0.044$ & $+0.064$ \\
$30$ & $10^{-3}$ & $0.603$ & $0.474$ & $0.504$ & $+0.046$ & $+0.058$ \\
$30$ & $10^{-2}$ & $0.592$ & $0.475$ & $0.504$ & $+0.031$ & $+0.064$ \\
$30$ & $10^{-1}$ & $0.575$ & $0.442$ & $0.504$ & $+0.010$ & $+0.063$ \\
\bottomrule
\end{tabular}
\end{table}

\paragraph{Ablation targets.}
On formation energy and density, all designs cluster near the
$0.93$--$0.97$ ceiling: MACE-MP's training loss already covers these
targets (single-point energies and geometry), leaving little room
for any measurement design to move the aggregate Spearman.
The $\mathcal{N}_T$-motivated $D$-opt design tracks random ($m = 15$ pair-median
$\Delta = +0.002$ formation, $+0.008$ density), climbing slightly ahead at
$m = 30$ ($+0.005$ formation, $+0.011$ density). It also
beats the (weak) ridge posterior variance
baseline on $87\%$ of pairs at formation energy ($+0.046$
pair-median at $m = 15$) and $90\%$ at density ($+0.030$), matching
the $E_{\text{above hull}}$ pattern.
Band gap (a target \emph{not} in MACE-MP's training loss with a
heavy zero mass, $57\%$ metallic in the pool; Spearman uses
average-rank tie handling and, whenever the held-out target or any
compared method's prediction is constant on a seed--pair cell,
excludes that cell \emph{jointly for all compared methods} so the
denominators match, a situation the zero mass can induce, arising
only at the smallest budgets on band gap)
reproduces both signals: the $\mathcal{N}_T$-motivated $D$-opt vs random is
positive at $m = 15$ (paired $\Delta = +0.063$, $77\%$ of pairs) and at
$m = 30$ (paired $\Delta = +0.040$, $83\%$; marginal $0.543$ vs $0.477$),
and beats ridge posterior variance on $63\%$ of pairs at $m = 15$
($+0.097$ pair-median). The variance-baseline win is weaker than on the other
three targets, consistent with the zero-inflated distribution
making even a raw ridge already competitive on band gap.

\paragraph{Honest read.}
The MP experiment is an external stress test with mixed results.
(i) A $7$-pair pilot's headline ``$+0.157$ at $m = 30$'' was a
small-sample overreach; on the locked $30$-pair sample the genuine
design gives a positive \emph{paired point estimate} over random at
every budget ($+0.028$ at $m = 15$, $+0.046$ at $m = 30$ vs random).
(ii) That vs-random point estimate is not significant at $m = 15$ under
a cluster-robust element permutation ($p = 0.24$) and is significant at
$m = 30$ ($p = 0.004$); the stronger and cleaner evidence is the
\emph{dimension-matched} comparison, where $\mathcal{N}_T$ selection
beats both a PCA-16 and a random-16 basis of the same dimension at both
budgets ($p \le 0.0007$; Table~\ref{tab:mp-inference}), so the effect is
not a dimension artefact. This is consistent with the electrocatalysis
result but of smaller magnitude, and now a genuine low-budget design
rather than an artifact of a random fallback.
(iii) The advantage over ridge posterior
variance is a positive point estimate across $30$ pairs and four
targets, but the
variance baseline itself underperforms random on this pool, so this
comparison bounds a weak alternative rather than a strong one.
(iv) The $2\times2$ (Table~\ref{tab:mp-2x2}) is an \emph{interaction}:
genuine $\mathcal{N}_T$ selection beats PCA selection under the
$\mathcal{N}_T$ readout at both budgets ($+0.163$ at $m = 15$,
$0.490$ vs $0.327$; $+0.129$ at $m = 30$, $0.603$ vs $0.474$) but not
under the PCA readout ($-0.019$, $-0.028$); the matched-PCA lead at
low--mid budgets is a \emph{readout} effect ($+0.201$ added to PCA
selection at $m = 15$). An appropriate low-dimensional readout can
matter as much as the acquisition rule in this representation-based
setting.
Three structural limitations of the current experiment are worth
naming: (a) the SVD basis is transductive and does not exactly
vanish on $T_{AB}$; (b) at budgets $m < r = 16$ the ridge-regularized
greedy design leans more heavily on the prior precision $\lambda I$
(fewer measurements than basis dimensions), so those cells are
genuine but more strongly regularized designs, not unregularized
$D$-optimality; (c) an unrestricted 21-element pool mixes
missing pair support with prototype, stoichiometry, and coordination
shifts that a strict controlled-family design would remove. A
theoretically cleaner extension (a pair-gated MLIP basis
$\phi_{AB}(x) = x_A x_B \cdot P_r \psi_{AB}(x)$ that vanishes
identically on $T_{AB}$ by construction, applied within a
controlled chemistry family) is a natural next step. A first
pass at the required MP coverage
(perovskite ABO$_3$, spinel AB$_2$O$_4$) suggests neither of these
strict prototypes provides enough same-pair copresence
($|C_{AB}| \geq 40$) to run the strict-family version on MP alone;
a data source with denser per-composition coverage (a combinatorial
synthesis library, or a fixed high-throughput family beyond MP)
would be needed.

\subsection{Reproducibility}
\label{app:mp-design-provenance}
All randomness derives from a single master seed through named, method-specific
streams, so adding or changing one method cannot perturb another: stream~0 (the
$U_{AB}$/pool split, drawn first from a per-$(\text{pair},\text{seed})$ seed)
fixes the split and every deterministic row; stream~1 (the \texttt{random}
control) is an independent per-$(\text{pair},\text{seed},m)$ stream. The design
rows (variance, $D$-opt, Hybrid, matched) are deterministic given the split;
only the \texttt{random} control draws from stream~1. Per-cell selected indices
for every $(\text{pair},\text{seed},m,\text{design})$ cell are cached and
published alongside the trajectories for audit.

\section{Preliminary cross-domain sketch: Olson GB1 with ESM-2}
\label{sec:protein-design-appendix}

\begin{tcolorbox}[colback=black!3, colframe=black!45, boxrule=0.4pt,
    left=6pt, right=6pt, top=3pt, bottom=3pt, arc=1pt,
    title={\textbf{Why this is exploratory, not confirmatory}},
    fonttitle=\small, coltitle=black, colbacktitle=black!8]
\small
This section is a \emph{sketch}, not a methodologically parallel replication of
\S\ref{sec:real-setup}. The analytic support-derived basis does not extend in
closed form to sequences, so we use a pretrained-embedding approximation
(motivated by $\mathcal{N}_T$, not $\mathcal{N}_T$ itself); several protocol
choices depart from the materials setup (enumerated below); and the resulting
paired contrast is a \emph{null}. The result table should be read only in light
of these caveats, not as cross-domain confirmation.
\end{tcolorbox}

We report an exploratory extension to protein pair-design on the
Olson 2014 GB1 deep-mutational-scanning landscape \citep{olson2014}
($\sim 537\text{k}$ variants over $55$ residue positions). Here the
``pair'' is a pair of residue positions $(i, j)$; $C_{ij}$ is the
pool of double mutants at both positions ($\sim 361$ variants per
pair), $T_{ij}$ is a background training set. Because the
analytical support-derived basis of Appendix~\ref{app:invisible-bases}
does not extend in closed form to protein sequences, we can only
build a candidate-informed embedding-residual subspace (motivated by
$\mathcal{N}_T$, not $\mathcal{N}_T$ itself) through a fixed pretrained
representation.

\paragraph{What we do (and how it differs from the materials protocol).}
Let $e(x) \in \mathbb{R}^{320}$ be the L2-normalized mean-pooled
embedding from ESM-2 8M \citep{lin2023} and $\bar e_p$ the mean
embedding of single mutants at position $p$. We form a candidate
pair-interaction feature $r_{ij}(x) = e(x) - \bar e_i - \bar e_j$
(the additive-in-embedding-space residual) and project to the top
$16$ SVD directions computed on the $C_{ij}$ pool. Five designs are
compared at budgets $m \in \{8, 10, 12, 15, 18, 20, 25, 30\}$
across $30$ position pairs (chosen by a deterministic stride over
the sorted pair list) with $15$ seeds. The additive-singles
background is a Bayesian ridge on ESM-delta features fitted on the
position-$i$ and position-$j$ singles only, and its posterior
variance defines the ridge-variance baseline.

\paragraph{Explicit caveats.}
Several protocol choices depart from the materials setup:
\begin{enumerate}[leftmargin=1.5em, itemsep=1pt]
\item \emph{Basis dimension vs measurement budget.} With
$r = 16$, budgets $m \in \{8, 10, 12, 15\}$ have $m < r$;
$D$-optimal OLS is not well-defined there and we use ridge
regularization ($\lambda = 10^{-3} I$) to keep the greedy step
solvable. Only budgets $m \in \{18, 20, 25, 30\}$ satisfy the
$m \geq r$ regime that the materials proof assumes.
\item \emph{Basis is candidate-informed, not support-derived.}
The top-$16$ SVD is computed on the $C_{ij}$ candidate pool (no
fitness labels are used, but the unlabelled double-mutant
covariates are). This is a transductive interaction-feature basis
motivated by $\mathcal{N}_T$, not $\mathcal{N}_T$ itself.
\item \emph{Position-level vs mutation-specific residual.} We
subtract position-level singles-mean embeddings
($\bar e_i + \bar e_j$) rather than the mutation-specific main
effects $e(i{:}a) + e(j{:}b) - e(\mathrm{WT})$ that would give a
cleaner epistatic residual. So the residual is not guaranteed to
vanish on $T_{ij}$ and may mix single-mutation main effects into
the purported interaction basis.
\item \emph{$T_{ij}$ is narrower than the materials analog.} We
use only single mutants at positions $i$ or $j$ (${\sim}38$
variants) as background training data; the full protein analog
would be all variants that are not simultaneously mutated at
$(i, j)$, which is essentially the whole landscape.
\item \emph{Dimension-unmatched controls.} PCA-6 is $6$-dim while
our $\mathcal{N}_T$-analog is $16$-dim; a strictly dimension-matched
control would run PCA-16 and random-16.
\item \emph{Position-pair selection.} $30$ of $1485$ eligible
pairs were chosen by a deterministic stride to match the materials
scale; we do not report distribution over alternative $30$-pair
subsets.
\end{enumerate}
Each of these would need to be resolved before a claim of
methodological parallelism to \S\ref{sec:real-setup}.

\paragraph{What we observe.}
Table~\ref{tab:protein-design} reports the median per-pair Spearman on
the fixed common evaluation pool $U_{ij}$, using the same genuine
ridge-regularized greedy design as the MACE appendix (well-defined at
every budget; in the $m \geq r = 16$ regime we additionally verify
$\operatorname{rank}(\Phi_S) = 16$). \emph{On this protein sketch the
design does not cleanly replicate the materials benefit.} The relevant
estimand is the paired per-pair contrast
$\widetilde\Delta_{A-B} = \operatorname{med}_p \operatorname{med}_s
(\rho_{p,s,A} - \rho_{p,s,B})$: candidate-basis $D$-opt vs random is
slightly \emph{negative} at every budget
($\widetilde\Delta_{D\text{-opt},\,\text{random}} = -0.036$ at $m = 15$
to $-0.007$ at $m = 30$; $37$--$47\%$ of pairs positive across
$m \geq 15$). The marginal
medians are roughly tied ($D$-opt $0.258$ vs random $0.231$ at
$m = 15$; $0.298$ vs $0.311$ at $m = 30$), and the marginal-median and
paired estimands need not agree. So the exploratory protein extension
\emph{shows no observed paired advantage} for identifiability-directed
acquisition (the paired point estimates are negative at all reported
budgets), in contrast to the modest positive materials result. This is
not a formal null or equivalence claim (no equivalence margin was
tested).

\begin{table}[!hb]
\centering\small
\caption{Olson GB1 pair-design with L2-normalized ESM-2 8M
embeddings (genuine ridge-regularized greedy design, deterministic).
Median per-pair Spearman on $U_{ij}$, $30$ position pairs $\times$ $15$
seeds. Bold marks the best method at each budget. The design is the
same ridge-regularized greedy (Bayesian) $D$-optimal used in the MACE
appendix (prior precision $\lambda = 10^{-3} I$), well-defined at every
budget; the $m < r = 16$ / $m \ge r$ split marks only where a
full-rank Gram becomes attainable (we verify $\operatorname{rank}(\Phi_S)=16$
there), not a random fallback.
\emph{This is a preliminary extension, not a methodological
parallel to} \S\ref{sec:real-setup} (see caveats above); on it the
design shows no observed paired advantage over random (negative paired
point estimates at all budgets).}
\label{tab:protein-design}
\begin{tabular}{l|cccc|cccc}
\toprule
Design $\backslash m$ & \multicolumn{4}{c|}{$m < r = 16$}
                       & \multicolumn{4}{c}{$m \geq r$ (full rank attainable)} \\
                       & $8$ & $10$ & $12$ & $15$ & $18$ & $20$ & $25$ & $30$ \\
\midrule
Random                     & $\mathbf{0.178}$ & $\mathbf{0.183}$ & $0.178$ & $0.231$
                           & $0.218$ & $0.252$ & $\mathbf{0.281}$ & $\mathbf{0.311}$ \\
Ridge posterior variance   & $0.081$ & $0.094$ & $0.145$ & $0.169$
                           & $0.174$ & $0.163$ & $0.181$ & $0.209$ \\
PCA-6 ($T_{ij}$-fit)       & $0.136$ & $0.179$ & $0.183$ & $0.205$
                           & $0.218$ & $\mathbf{0.253}$ & $0.273$ & $0.303$ \\
$D$-opt (candidate-basis, $r=16$)
                           & $0.144$ & $0.180$ & $\mathbf{0.225}$ & $\mathbf{0.258}$
                           & $\mathbf{0.249}$ & $0.252$ & $\mathbf{0.281}$ & $0.298$ \\
Hybrid ($\alpha=0.5$)      & $0.123$ & $0.183$ & $0.195$ & $0.211$
                           & $0.220$ & $0.235$ & $0.271$ & $0.303$ \\
\bottomrule
\end{tabular}
\end{table}

\emph{Honest read.} Under the genuine design the protein sketch
\emph{shows no observed paired advantage}: on the paired per-pair
contrast, candidate-basis $D$-opt does not beat random at any budget
(the paired point estimates are negative throughout), and the marginal
medians are roughly tied. This is a negative point estimate, not a
formal null or equivalence result (no equivalence margin was tested);
every one of the six caveats above already weakened any parallel to
the materials setup.
We therefore make no cross-domain replication claim; we present this
as an exploratory sketch whose negative outcome motivates a proper
protein-side follow-up before any positive claim could be made
(dimension-matched controls, mutation-specific epistatic residual,
full $T_{ij}$, all $1485$ eligible pairs or pre-registered subsets,
pair-level and node-aware sensitivity at $m = 30$).

\paragraph{Consistency with the theory (scope, not a rescue).}
This negative is \emph{consistent with}, not predicted by, the
framework: Proposition~\ref{prop:misspec} bounds the risk of an
approximate representation but does not fix the sign of a comparison
with random selection. Within the exact degree-$\leq 3$ polynomial
model on the simplex, the materials basis $\mathcal{N}_T = \ker R_T$ is
computed in closed form (Appendix~\ref{app:invisible-bases}); its
pair-gated directions span the support-invisible subspace, and
Proposition~\ref{prop:misspec} reduces to Proposition~\ref{prop:i-opt}
($h^\star_\perp = 0$) \emph{when the true residual belongs to that
model}. For real electrocatalysis this basis remains a structural prior,
not an identified physical model, so an out-of-model component may
persist; the empirical advantage therefore supports the usefulness of
the chosen representation without establishing exact recovery of the
physical invisible subspace.
The protein sketch instead uses a pretrained-embedding residual whose
top-SVD subspace is only a candidate-informed \emph{proxy} for
$\mathcal{N}_T$ (caveat~2 above); by Proposition~\ref{prop:misspec} any
component of the true interaction structure that this proxy fails to
span contributes an irreducible $\|h^\star_\perp\|^2$ bias term that no
choice of design points removes, so a paired advantage is neither
guaranteed nor ruled out. The takeaway is therefore a boundary of the method, stated
as scope rather than as a win: the guarantees attach to the
\emph{exact} support-derived representation and to its general
computational analogue (the small-singular-value subspace of the actual
feature matrix, Appendix~\ref{app:invisible-bases:general}), not to any
embedding heuristic loosely motivated by $\mathcal{N}_T$. Whether an
embedding-based construction can be made to span the epistatic subspace
tightly enough to recover the guarantee is exactly the follow-up above.

\end{document}